\documentclass{article}

\PassOptionsToPackage{numbers, sort, compress}{natbib}

\usepackage[final]{neurips_2018}

\usepackage[utf8]{inputenc} 
\usepackage[T1]{fontenc}    
\usepackage{hyperref}       
\usepackage{url}            
\usepackage{booktabs}       
\usepackage{amsfonts}       
\usepackage{nicefrac}       
\usepackage{microtype}      

\usepackage{calc}
\usepackage{amsmath, amssymb, amsthm}
\usepackage{DejaVuSans}
\usepackage{parskip}
\usepackage{color}
\usepackage{epsfig}
\usepackage{subfig}
\usepackage{verbatim}
\usepackage{rotating}
\usepackage[ruled,noend]{algorithm2e}
\usepackage{algorithmic}
\usepackage{enumitem}
\usepackage{wrapfig}
\usepackage{cleveref}
\usepackage{bbm}
\usepackage{array}
\usepackage{soul}
\usepackage{xspace}
\usepackage{xcolor}
\usepackage{colortbl}
\usepackage{multirow}
\usepackage{kky}
\usepackage{koopmansDefns}

\title{\Large Neural Architecture Search \\
with Bayesian Optimisation and Optimal Transport}

\newcommand{\authspace}{\;\,}

\author{
  Kirthevasan Kandasamy,\authspace
  Willie Neiswanger,\authspace
  Jeff Schneider,\authspace
  Barnab\'as P\'oczos,\authspace
  Eric P Xing \\
  Carnegie Mellon University, \quad Petuum Inc. \\
  \incmtt{\{kandasamy, willie, schneide, bapoczos, epxing\}@cs.cmu.edu} \\
}

\begin{document}

\maketitle


\newcommand{\insertFigNNEgs}{
\newcommand{\nnfigwidth}{1.05in}
\newcommand{\nnfigheight}{2.38in}
\newcommand{\nnfighsp}{\hspace{0.025in}}
\begin{figure*}
\begin{center}
\centering
\subfloat[]{
\includegraphics[height=\nnfigheight]{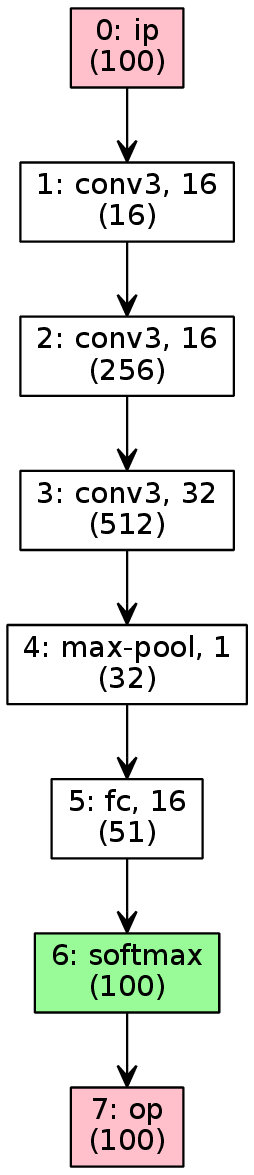}
\label{fig:nneg1}}\nnfighsp
\subfloat[]{
\includegraphics[height=\nnfigheight]{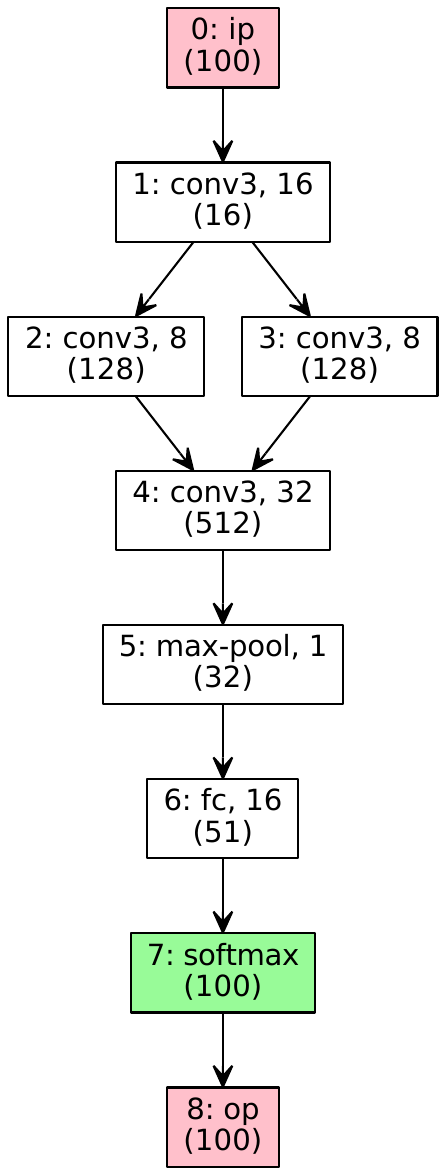} \nnfighsp
\label{fig:nneg2}}\nnfighsp
\subfloat[]{
\includegraphics[height=\nnfigheight]{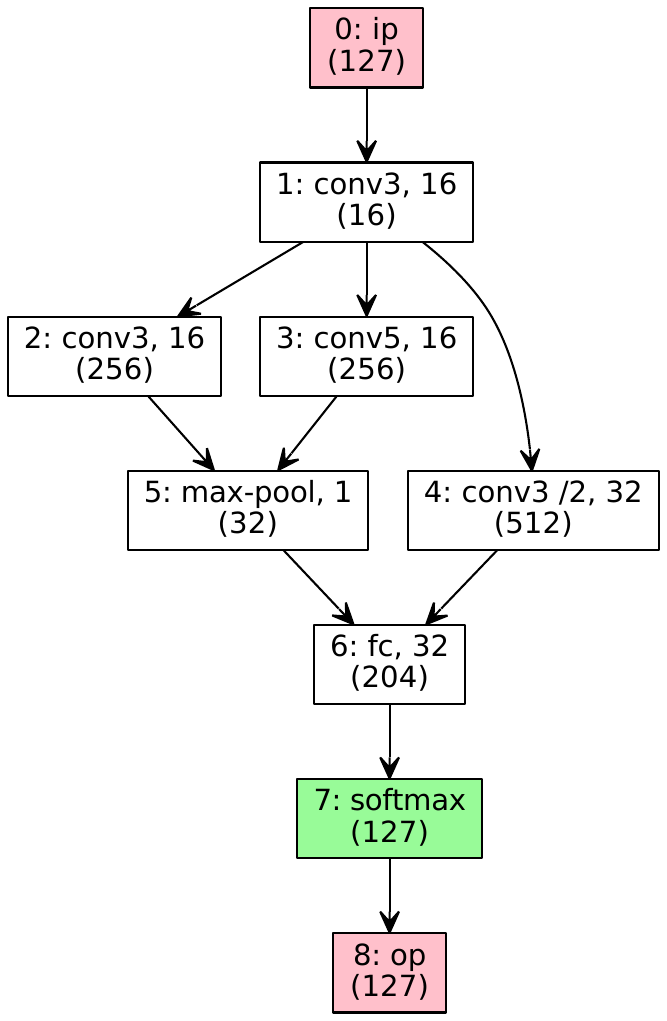} \nnfighsp
\label{fig:nneg3}}\nnfighsp
\subfloat[]{
\includegraphics[height=\nnfigheight]{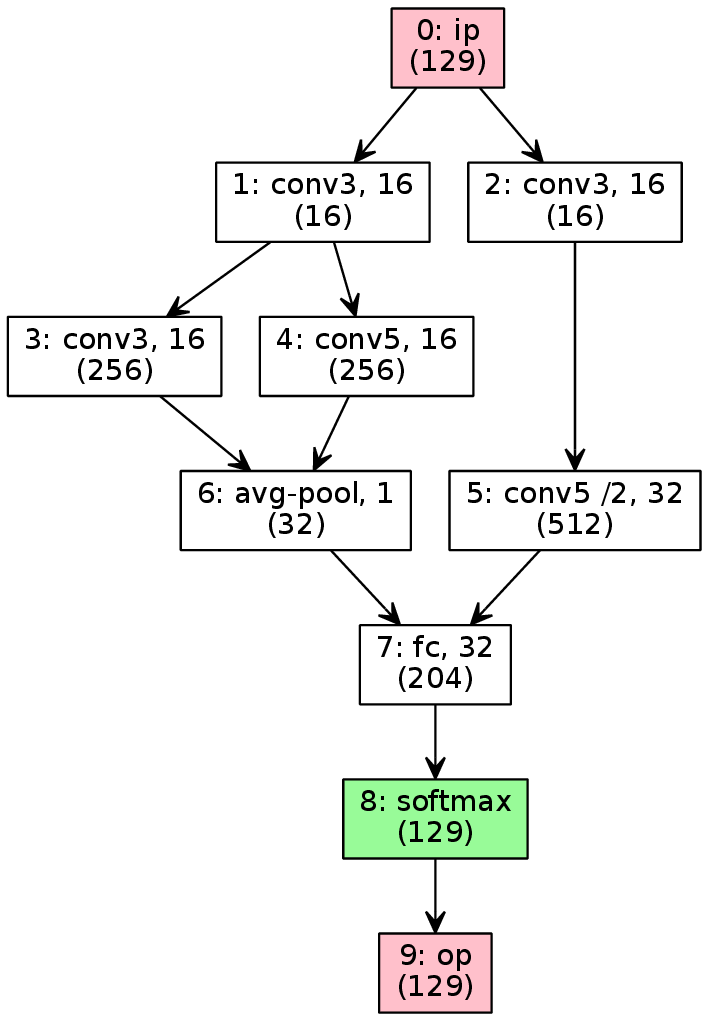}
\label{fig:nneg4}}\nnfighsp
\subfloat[]{
\includegraphics[height=\nnfigheight]{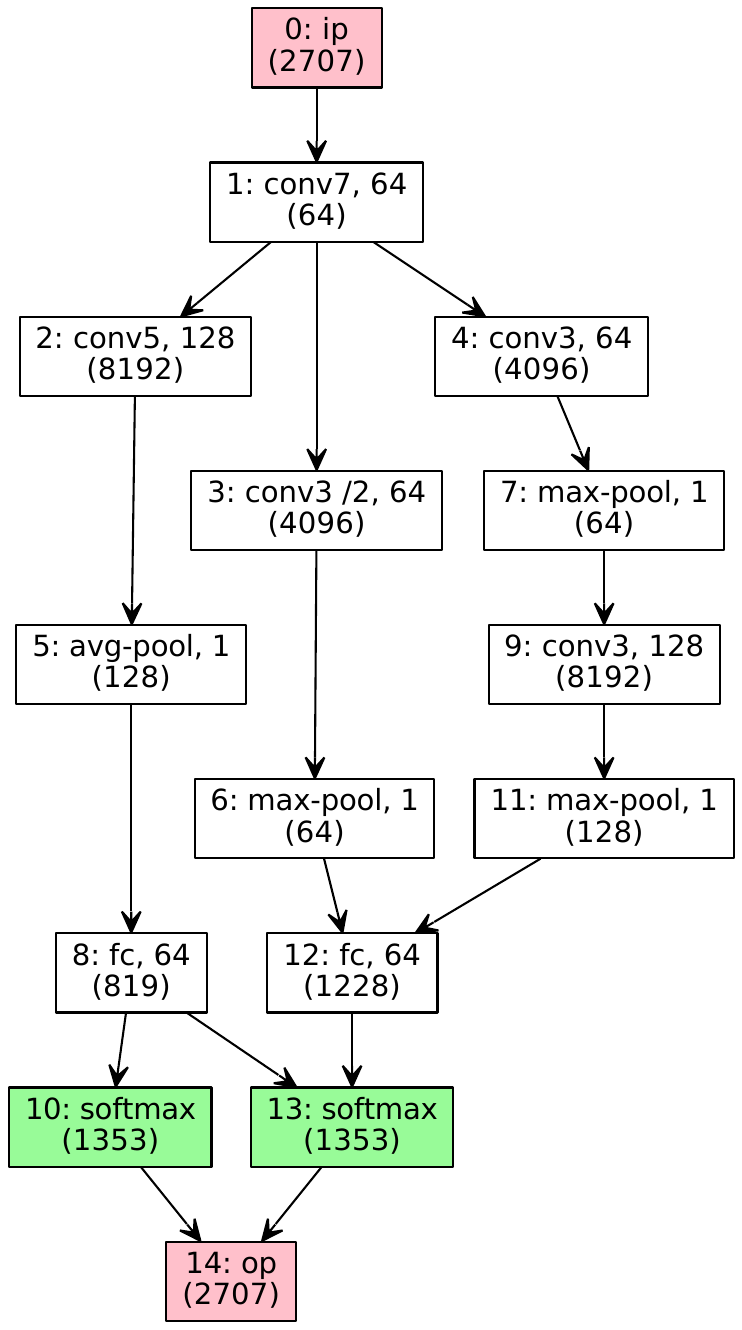}
\label{fig:nneg5}}\nnfighsp
\vspace{-0.10in}
\caption{
An illustration of some CNN architectures.
In each layer, $i$: indexes the layer, followed by the label (e.g \convthree),
and followed by the number of units (e.g. number of convolutional filters).
`\inlabelfont{/2}' indicates stride 2 when performing convolutions.
The layer mass is denoted in parantheses.
The input and output layers are shaded in pink while the decision (\softmax) layers
are shaded in green.
\label{fig:nnegs}
}
\end{center}
\end{figure*}
}

\newcommand{\insertFigMLPEgs}{
\newcommand{\mlpfigwidth}{1.1in}
\newcommand{\mlpfigheight}{2.38in}
\newcommand{\mlpfighsp}{\hspace{0.025in}}
\begin{figure*}
\begin{center}
\centering
\subfloat[]{
\includegraphics[height=\mlpfigheight]{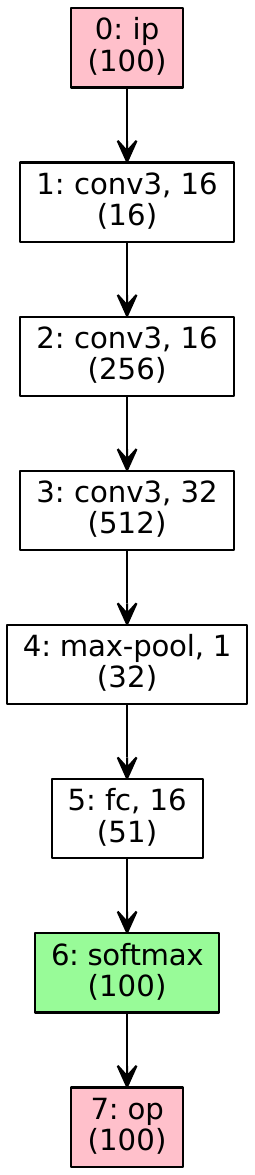}
\label{fig:mlpeg1}}\mlpfighsp
\subfloat[]{
\includegraphics[height=\mlpfigheight]{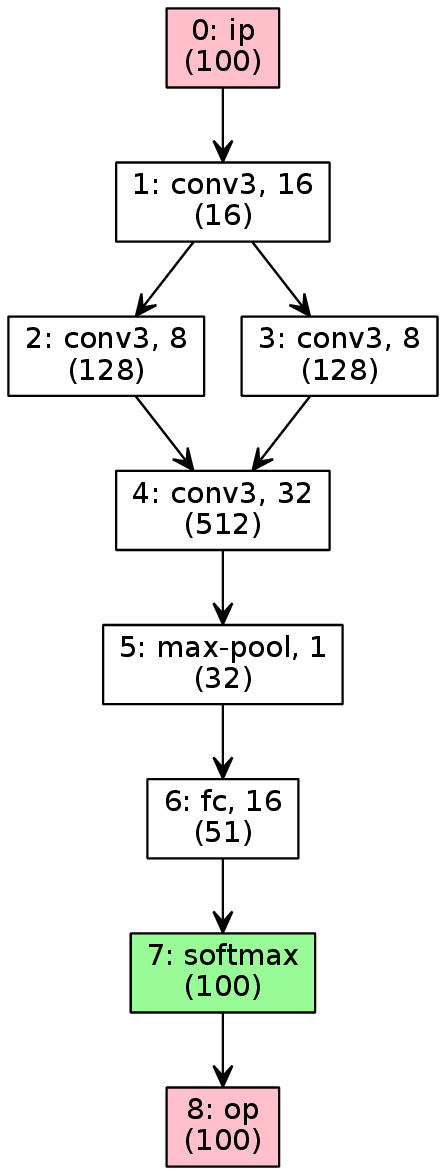} \mlpfighsp
\label{fig:mlpeg2}}\mlpfighsp
\subfloat[]{
\includegraphics[height=\mlpfigheight]{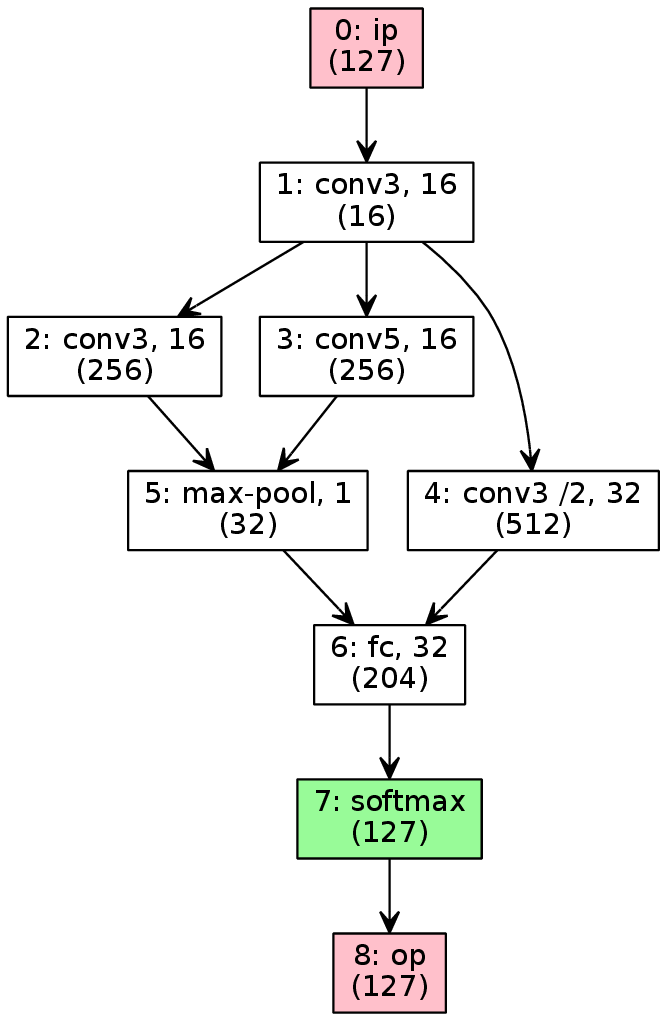} \mlpfighsp
\label{fig:mlpeg3}}\mlpfighsp
\subfloat[]{
\includegraphics[height=\mlpfigheight]{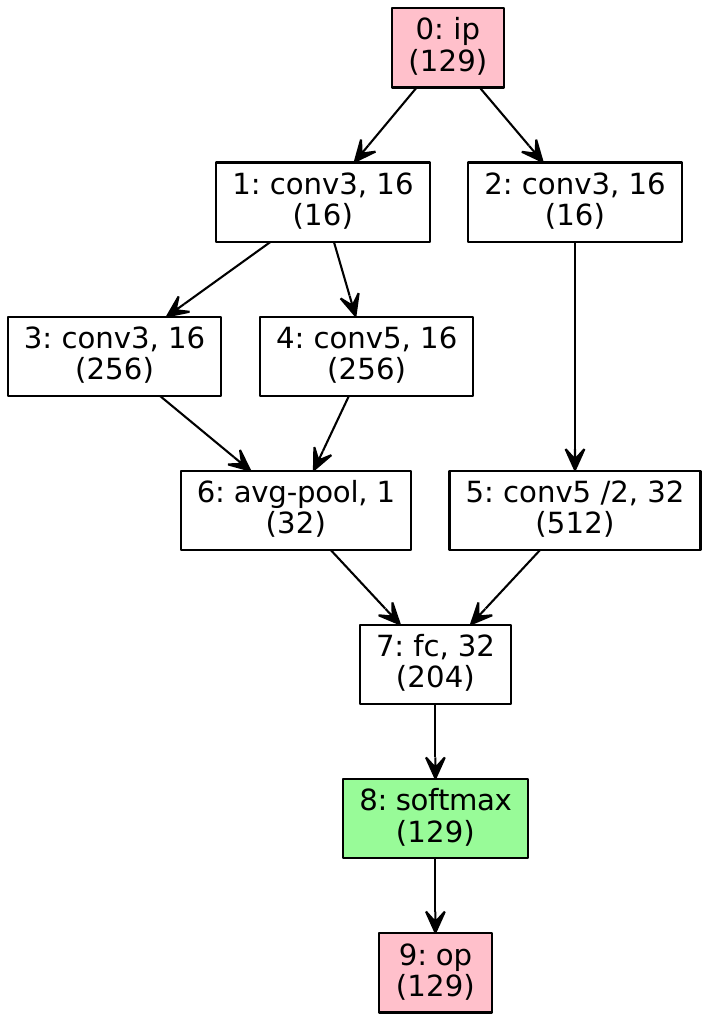}
\label{fig:mlpeg4}}\mlpfighsp
\subfloat[]{
\includegraphics[height=\mlpfigheight]{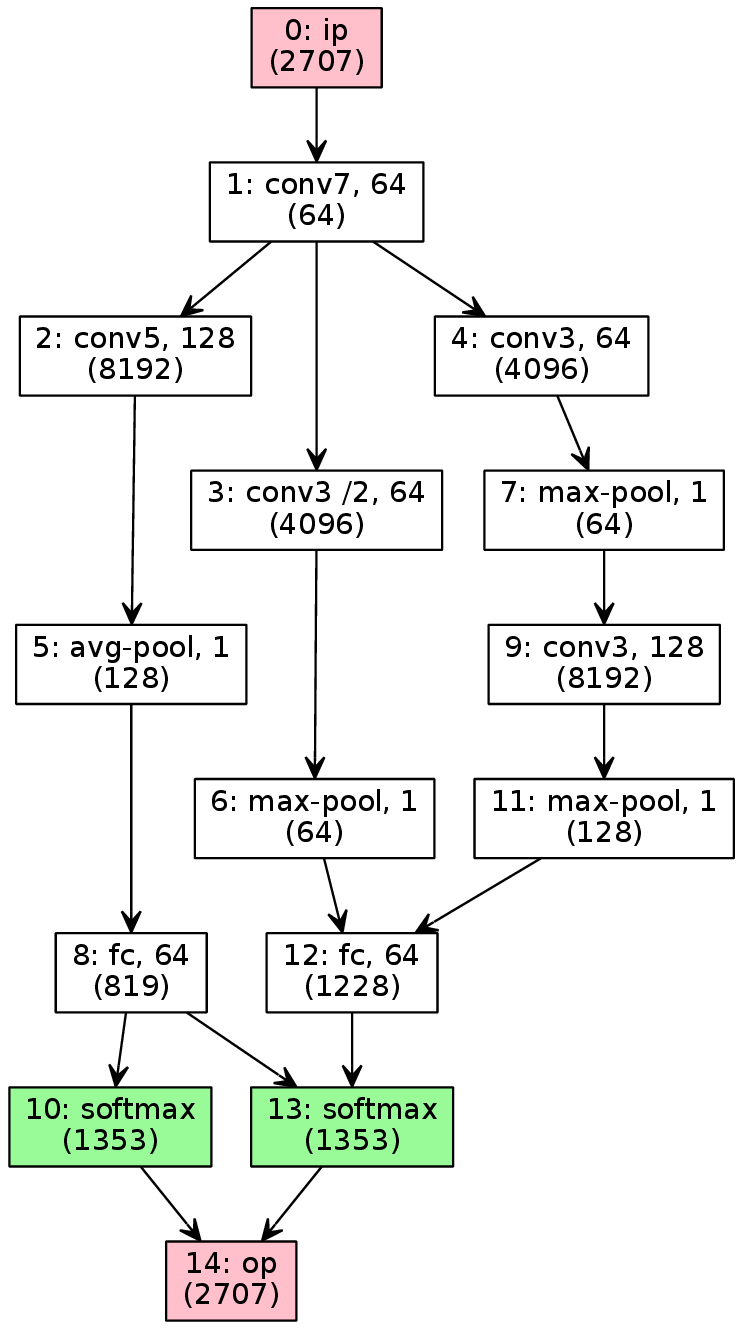}
\label{fig:mlpeg5}}\mlpfighsp
\vspace{-0.10in}
\caption{
An illustration of some CNN architectures.
In each layer, $i$: indexes the layer, followed by the label (e.g \convthree),
and followed by the number of units (e.g. number of convolutional filters).
`\inlabelfont{/2}' indicates stride 2 when performing convolutions.
The layer mass is denoted in parantheses.
The input and output layers are shaded in pink while the decision (\softmax) layers
are shaded in green.
\label{fig:mlpegs}
}
\end{center}
\end{figure*}
}

\newcommand{\insertFigPseudoDistanceIllus}{
\newcommand{\nnfigpdwidth}{1.1in}
\newcommand{\nnfigpdheight}{2.15in}
\newcommand{\nnfigpdhsp}{\hspace{0.250in}}
\begin{figure}
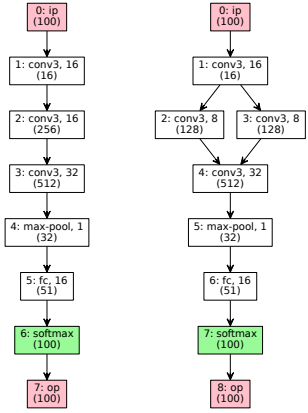

\begin{minipage}{2.5in}
\begin{center}
\centering
\subfloat{
\includegraphics[height=\nnfigpdheight]{figs/cnn_egs_2/1}
\label{fig:nneg1}}\nnfigpdhsp
\subfloat{
\includegraphics[height=\nnfigpdheight]{figs/cnn_egs_2/0}
\label{fig:nneg2}}
\end{center}
\end{minipage}
\hspace{0.01in}
\begin{minipage}{2.7in}
\caption{
An example of $2$ CNNs which have $d=\dbar=0$ distance.
The OT solution matches the mass in each layer in the network on the left to the layer
horizontally opposite to it on the right with $0$ cost.
For layer 2 on the left, its mass is mapped to layers 2 and 3 on the left.
However,
while the descriptor of these networks is different,
their functional behaviour is the same.
\label{fig:nnpd}
\vspace{-0.2in}
}
\end{minipage}
\end{figure}
}

\newcommand{\insertFigNNEgsMain}{
\newcommand{\mainnnfigwidth}{1.1in}
\newcommand{\mainnnfigheight}{2.25in}
\newcommand{\mainnnfighsp}{\hspace{0.000in}}
\begin{figure}
\begin{minipage}{3.0in}
\vspace{-0.15in}
\begin{center}
\subfloat[]{
\includegraphics[height=\mainnnfigheight]{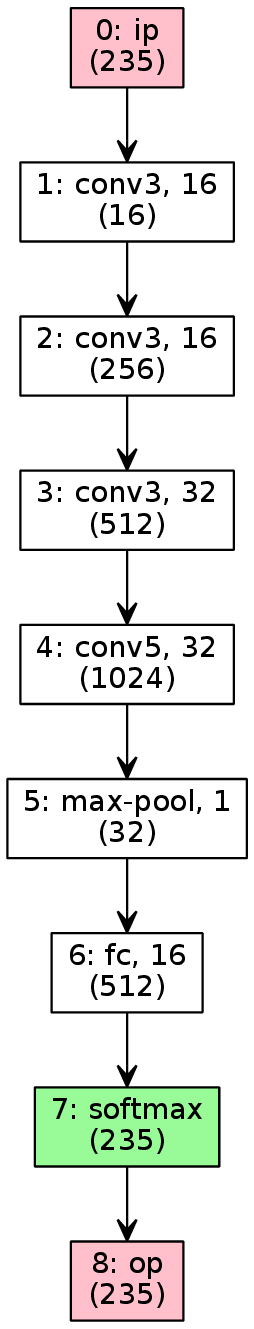}
\label{fig:mainnneg1}}\mainnnfighsp
\subfloat[]{
\includegraphics[height=\mainnnfigheight]{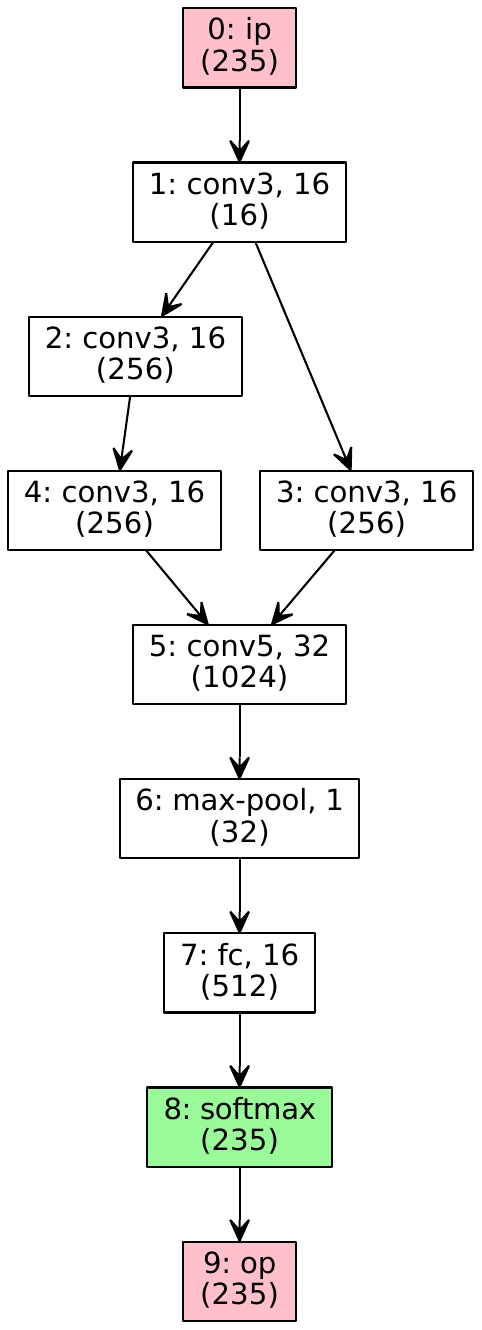}
\label{fig:mainnneg2}}\mainnnfighsp
\subfloat[]{
\includegraphics[height=\mainnnfigheight]{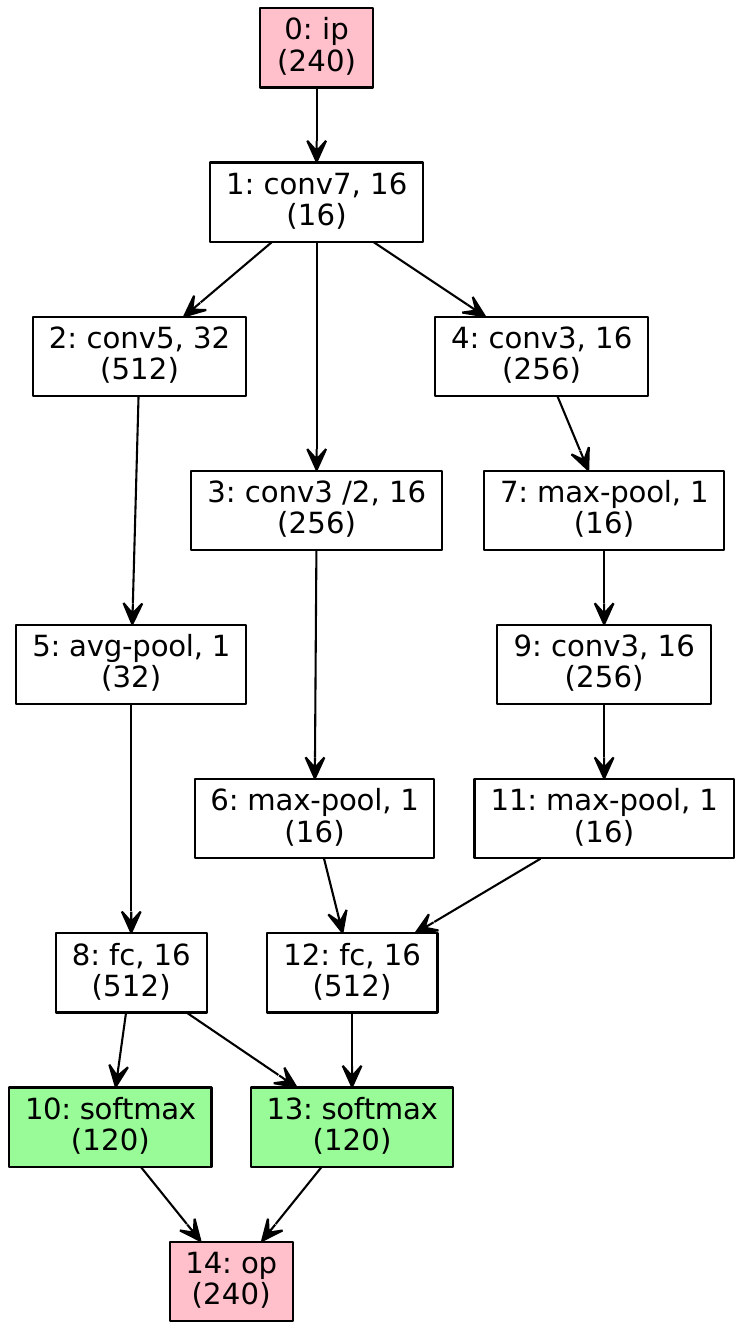} 
\label{fig:mainnneg3}}
\end{center}
\end{minipage}
\hspace{-0.00in}
\begin{minipage}{2.2in}
\caption{
An illustration of some CNN architectures.
In each layer, $i$: indexes the layer, followed by the label (e.g \convthree),
and then the number of units
(e.g. number of filters).
The input and output layers are pink while the decision (\softmax) layers
are green. \\[0.05in]
\emph{From Section~\ref{sec:nndistmain}:}
The layer mass is denoted in parentheses.
The following are the normalised and unnormalised distances $d, \dbar$ .
All self distances are $0$, i.e.
$d(\Gcal,\Gcal) = $ $\dbar(\Gcal,\Gcal) = 0$.
Unnormalised:
$d(\textrm{a},\textrm{b}) = 175.1$,\;
$d(\textrm{a},\textrm{c}) = 1479.3$,\;
$d(\textrm{b},\textrm{c}) = 1621.4$.
Normalised:
$\dbar(\textrm{a},\textrm{b}) = 0.0286$,\;
$\dbar(\textrm{a},\textrm{c}) = 0.2395$,\;
$\dbar(\textrm{b},\textrm{c}) = 0.2625$.%
\hspace{-0.1in}
\label{fig:mainnnegs}
}
\end{minipage}
\vspace{-0.2in}
\end{figure}
}

\newcommand{\insertFigMLP}{
\newcommand{\modselfigwidth}{1.015in}
\newcommand{\modselfighsp}{\hspace{-0.065in}}
\newcommand{\modselfighsptwo}{\hspace{0.15in}}
\begin{figure*}[t]
\subfloat{
\includegraphics[height=\modselfigwidth]{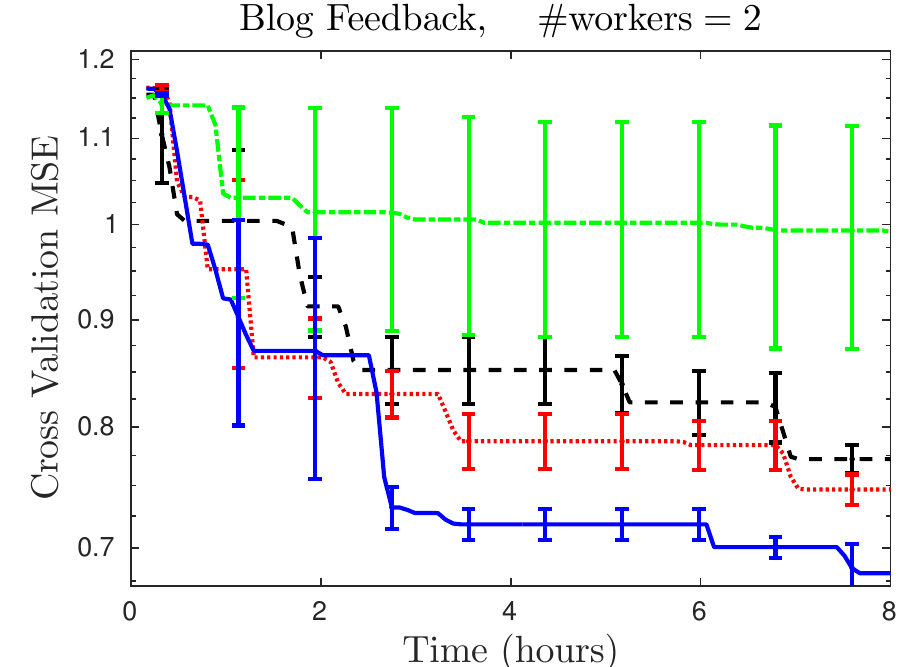}
\label{fig:modselblog}} \modselfighsp
\subfloat{
\includegraphics[height=\modselfigwidth]{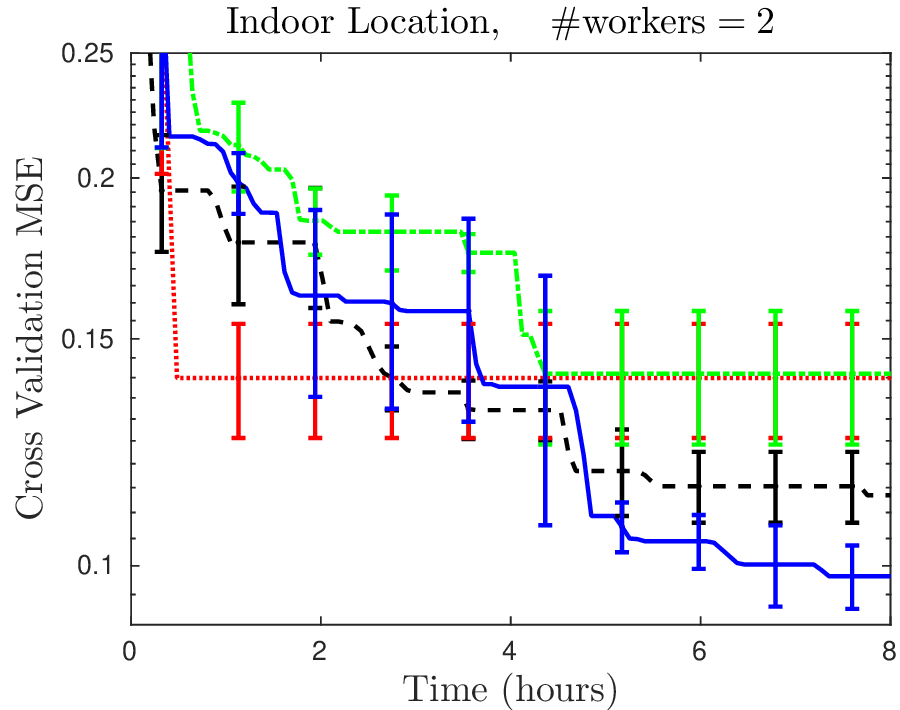}
\label{fig:modselindoor}} \modselfighsp
\subfloat{
\includegraphics[height=\modselfigwidth]{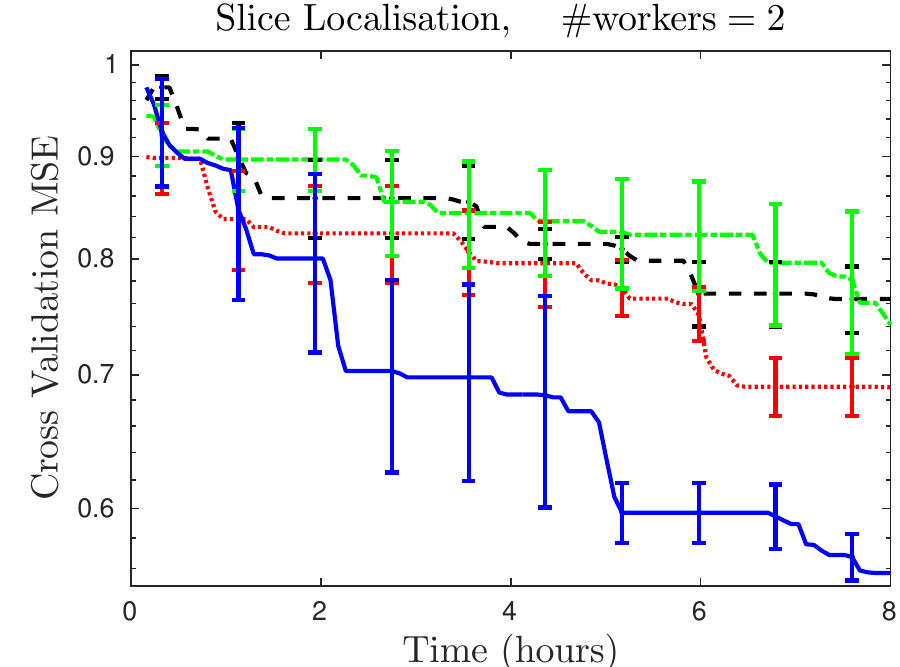}
\label{fig:modselslice}} \modselfighsp
\subfloat{
\includegraphics[height=\modselfigwidth]{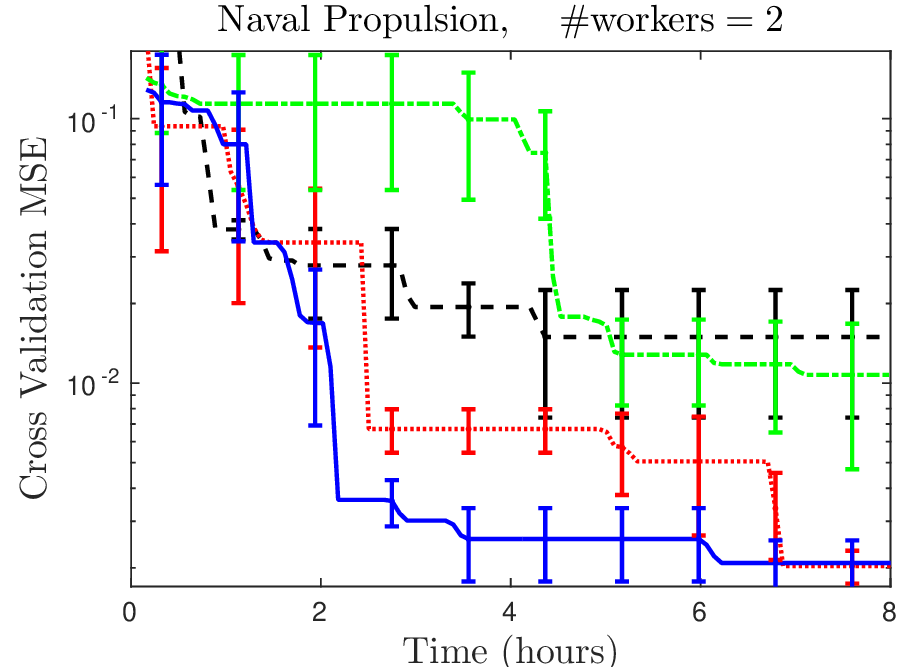}
\label{fig:modselnaval}} \modselfighsp
\\[-0.07in]
\subfloat{
\includegraphics[height=\modselfigwidth]{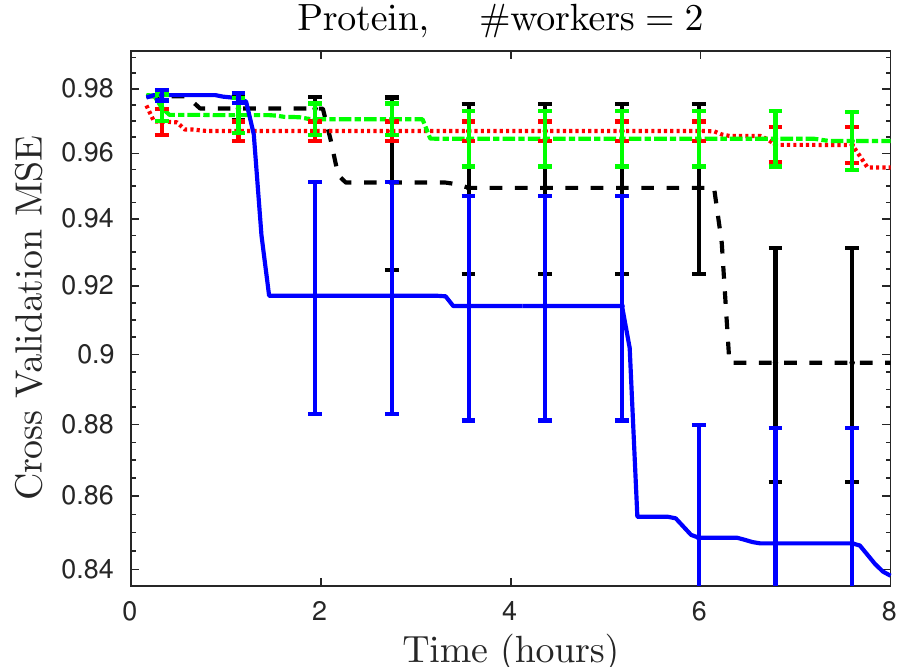}
\label{fig:modselprotein}} \modselfighsp
\subfloat{
\includegraphics[height=\modselfigwidth]{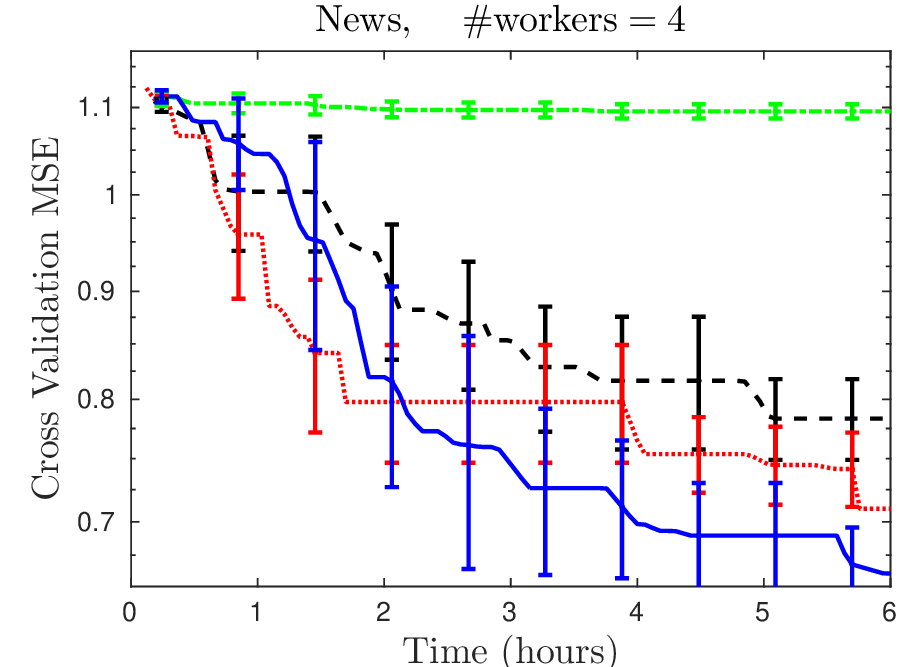}
\label{fig:modselnews}} \modselfighsp
\subfloat{
\includegraphics[height=\modselfigwidth]{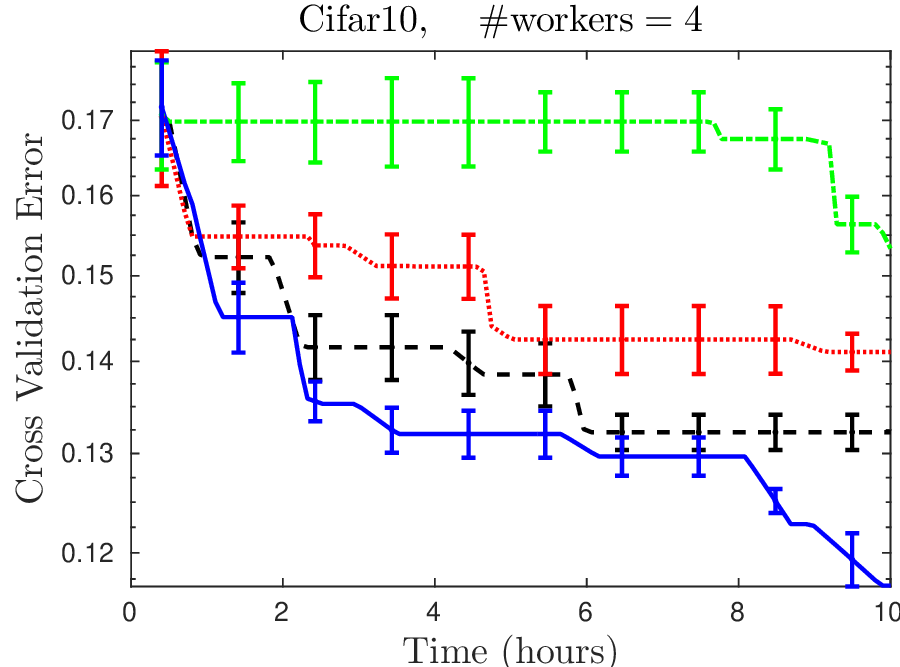}
\label{fig:modselcifar}} \modselfighsp
\hspace{0.35in}
\subfloat{
\includegraphics[width=0.7in]{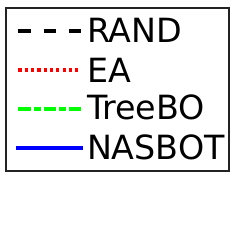}
\label{fig:modselcifar}
} 
\vspace{-0.02in}
\caption{\small
\emph{Cross validation results:}
In all figures, the $x$ axis is time.
The $y$ axis is the mean squared error (MSE) in the first 6 figures and
the classification error in the last. Lower is better in all cases.
The title of each figure states the dataset and the number of parallel workers (GPUs).
All figures were averaged over at least $5$ independent runs of each method.
Error bars indicate one standard error. \hspace{-0.1in}
\label{fig:modsel}
\vspace{-0.15in}
}
\end{figure*}
}

\newcommand{\insertdistembedding}{
\newcommand{\embedwidth}{5.5in}
\begin{figure}
\begin{center}
\subfloat{
\includegraphics[width=\embedwidth]{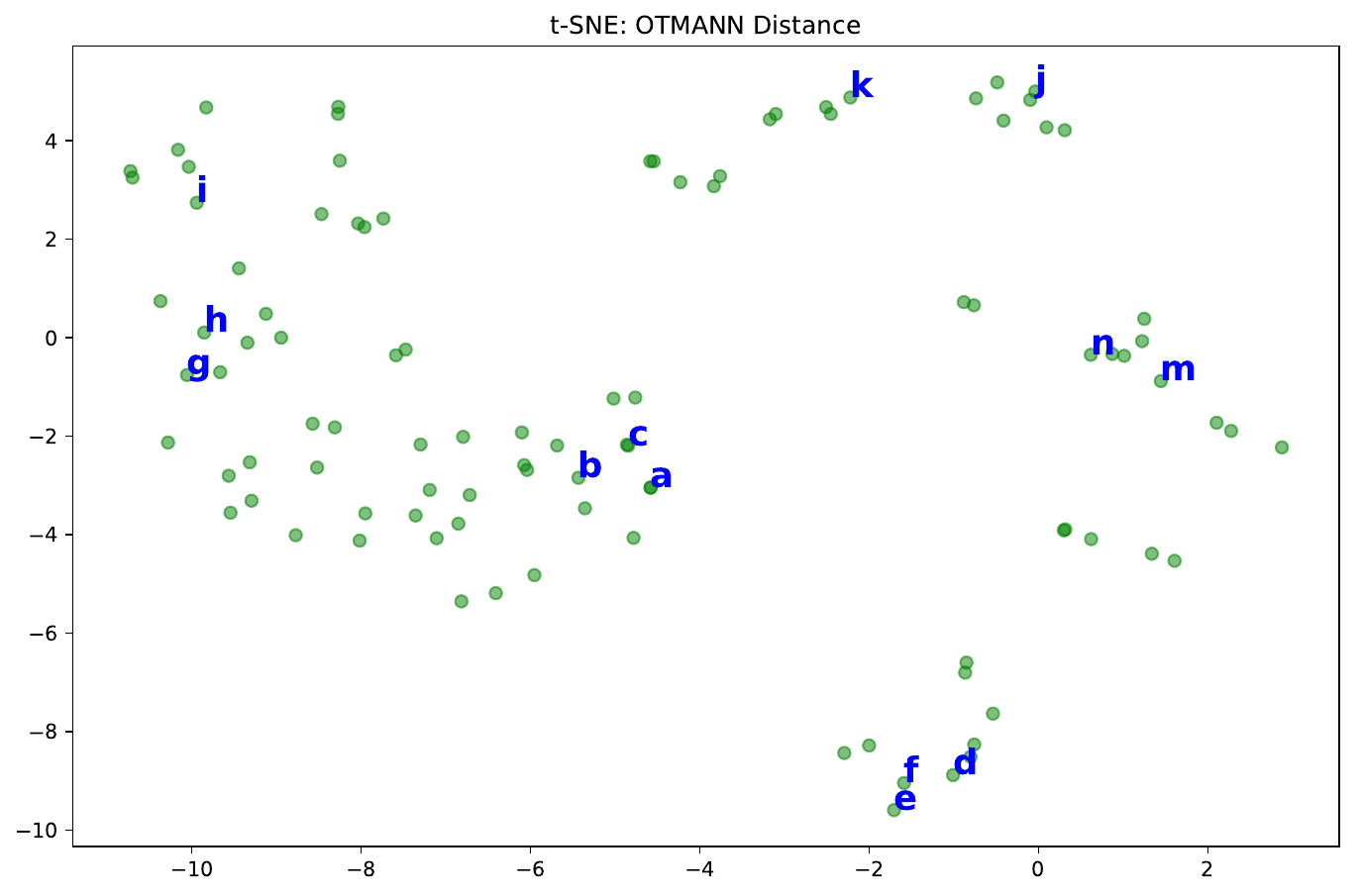}
\label{fig:tsne_unnorm}} \\
\subfloat{
\includegraphics[width=\embedwidth]{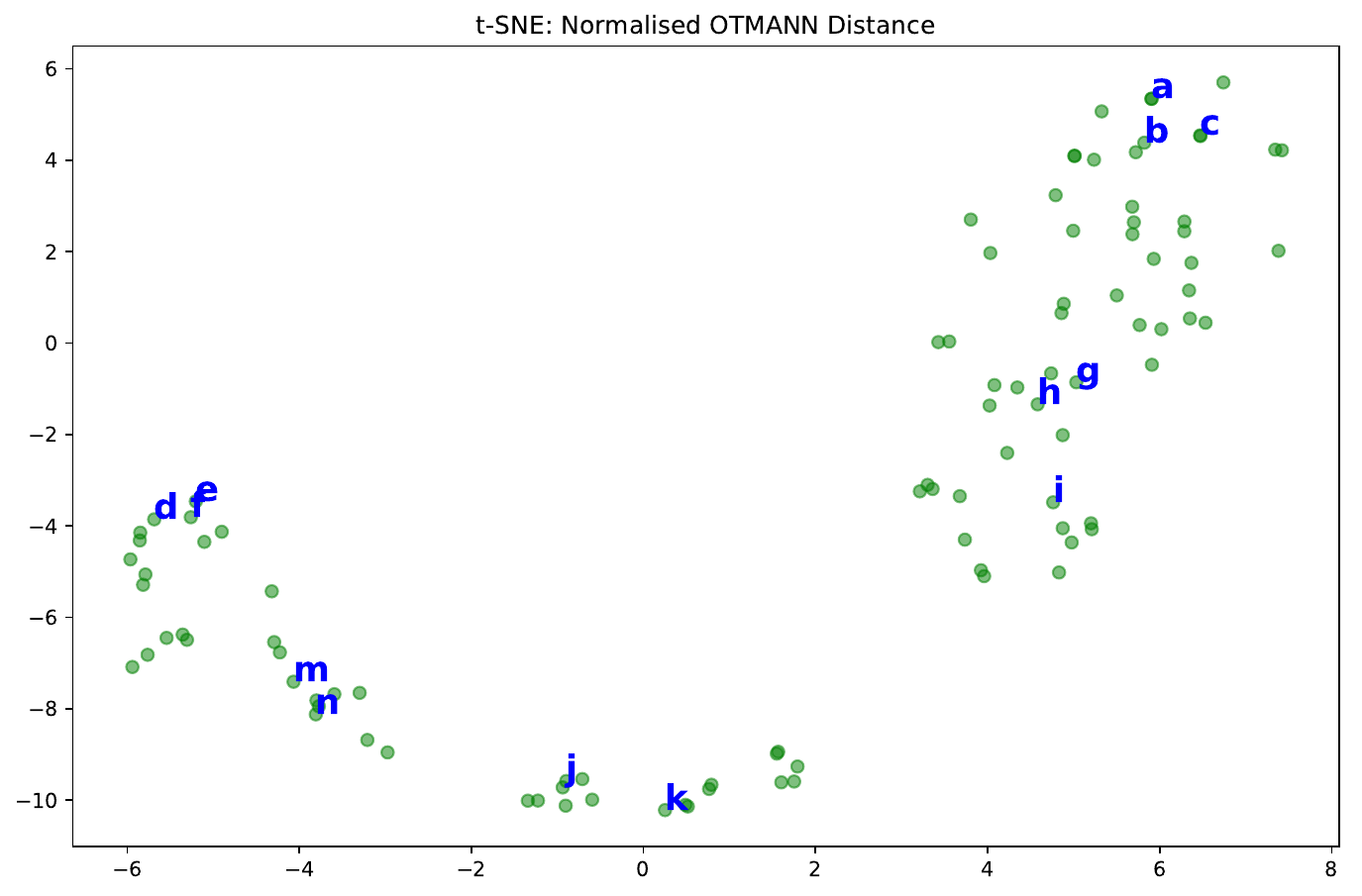}
\label{fig:tsne_norm}}
\vspace{-0.00in}
\caption{
Two dimensional
t-SNE embeddings of $100$ randomly generated CNN architectures based on
the \nndists distance (top) and its normalised version (bottom).
Some networks have been indexed a-n in the figures; these network architectures are
illustrated in Figure~\ref{fig:tsne_nns}.
Networks that are similar are embedded close to each other indicating that
the \nndists induces a meaningful topology among neural network architectures.
\label{fig:tsne}
}
\end{center}
\end{figure}
}

\newcommand{\insertdistembeddingnns}{
\begin{figure}
\begin{center}
\includegraphics[width=5.55in]{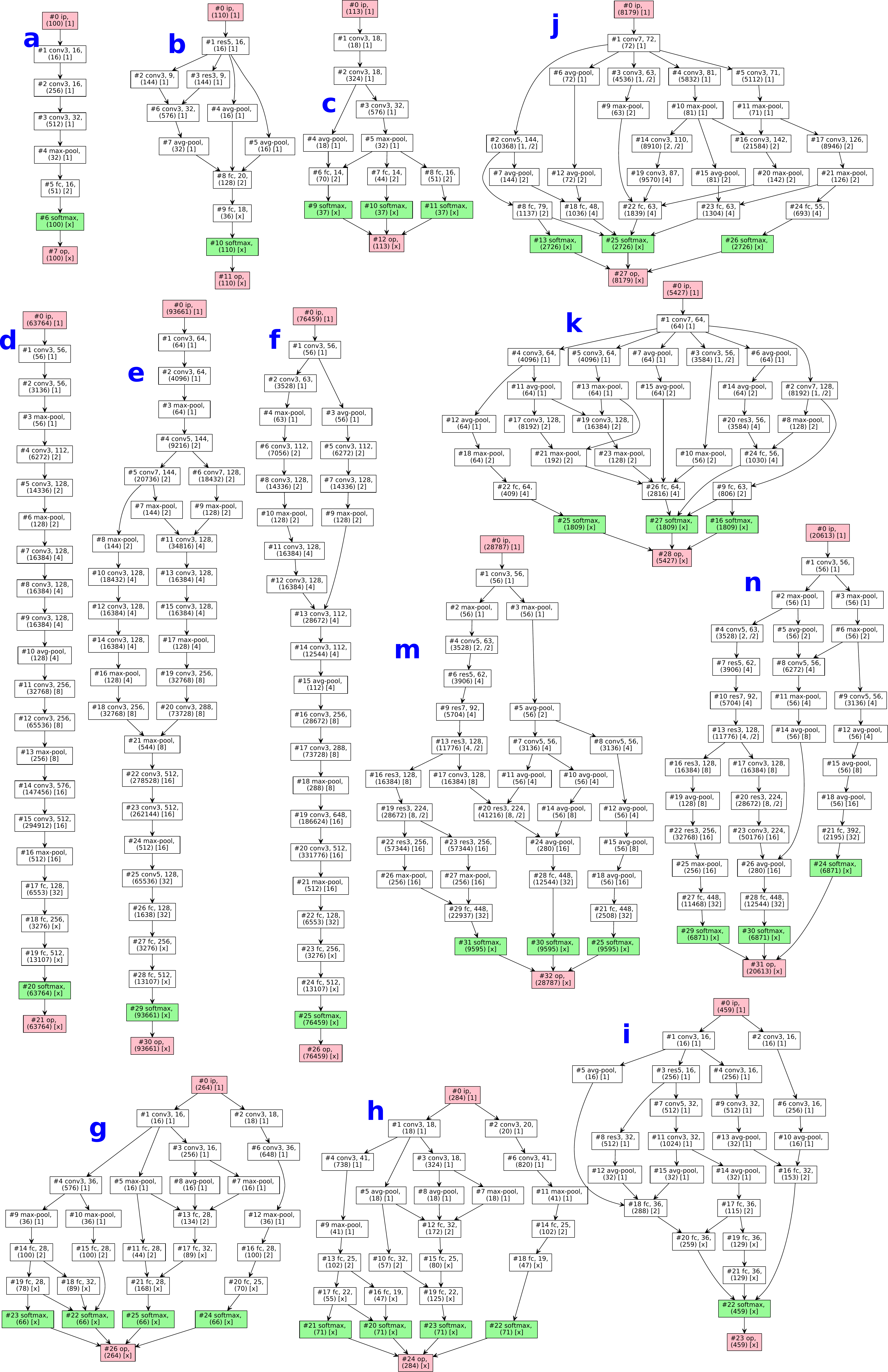}
\vspace{-0.02in}
\caption{
Illustrations of the nextworks indexed a-n in Figure~\ref{fig:tsne}.
\label{fig:tsne_nns}
}
\end{center}
\end{figure}
}

\newcommand{\insertdistcorrelations}{
\newcommand{\corrwidth}{1.8in}
\begin{figure}
\begin{center}
\subfloat{
\includegraphics[width=\corrwidth]{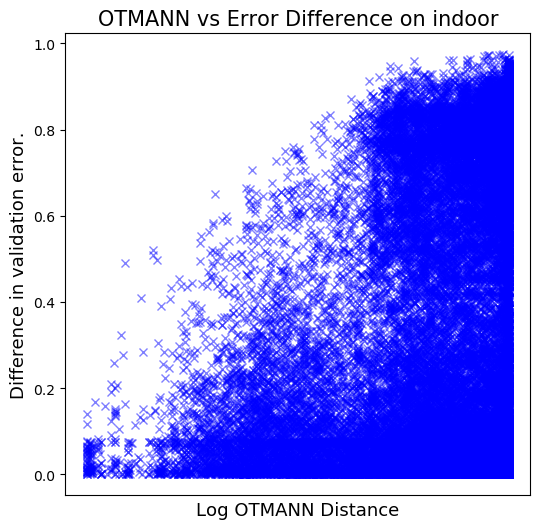}
} 
\subfloat{
\includegraphics[width=\corrwidth]{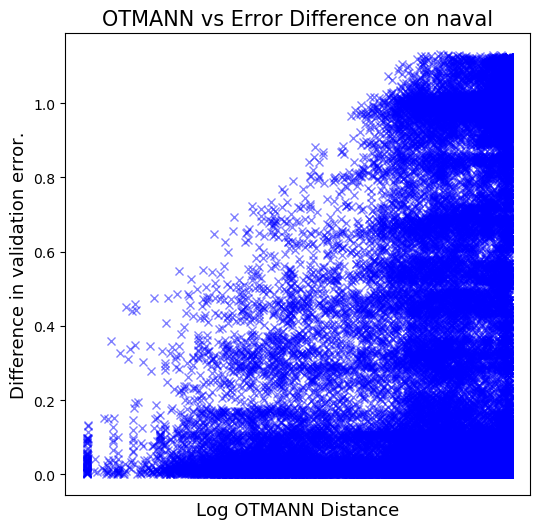}
} 
\subfloat{
\includegraphics[width=\corrwidth]{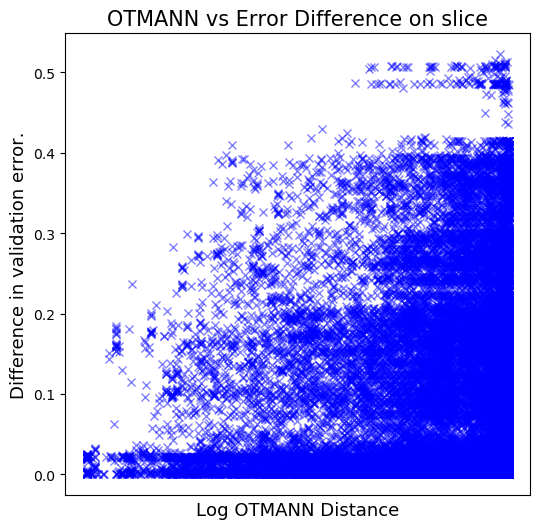}
}
\vspace{-0.00in}
\caption{
\label{fig:distcorr}
Each point in the scatter plot indicates the log distance between two architectures ($x$
axis) and the difference in the validation error ($y$ axis),
on the Indoor, Naval and Slice datasets.
We used $300$ networks, giving rise to $\sim 45K$ pairwise points.
On all datasets, when the distance is small, so is the difference in the validation error.
As the distance increases, there is more variance in the validation error difference.
Intuitively, one should expect that while networks that are far apart could perform
similarly or differently, networks with small distance should perform similarly.
}
\end{center}
\end{figure}
}

\newcommand{\insertFigAblation}{
\newcommand{\ablationfigwidth}{1.3in}
\newcommand{\ablationfighsp}{\hspace{-0.065in}}
\newcommand{\ablationfighsptwo}{\hspace{0.15in}}
\begin{figure*}[t]
\centering
\subfloat[]{
\includegraphics[height=\ablationfigwidth]{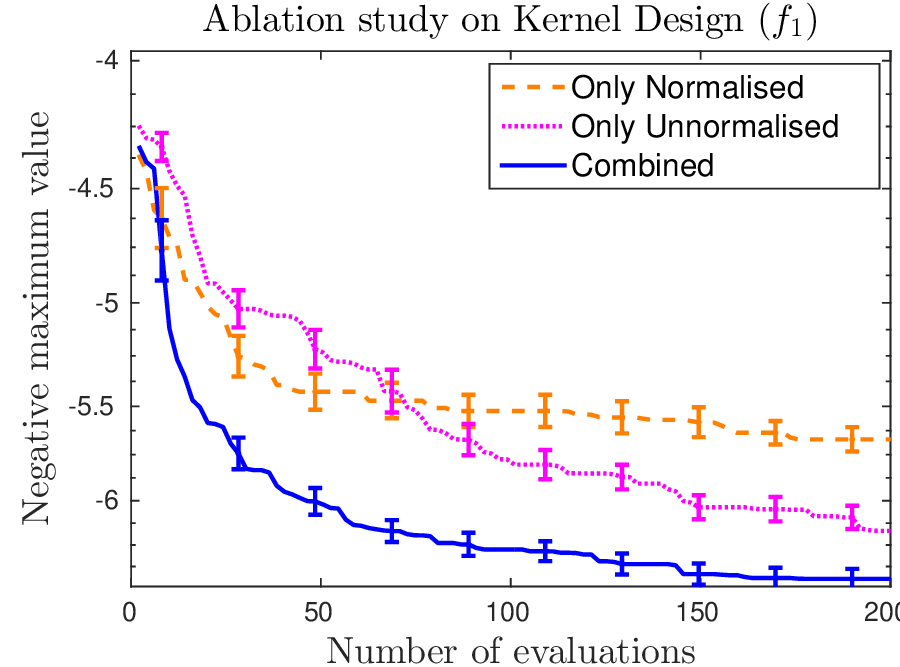}
\label{fig:ablkernel}} \ablationfighsp
\subfloat[]{
\includegraphics[height=\ablationfigwidth]{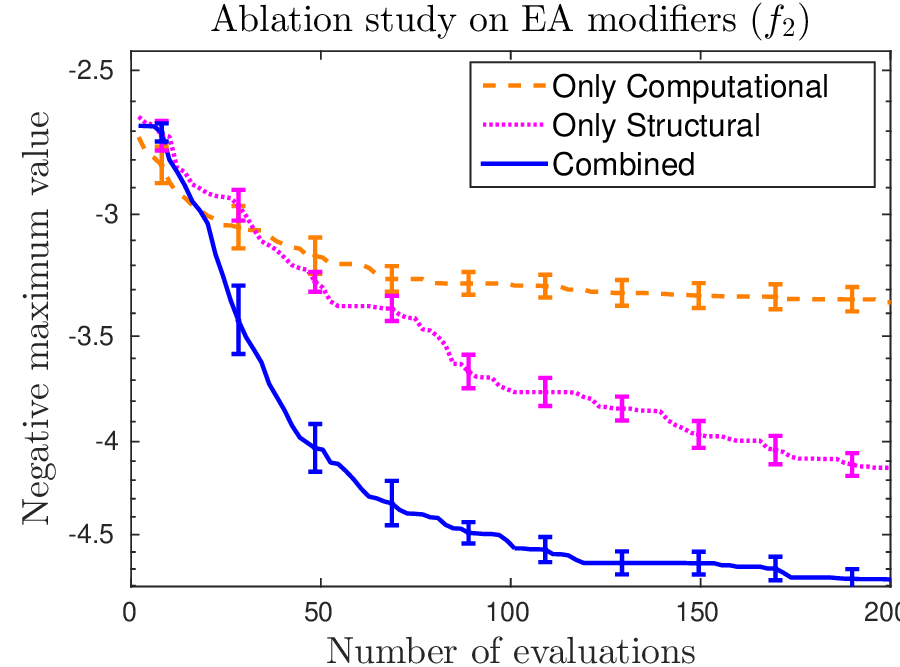}
\label{fig:ablea}} \ablationfighsp
\subfloat[]{
\includegraphics[height=\ablationfigwidth]{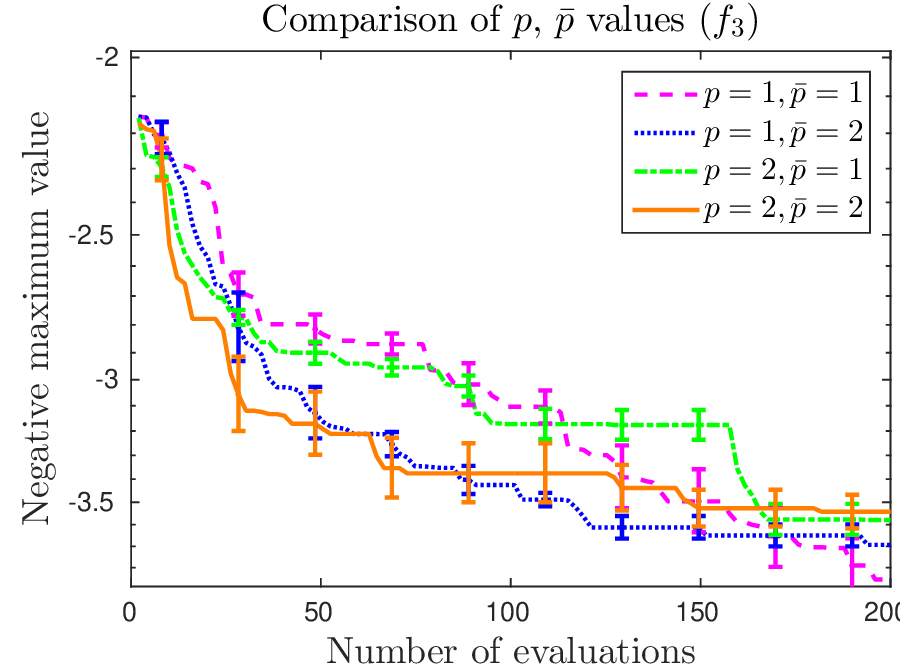}
\label{fig:ablppbar}}
\caption{\small
We compare \nnbos for different design choices in our framework.
\protect\subref{fig:ablkernel}:
Comparison of \nnbos using only the normalised distance 
$e^{-\beta \dbar}$,
only the unnormalised distance $d^{-\beta d}$, and the combination 
$e^{-\beta d} + e^{-\betabar \dbar}$.
\protect\subref{fig:ablea}:
Comparison of \nnbos using only the \evoalgs modifiers which change the
computational units (top 4 in
Table~\ref{tb:nnmodifiers}),
modifiers which only change the structure of the networks (bottom 5 in
Table~\ref{tb:nnmodifiers}),
and all 9 modifiers.
\protect\subref{fig:ablppbar}:
Comparison of \nnbos with different choices for $p$ and $\pbar$.
In all figures, the $x$ axis is the number of evaluations and the $y$ axis
is the negative maximum value (lower is better).
All figures were produced by averaging over at least 10 runs.
\label{fig:ablation}
\vspace{-0.15in}
}
\end{figure*}
}

\newcommand{\fullPageFigHeight}{7in}
\newcommand{\halfPageFigHeight}{3.5in}
\newcommand{\halfPageFigWidth}{6in}
\newcommand{\initPoolFigHeight}{8in}


\newcommand{\insertFigBestNetCifarHEI}{
\begin{figure*}
\centering
\includegraphics[height=6.5in]{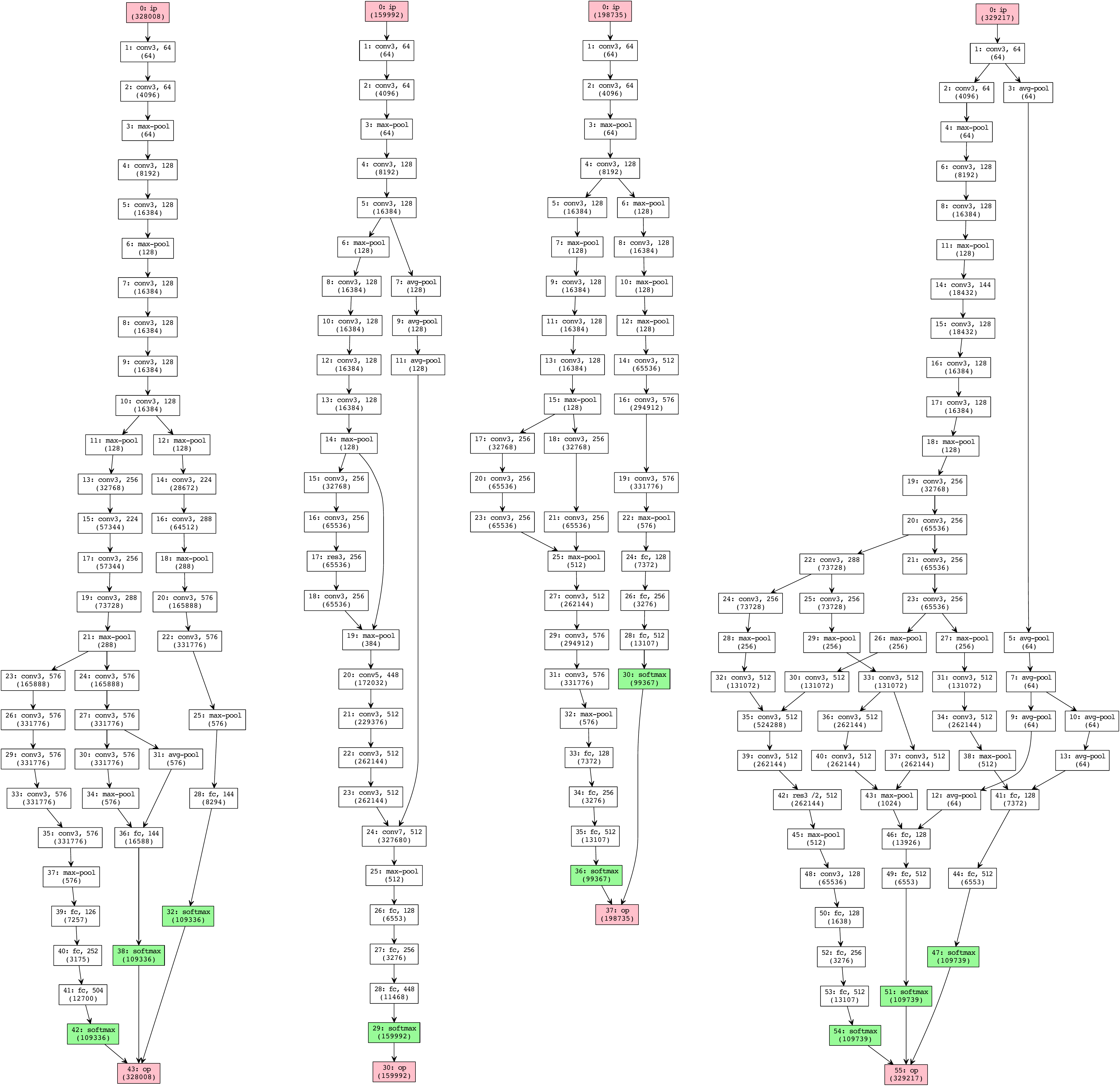}
\vspace{-0.1in}
\caption{\small
\label{fig:bestNetCifarHEI}
{Optimal network architectures found with \nnbos on Cifar10 data.}
\vspace{-0.15in}
}
\end{figure*}
}

\newcommand{\insertFigBestNetCifarGA}{
\begin{figure*}
\centering
\includegraphics[height=\fullPageFigHeight]{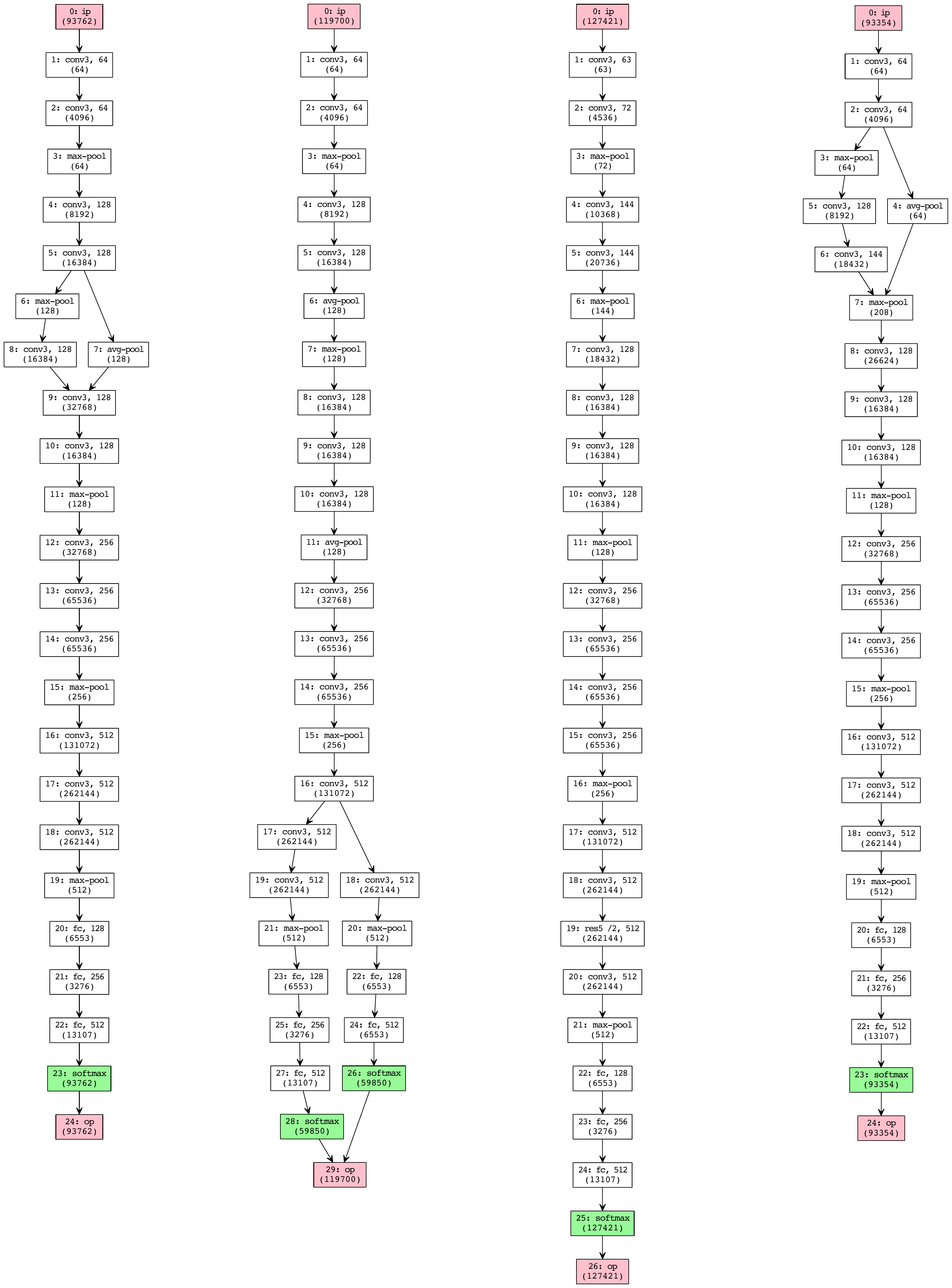}
\vspace{-0.1in}
\caption{\small
\label{fig:bestNetCifarGA}
{Optimal network architectures found with \evoalgs on Cifar10 data.}
\vspace{-0.15in}
}
\end{figure*}
}

\newcommand{\insertFigBestNetCifarRAND}{
\begin{figure*}
\centering
\includegraphics[height=\fullPageFigHeight]{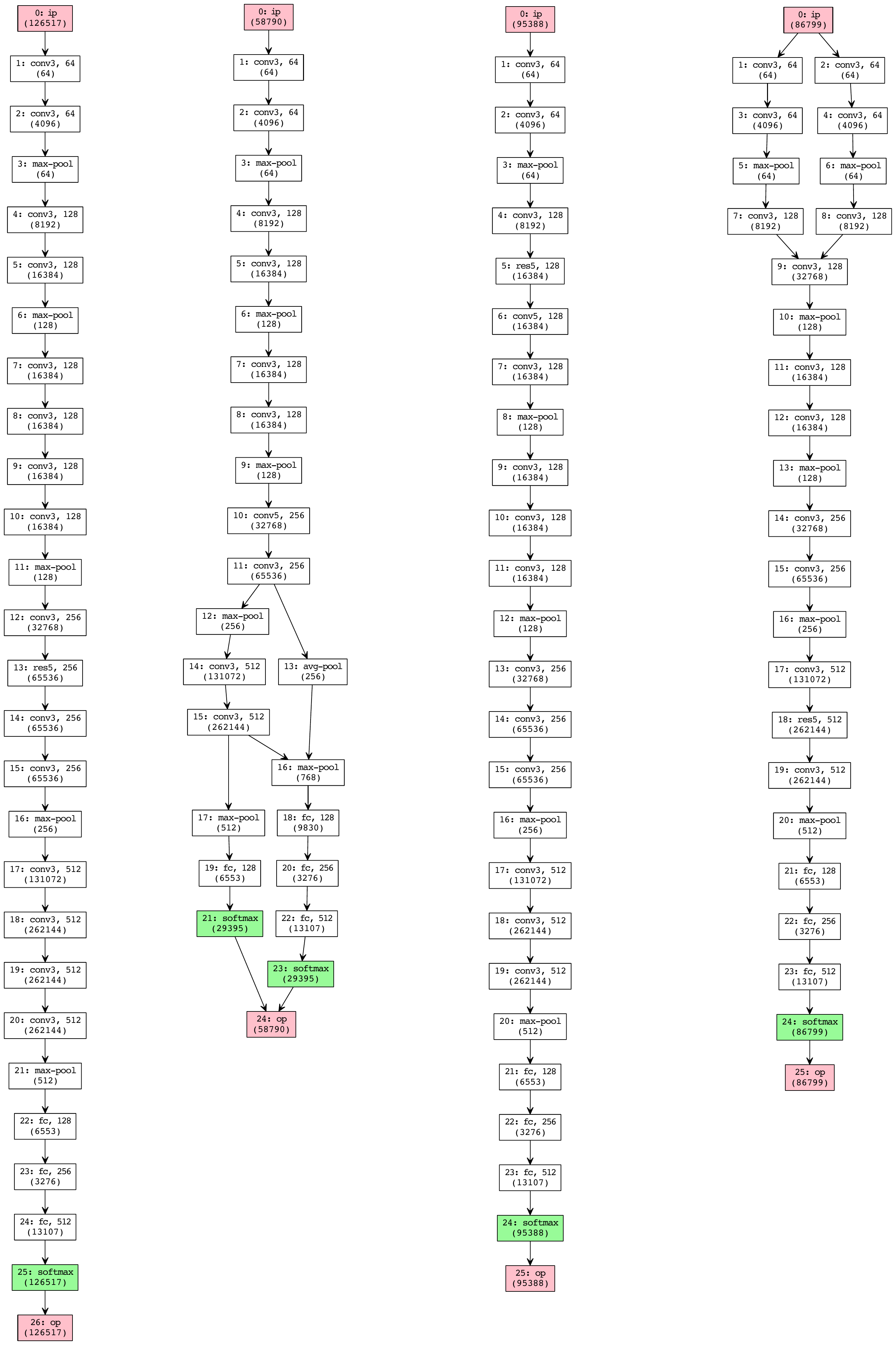}
\vspace{-0.1in}
\caption{\small
\label{fig:bestNetCifarRAND}
{Optimal network architectures found with \rands on Cifar10 data.}
\vspace{-0.15in}
}
\end{figure*}
}

\newcommand{\insertFigBestNetCifarTREE}{
\begin{figure*}
\centering
\includegraphics[height=\fullPageFigHeight]{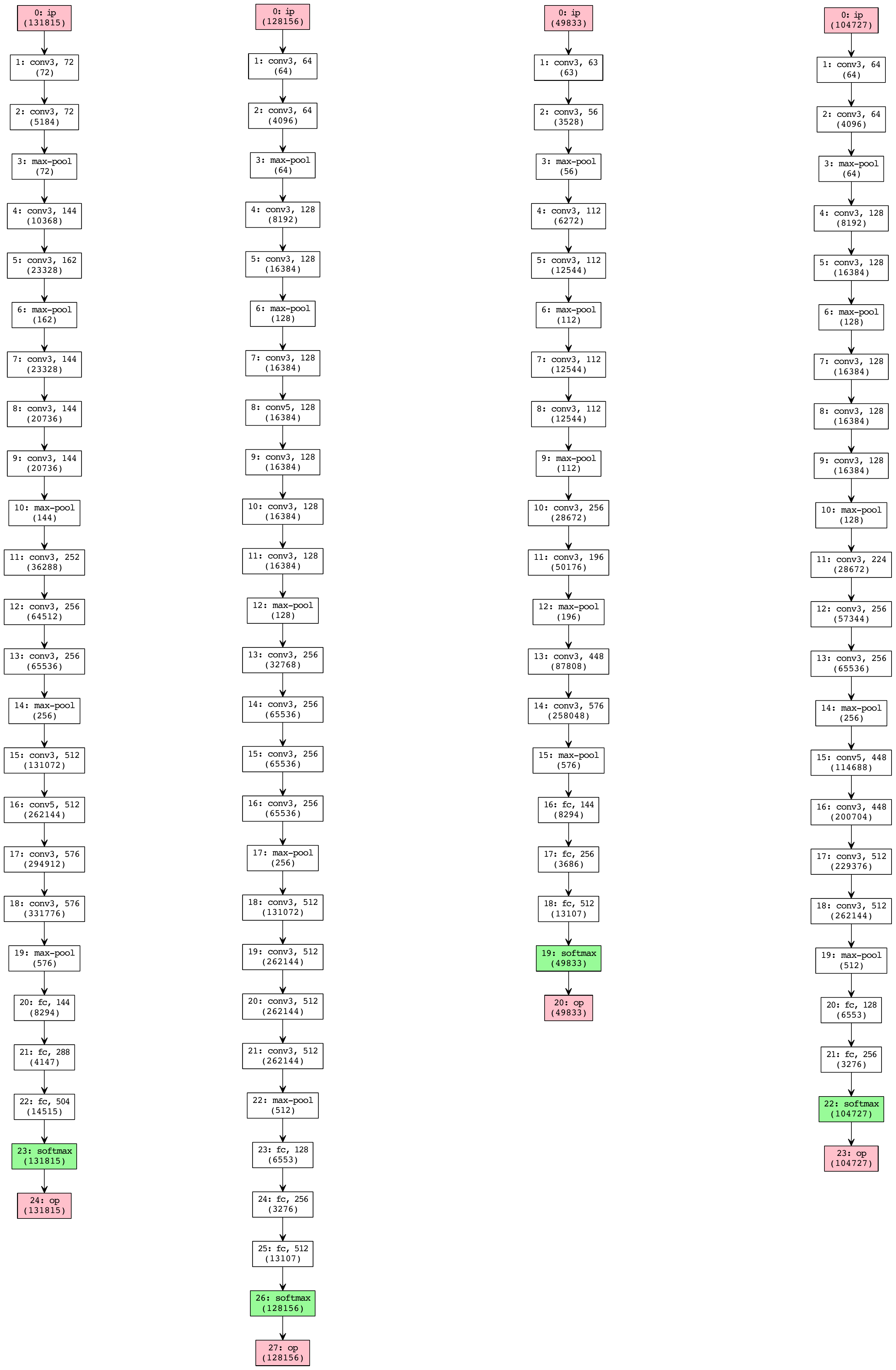}
\vspace{-0.1in}
\caption{\small
\label{fig:bestNetCifarTREE}
{Optimal network architectures found with \treebos on Cifar10 data.}
\vspace{-0.15in}
}
\end{figure*}
}


\newcommand{\insertFigBestNetIndoorHEI}{
\begin{figure*}
\centering
\includegraphics[height=\halfPageFigHeight]{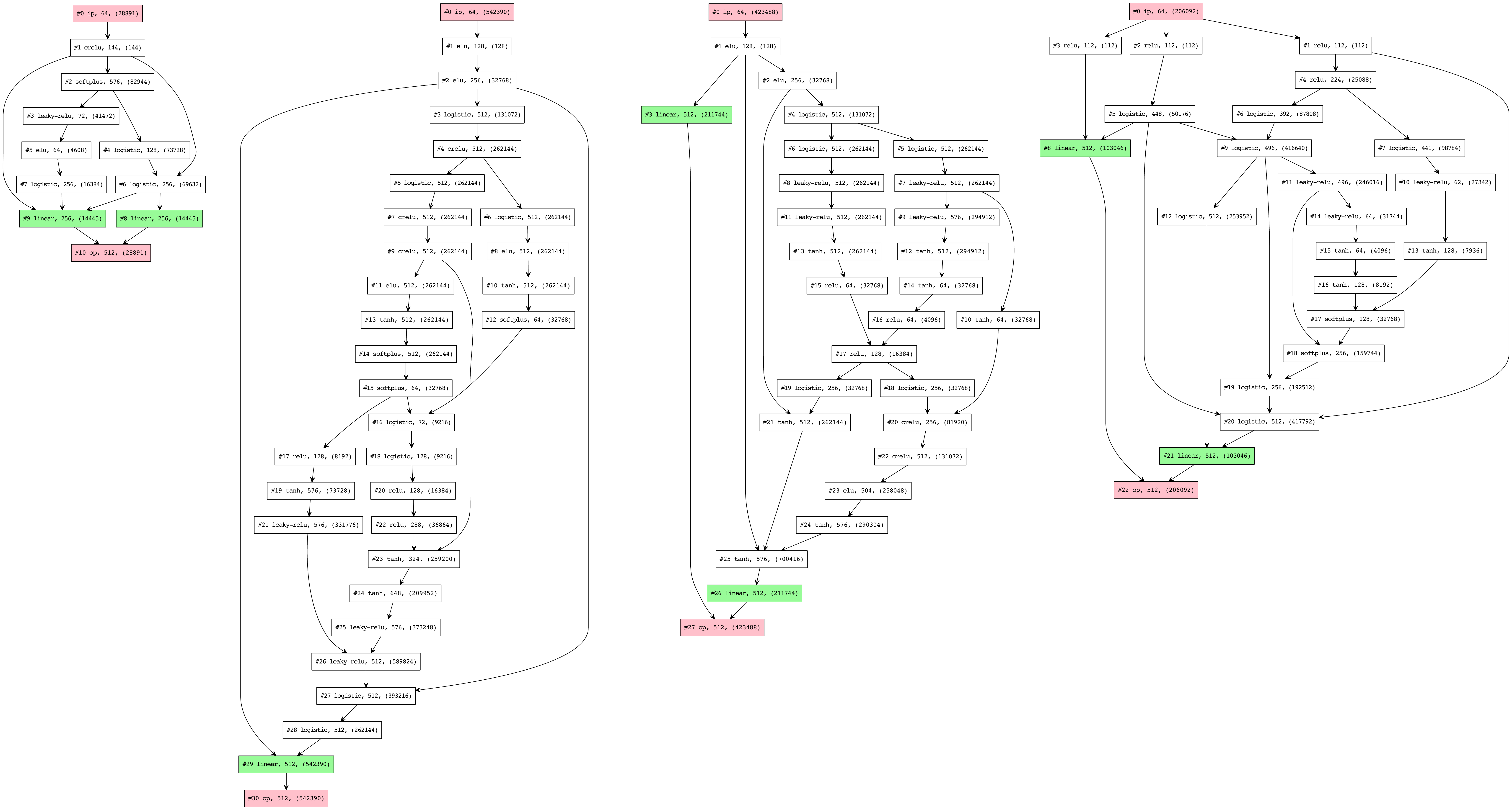}
\vspace{-0.1in}
\caption{\small
\label{fig:bestNetIndoorHEI}
{Optimal network architectures found with \nnbo on Indoor data.}
\vspace{-0.15in}
}
\end{figure*}
}

\newcommand{\insertFigBestNetIndoorGA}{
\begin{figure*}
\centering
\includegraphics[height=3.9in]{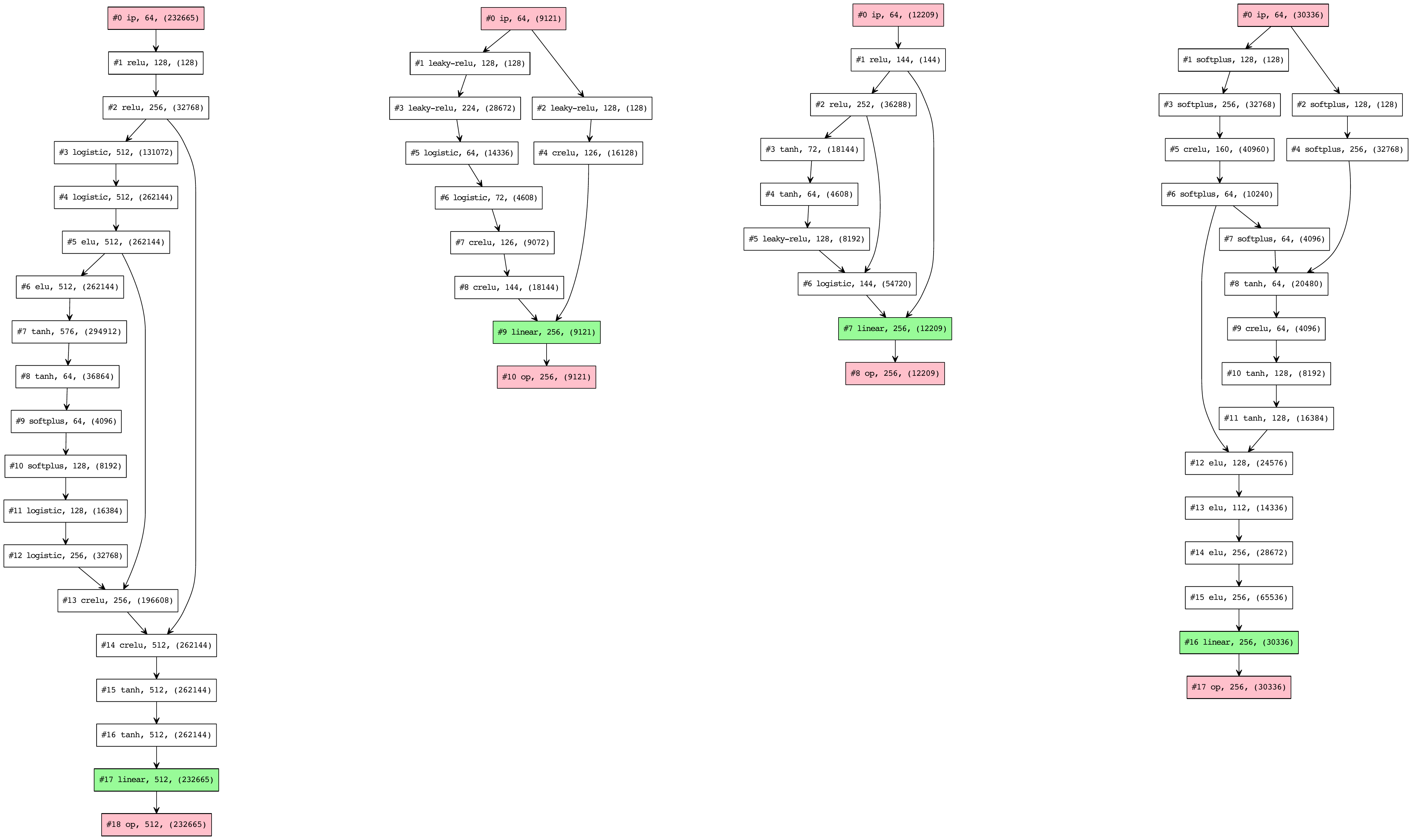}
\vspace{-0.1in}
\caption{\small
\label{fig:bestNetIndoorGA}
{Optimal network architectures found with \evoalgs on Indoor data.}
\vspace{-0.15in}
}
\end{figure*}
}

\newcommand{\insertFigBestNetIndoorRAND}{
\begin{figure*}
\centering
\includegraphics[height=3.9in]{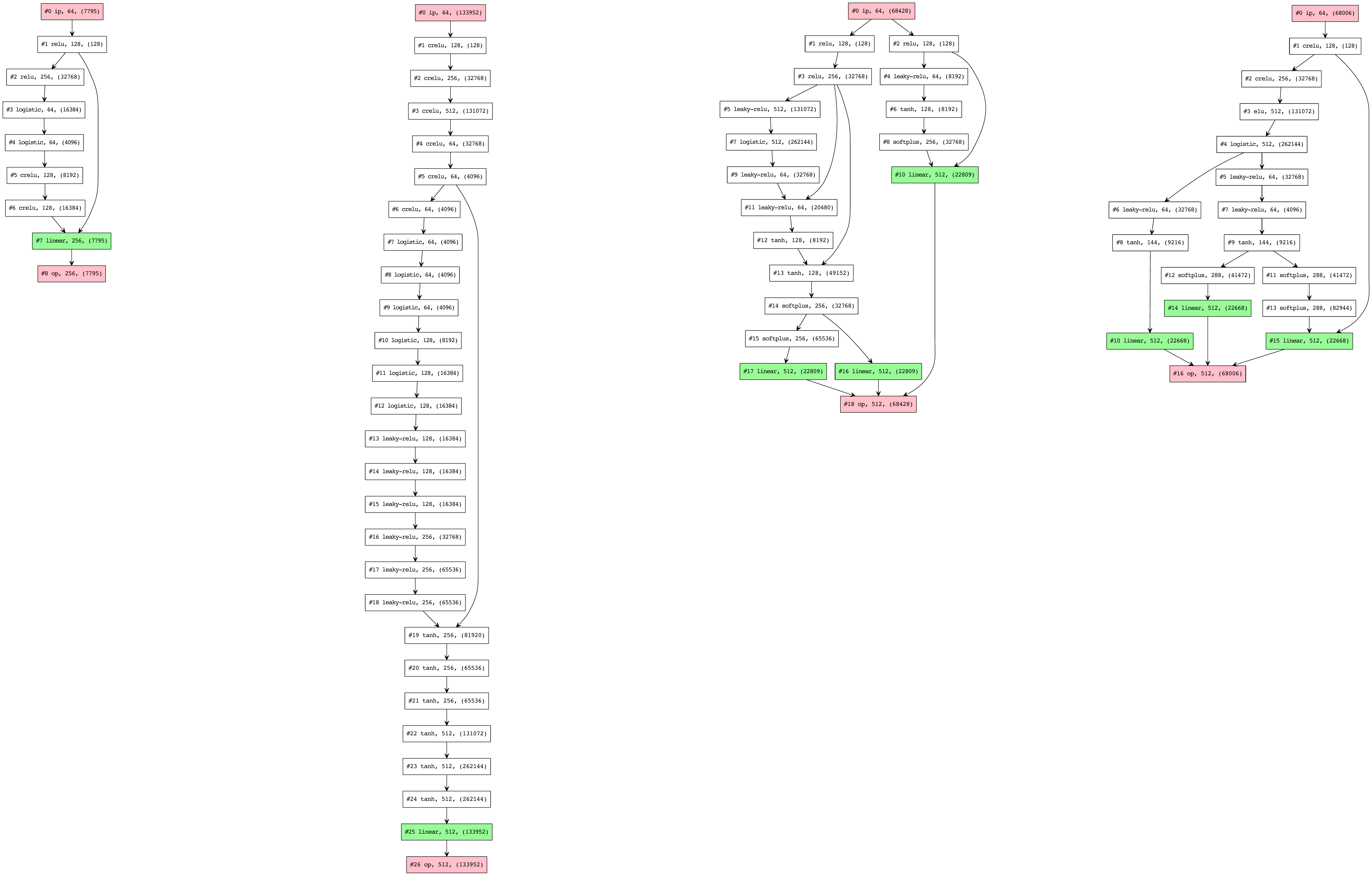}
\vspace{-0.1in}
\caption{\small
\label{fig:bestNetIndoorRAND}
{Optimal network architectures found with \rands on Indoor data.}
\vspace{-0.15in}
}
\end{figure*}
}

\newcommand{\insertFigBestNetIndoorTREE}{
\begin{figure*}
\centering
\includegraphics[width=\halfPageFigWidth]{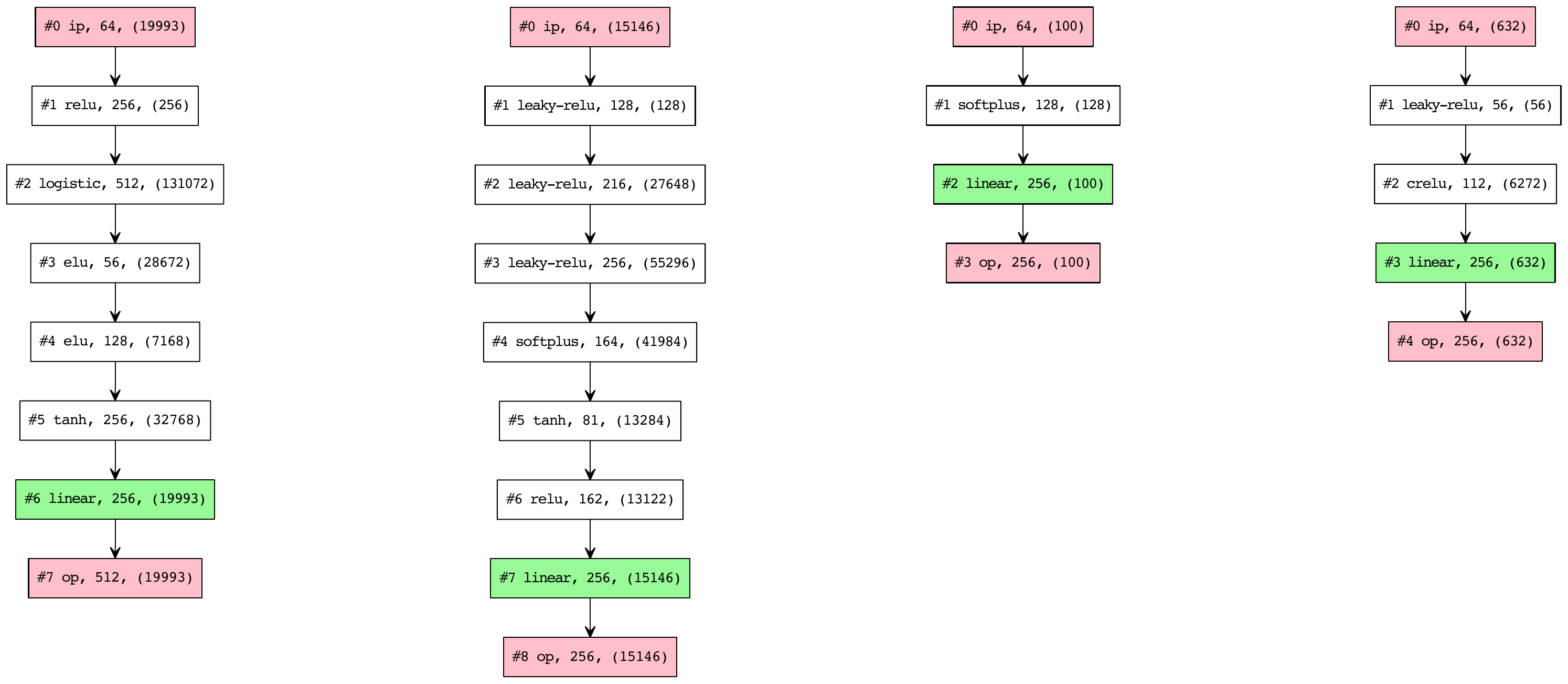}
\vspace{-0.1in}
\caption{\small
\label{fig:bestNetIndoorTREE}
{Optimal network architectures found with \treebos on Indoor data.}
\vspace{-0.15in}
}
\end{figure*}
}


\newcommand{\insertFigBestNetSliceHEI}{
\begin{figure*}
\centering
\includegraphics[height=3.9in]{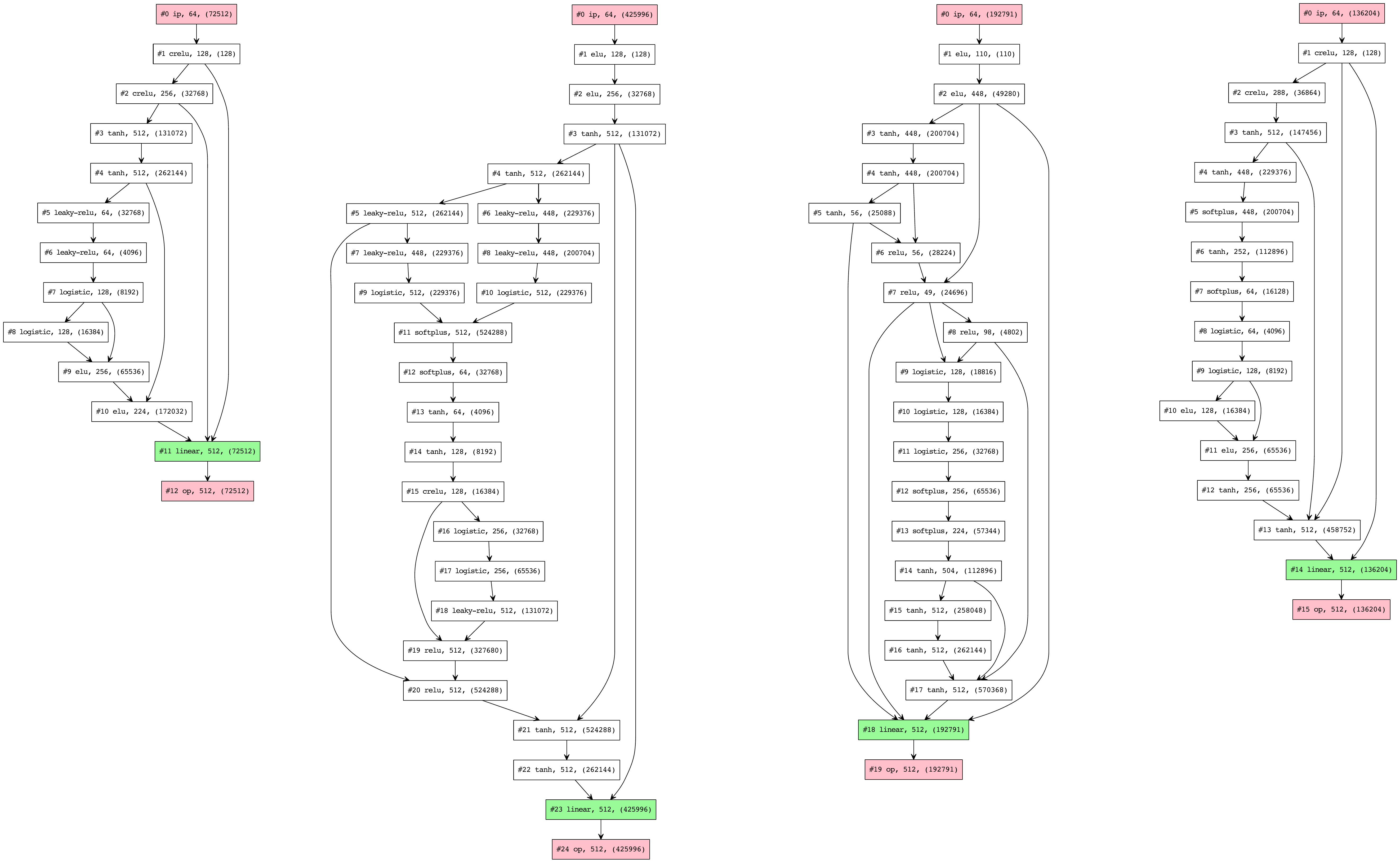}
\vspace{-0.1in}
\caption{\small
\label{fig:bestNetSliceHEI}
{Optimal network architectures found with \nnbo on Slice data.}
\vspace{-0.15in}
}
\end{figure*}
}

\newcommand{\insertFigBestNetSliceGA}{
\begin{figure*}
\centering
\includegraphics[height=3.7in]{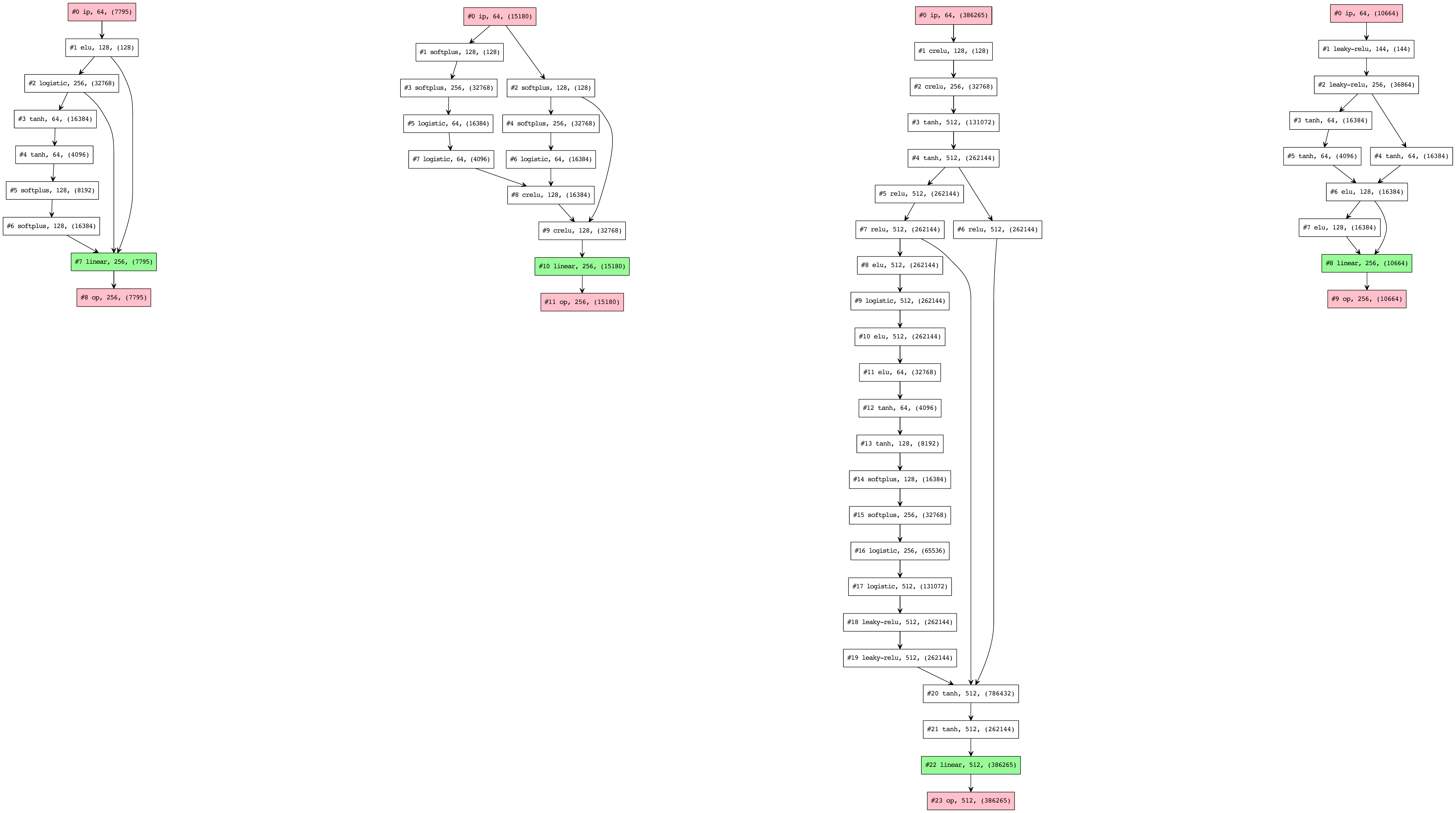}
\vspace{-0.1in}
\caption{\small
\label{fig:bestNetSliceGA}
{Optimal network architectures found with \evoalgs on Slice data.}
\vspace{-0.15in}
}
\end{figure*}
}

\newcommand{\insertFigBestNetSliceRAND}{
\begin{figure*}
\centering
\includegraphics[height=3.9in]{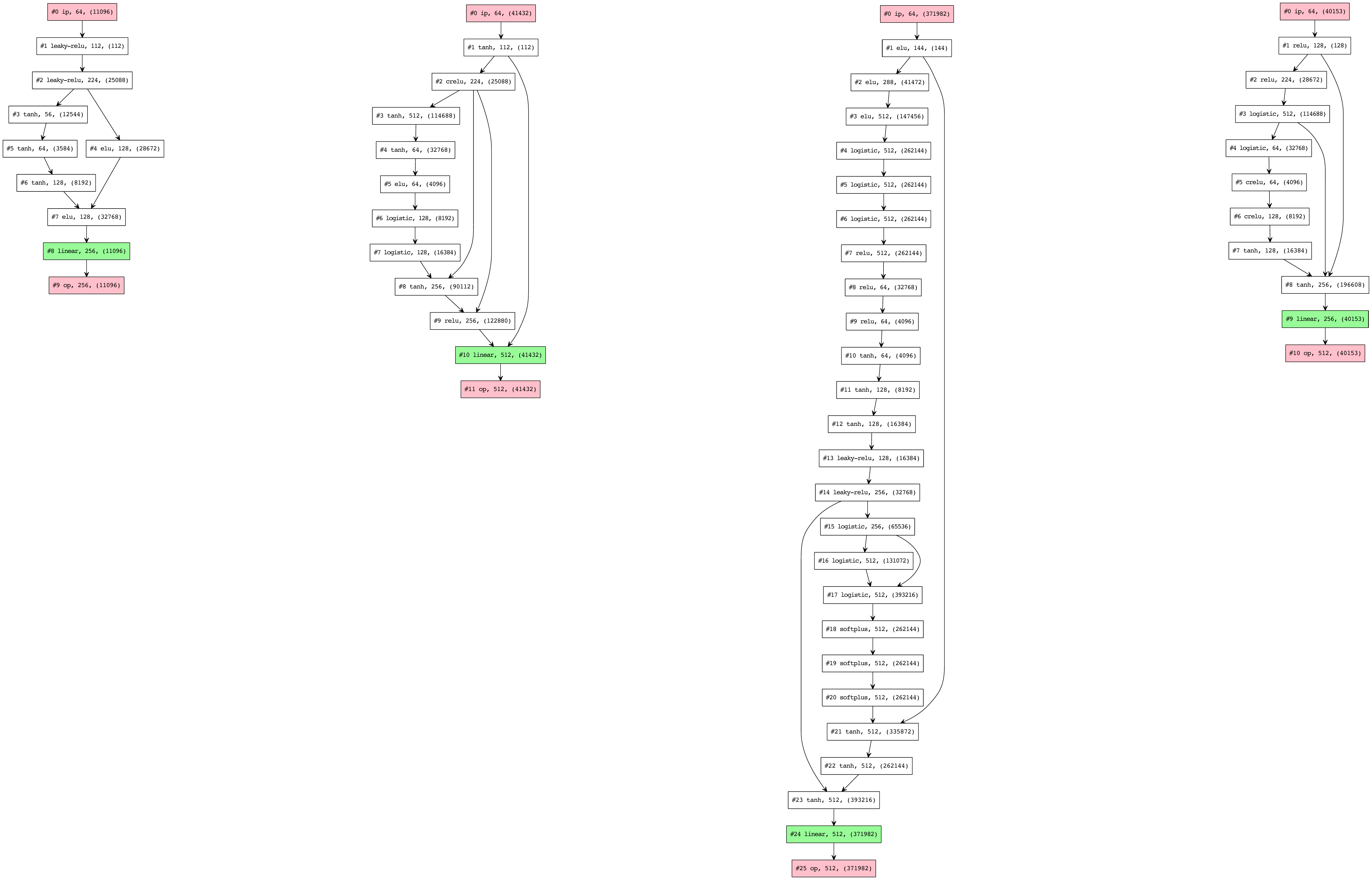}
\vspace{-0.1in}
\caption{\small
\label{fig:bestNetSliceRAND}
{Optimal network architectures found with \rands on Slice data.}
\vspace{-0.15in}
}
\end{figure*}
}

\newcommand{\insertFigBestNetSliceTREE}{
\begin{figure*}
\centering
\includegraphics[width=\halfPageFigWidth]{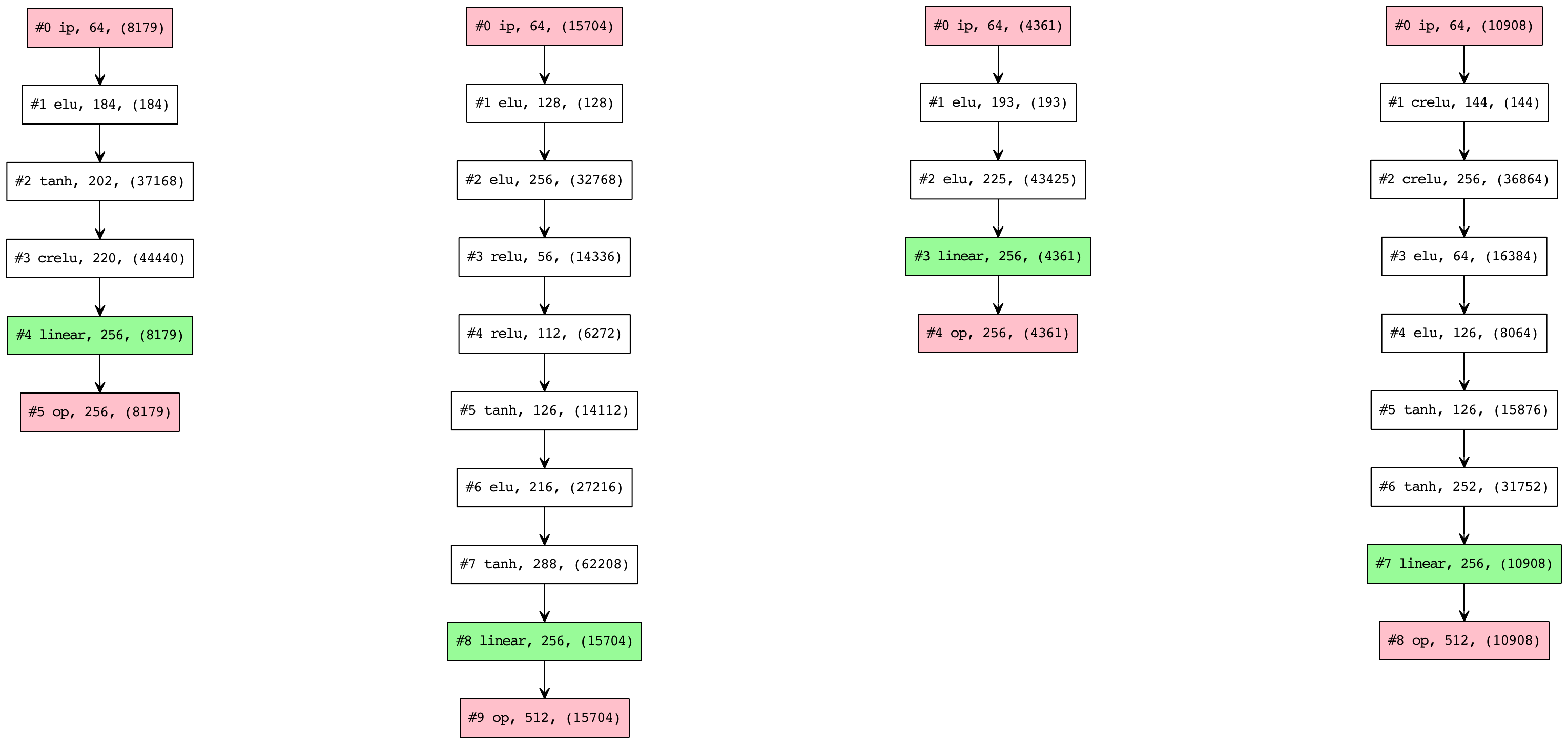}
\vspace{-0.1in}
\caption{\small
\label{fig:bestNetSliceTREE}
{Optimal network architectures found with \treebos on Slice data.}
\vspace{-0.15in}
}
\end{figure*}
}


\newcommand{\insertFigInitialPoolMLP}{
\begin{figure*}
\centering
\includegraphics[height=\initPoolFigHeight]{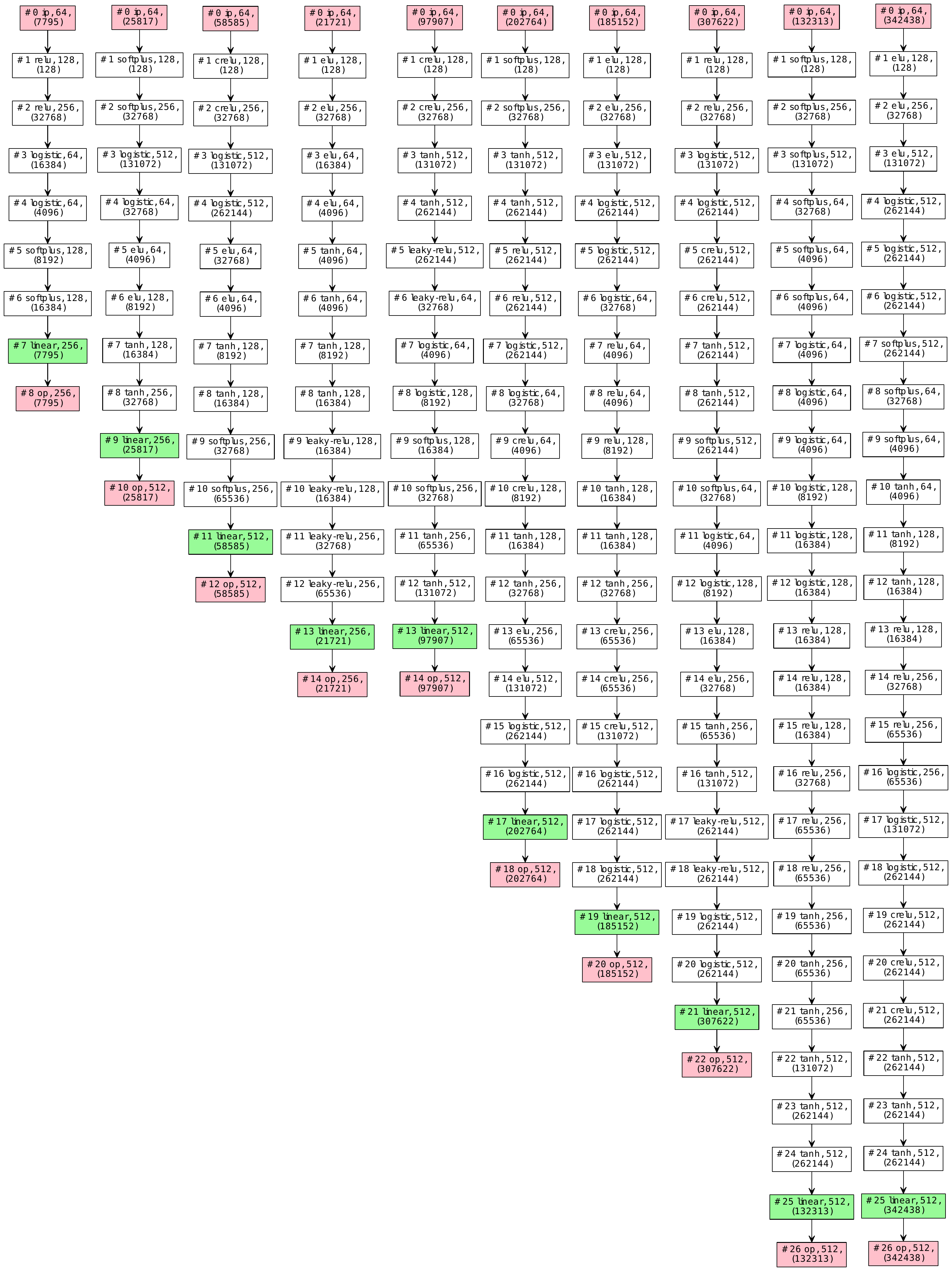}
\vspace{-0.1in}
\caption{\small
\label{fig:initPoolMLP}
{Initial pool of MLP network architectures.
}
\vspace{-0.15in}
}
\end{figure*}
}

\newcommand{\insertFigInitialPoolCNN}{
\begin{figure*}
\centering
\includegraphics[height=\initPoolFigHeight]{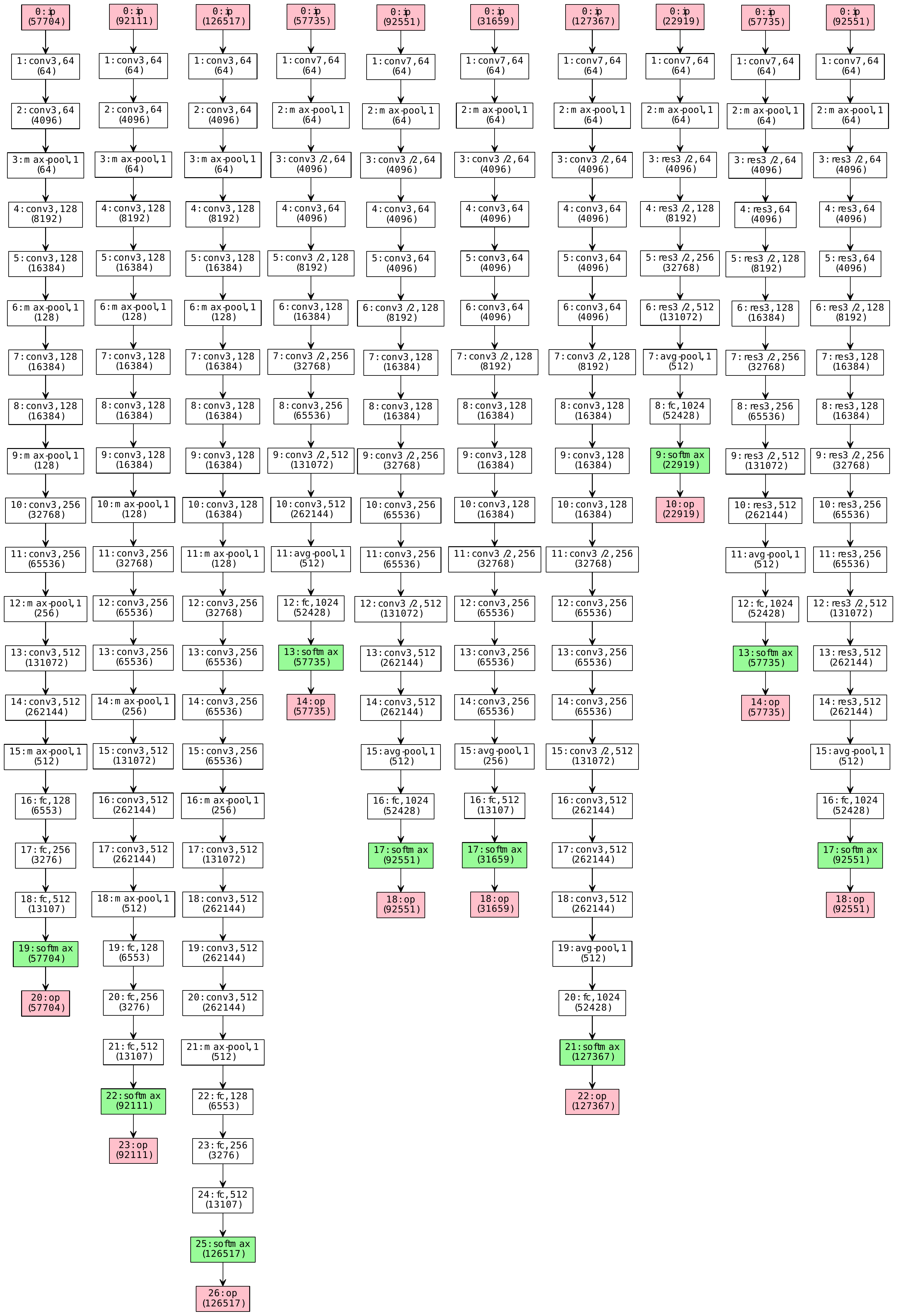}
\vspace{-0.1in}
\caption{\small
\label{fig:initPoolCNN}
{Initial pool of CNN network architectures.
The first $3$ networks have structure similar to the VGG
nets~\citep{simonyan2014very} and the remaining have blocked feed forward structures
as in~\citet{he2016deep}.
}
\vspace{-0.15in}
}
\end{figure*}
}


\newcommand{\insertLabelMismatchTable}{
\begin{table}
\begin{tabular}{c|cccccc}
& \inlabelfont{c3} & \inlabelfont{c5} & \inlabelfont{c7} & \inlabelfont{mp} &
\inlabelfont{ap} & \inlabelfont{fc} \\
\hline
\inlabelfont{c3} & $0$ & $0.2$ & $0.28$  & $\infty$ & $\infty$ & $\infty$  \\
\inlabelfont{c5} & $0.2$ & $0$ & $0.2$  & $\infty$ & $\infty$ & $\infty$  \\
\inlabelfont{c7} & $0.28$ & $0.2$ & $0$  & $\infty$ & $\infty$ & $\infty$  \\
\inlabelfont{mp} & $\infty$ & $\infty$ & $\infty$  & $0$ & $0.25$ & $\infty$  \\
\inlabelfont{ap} & $\infty$ & $\infty$ & $\infty$  & $0.25$ & $0$ & $\infty$  \\
\inlabelfont{fc} & $\infty$ & $\infty$ & $\infty$  & $\infty$ & $\infty$ & $0$  \\
\end{tabular}
\caption{\small
Part of the label mismatch cost matrix $\mislabmat$ we used in our CNN
experiments.
$\mislabmat(\textrm{\inlabelfont{x}},\textrm{\inlabelfont{y}})$ denotes the penalty for
transporting a unit
mass from a layer with label \inlabelfont{x} to a layer with label \inlabelfont{y}.
The labels abbreviated are
\convthree, \convfive, \convseven, \maxpool, \avgpool, and \fc{} in order.
There is zero cost for transporting mass among identical layers,
small cost for similar layers,
and infinite cost for highly disparate layers.
\label{tb:mislabmat}
}
\end{table}
}

\newcommand{\insertLabelMismatchTableSmall}{
\newcommand{\lmmsmalltablehspace}{\hspace{-0.05in}}
\newcommand{\mislabmatcolwidth}{8mm}
\begin{table}
\centering
\small
\begin{minipage}{3.4in}
\begin{tabular}{l|cccccc}
& \lmmsmalltablehspace{\small\convthree} \lmmsmalltablehspace
& \lmmsmalltablehspace{\small\convfive} \lmmsmalltablehspace
& \lmmsmalltablehspace{\small\maxpool} \lmmsmalltablehspace
& \lmmsmalltablehspace{\small\avgpool} \lmmsmalltablehspace
& \lmmsmalltablehspace{\small\fc} \lmmsmalltablehspace\\
\hline
{\small \convthree} & $0$ & $0.2$  & $\infty$ & $\infty$ & $\infty$  \\
{\small \convfive} & $0.2$ & $0$ & $\infty$ & $\infty$ & $\infty$  \\
{\small \maxpool}\hspace{-0.05in} & $\infty$ & $\infty$  & $0$ & $0.25$ & $\infty$  \\
{\small \avgpool}\hspace{-0.05in} & $\infty$ & $\infty$  & $0.25$ & $0$ & $\infty$  \\
{\small \fc} & $\infty$ & $\infty$  & $\infty$ & $\infty$ & $0$  \\
\end{tabular}
\end{minipage}
\begin{minipage}{2in}
\caption{\small
An example label mismatch cost matrix $\mislabmat$.
There is zero cost for matching identical layers,
$<1$ cost for similar layers,
and infinite cost for disparate layers.
\label{tb:mislabmatsmall}
}
\end{minipage}
\vspace{-0.13in}
\end{table}
}

\newcommand{\insertLabelMismatchTableWithSink}{
\begin{table}
\begin{tabular}{c|cccccc|c}
& \inlabelfont{c3} & \inlabelfont{c5} & \inlabelfont{c7} & \inlabelfont{mp} &
\inlabelfont{ap} & \inlabelfont{fc} & $\sinknode$ \\
\hline
\inlabelfont{c3} & $0$ & $0.1$ & $0.14$  & $\infty$ & $\infty$ & $\infty$ & 1 \\
\inlabelfont{c5} & $0.1$ & $0$ & $0.1$  & $\infty$ & $\infty$ & $\infty$ & 1 \\
\inlabelfont{c7} & $0.14$ & $0.1$ & $0$  & $\infty$ & $\infty$ & $\infty$ & 1 \\
\inlabelfont{mp} & $\infty$ & $\infty$ & $\infty$  & $0$ & $0.25$ & $\infty$ & 1 \\
\inlabelfont{ap} & $\infty$ & $\infty$ & $\infty$  & $0.25$ & $0$ & $\infty$ & 1 \\
\inlabelfont{fc} & $\infty$ & $\infty$ & $\infty$  & $\infty$ & $\infty$ & $0$ & 1 \\
\hline
$\sinknode$      & 1 & 1 & 1 & 1 & 1 & 1 & $0$ \\
\end{tabular}
\caption{\small
The first six rows and columns
depict part of the label mismatch cost matrix $\mislabmat$ we used in our CNN
experiments.
$\mislabmat(\textrm{\inlabelfont{x}},\textrm{\inlabelfont{y}})$ denotes the penalty for
transporting a unit
mass from a layer with label \inlabelfont{x} to a layer with label \inlabelfont{y}.
The labels abbreviated are
\convthree, \convfive, \convseven, \maxpool, \avgpool, and \fc{} in order.
There is zero cost for transporting mass among identical layers,
small cost for similar layers,
and infinite cost for highly disparate layers.
The last row and columnn show the cost for transporting to the non assignment layer
 as per the reformulation
in~\eqref{eqn:nndistdefntwo} -- i.e. there is cost $1$ for not assigning
a unit mass from one network to another in ~\eqref{eqn:nndistdefntwo}.
Accordingly, the optimiser is incentivised
to transport mass among similar layers rather than not assign it provided that
the structural penalty is not too large.
\label{tb:mislabmat}
}
\end{table}
}

\newcommand{\skipmultirowheight}{*}
\definecolor{lightblue}{rgb}{0.68, 0.85, 0.9}
\definecolor{lightskyblue}{rgb}{0.53, 0.81, 0.98}
\newcommand{\coloursmall}{\cellcolor{white}}
\newcommand{\colourbig}{\cellcolor{white}}

\newcommand{\insertTableEAModifiersSmall}{
\begin{table*}
\centering
\footnotesize
\begin{tabular}{|m{15.4mm}|m{4.55in}|}
\hline
\textbf{Operation} & \multicolumn{1}{c|}{\textbf{Description}} \\
\hline
{\tiny \decsingle} \coloursmall &
Pick a layer at random and decrease the number of units by
                          $1/8$. \\
\hline
\decenmasse
\colourbig & 
Pick several layers at random and decrease the number of units by $1/8$ for all of them.
\\
\hline
\incsingle \coloursmall & Pick a layer at random and increase the number of units by
                          $1/8$. \\
\hline
\incenmasse \colourbig &
Pick several layers  at random and increase the number of units by $1/8$ for all of them.
\\
\hline
\hline
\branch \colourbig &
Pick a random path $u_1,\dots,u_k$, duplicate
$u_2,\dots,u_{k-1}$ and connect them to $u_1$ and $u_k$.
\\
\hline
\removelayer \coloursmall &
Pick a layer at random and remove it.
Connect the layer's parents to its children if necessary.
 \\
\hline
\colourbig \skiplayer &
Randomly pick layers $u,v$ where $u$ is topologically
before $v$. Add $(u,v)$ to $\edges$.
\\
\hline
\swaplayer \coloursmall & Randomly pick a layer and change its label. \\
\hline
\wedgelayer \coloursmall & Randomly remove an edge $(u,v)$ from $\edges$.
Create a new layer $w$ and add  $(u,w), (w,v)$ to $\edges$.
\\
\hline
\end{tabular}
\vspace{-0.05in}
\caption{\small
Descriptions of modifiers to transform one network to another.
The first four change the number of units in the layers but do not change the
architecture, while the last five change the architecture.
\label{tb:nnmodifierssmall}
}
\vspace{-0.20in}
\end{table*}
}

\newcommand{\insertTableEAModifiers}{
\begin{table*}
\centering
\begin{tabular}{|c|p{4.5in}|}
\hline
\textbf{Operation} & \multicolumn{1}{c|}{\textbf{Description}} \\
\hline
\decsingle \coloursmall & Pick a layer at random and decrease the number of units by
                          $1/8$. \\
\hline
\decenmasse
\colourbig & 
First topologically order the networks, randomly pick $1/8$ of the layers (in order) and
decrease the number of units by $1/8$.
For networks with eight layers or fewer pick a $1/4$ of the layers (instead of 1/8)
and for those with four layers or fewer pick $1/2$.
\\
\hline
\incsingle \coloursmall & Pick a layer at random and increase the number of units by
                          $1/8$. \\
\hline
\incenmasse \colourbig & Choose a large sub set of layers, as for \decenmasse{}, and
  increase the number of units by $1/8$. \\
\hline
\hline
\branch \colourbig &
This modifier duplicates a random path in the network.
Randomly pick a node $u_1$ and then pick one of its children $u_2$ randomly.
Keep repeating to generate
a path $u_1,u_2,\dots,u_{k-1}, u_k$ until you decide to stop randomly.
Create duplicate layers $\tilde{u}_2,\dots,\tilde{u}_{k-1}$ where
$\tilde{u}_i = u_i$ for $i=2,\dots,k-1$.
Add these layers along with new edges $(u_1, \tilde{u}_2)$, $(\tilde{u}_{k-1}, u_k)$,
and $(\tilde{u}_{j}, \tilde{u}_{j+1})$ for $j=2,\dots,k-2$.
\\
\hline
\removelayer \coloursmall &
Picks a layer at random and removes it.
If this layer was the only child (parent) of  any of its parents (children) $u$,
then adds an edge from $u$ (one of its parents) to one of its children ($u$).
 \\
\hline
\colourbig \skiplayer &
Randomly picks layers $u,v$ where $u$ is topologically
before $v$ and $(u,v)\notin\edges$. Add $(u,v)$ to $\edges$.
\\
\hline
\swaplayer \coloursmall & Randomly pick a layer and change its label. \\
\hline
\wedgelayer \coloursmall & Randomly pick any edge $(u,v)\in\edges$.
Create a new layer $w$ with a random label $\laylabel(w)$.
Remove $(u,v)$ from $\edges$ and add $(u,w), (w,v)$.
If applicable, set the number of units $\layunits(w)$ to be $(\layunits(u) +
\layunits(v))/2$.
\\
\hline
\end{tabular}
\vspace{-0.05in}
\caption{\small
Descriptions of modifiers to transform one network to another.
The first four change the number of units in the layers but do not change the
architecture, while the last five change the architecture.
\label{tb:nnmodifiers}
}
\vspace{-0.15in}
\end{table*}
}

\newcommand{\insertTestResultsTableHor}{
\newcommand{\hrcolwidth}{11mm}
\begin{table*}
\centering
\footnotesize
\begin{tabular}{%
m{12mm}|m{\hrcolwidth}|m{\hrcolwidth}|m{\hrcolwidth}|m{\hrcolwidth}|m{\hrcolwidth}|%
m{\hrcolwidth}|m{\hrcolwidth}|m{\hrcolwidth}}
\toprule
Method
  & Blog \newline {\scriptsize $(60K, 281)$}
  & Indoor \newline {\scriptsize $(21K, 529)$}
  & Slice \newline {\scriptsize $(54K, 385)$}
  & Naval \newline {\scriptsize $(12K, 17)$}
  & Protein \newline {\scriptsize $(46K, 9)$}
  & News \newline {\scriptsize $(40K, 61)$}
  & Cifar10 \newline {\scriptsize $(60K, 3K)$}
  & Cifar10 \newline {\tiny $150K$ iters} \\
\midrule
\rand &  $0.780 \newline \pm 0.034$
      &  $\bf 0.115 \newline \pm 0.023$
      &  $0.758 \newline \pm 0.041$
      &  $0.0103 \newline \pm 0.002$
      &  $0.948 \newline \pm 0.024$
      &  $\bf 0.762 \newline \pm 0.013$
      &  $0.1342 \newline \pm 0.002$
      &  $0.0914 \newline \pm 0.008$
\\
\midrule
\evoalg &  $0.806 \newline \pm 0.040$
        &  $0.147 \newline \pm 0.010$
        &  $0.733 \newline \pm 0.041$
        &  $\bf 0.0079 \newline \pm 0.004$
        &  $1.010 \newline \pm 0.038$
        &  $\bf 0.758 \newline \pm 0.038$
        &  $0.1411 \newline \pm 0.002$
        &  $0.0915 \newline \pm 0.010$
\\
\midrule
\treebo &  $0.928 \newline \pm 0.053$
        &  $0.168 \newline \pm 0.023$
        &  $0.759 \newline \pm 0.079$
        &  $0.0102 \newline \pm 0.002$
        &  $0.998 \newline \pm 0.007$
        &  $0.866 \newline \pm 0.085$
        &  $0.1533 \newline \pm 0.004$
        &  $0.1121 \newline \pm 0.004$
\\
\midrule
\nnbo &  $\bf 0.731 \newline \pm 0.029$
      &  $\bf 0.117 \newline \pm 0.008$
      &  $\bf 0.615 \newline \pm 0.044$
      &  $\bf 0.0075 \newline \pm 0.002$
      &  $\bf 0.902 \newline \pm 0.033$
      &  $\bf 0.752 \newline \pm 0.024$
      &  $\bf 0.1209 \newline \pm 0.003$
      &  $\bf 0.0869 \newline \pm 0.004$
\\
\bottomrule
\end{tabular}
\vspace{-0.05in}
\caption{\small
The first row gives the number of samples $N$ and the dimensionality $D$
of each dataset in the form $(N, D)$.
The subsequent rows show the regression MSE or classification error (lower is better)
on the \emph{test set} for each method.
The last column is for Cifar10 where we took the best models found by each method in 24K
iterations and trained it for $120K$ iterations.
When we trained the VGG-19 architecture using our training procedure,
we got test errors $0.1718$ (60K iterations) and \textcolor{black}{$0.1018$ (150K iterations)}.
\label{tb:testresults}
}
\vspace{-0.20in}
\end{table*}
}

\newcommand{\insertTestResultsTableVer}{
\newcommand{\vrcolwidth}{11.57mm}
\begin{table}
\begin{center}
\small
\begin{tabular}{%
m{12.8mm}|m{\vrcolwidth}|m{12.79mm}|m{\vrcolwidth}|m{12.79mm}}
\toprule
\centering Method &  \centering\rand & \centering \evoalg &  \centering\treebo & \nnbo  \\
\toprule
\centering Blog \newline $(60K,281)$
      & \centering $0.780 \newline \pm 0.034$
      & \centering $0.806 \newline \pm 0.040$
      & \centering $0.928 \newline \pm 0.053$
      &            $\bf 0.731 \newline \pm 0.029$
\\
\midrule
\centering
Indoor  \newline $(21K,529)$
      & \centering $\bf 0.115 \newline \pm 0.023$
      & \centering $0.147 \newline \pm 0.010$
      & \centering $0.168 \newline \pm 0.023$
      &            $\bf 0.117 \newline \pm 0.008$
\\
\midrule
\centering
Slice  \newline $(54K,385)$
      & \centering $0.758 \newline \pm 0.041$
      & \centering $0.733 \newline \pm 0.041$
      & \centering $0.759 \newline \pm 0.079$
      &            $\bf 0.615 \newline \pm 0.044$
\\
\midrule
\centering
Naval \newline $(12K, 17)$
      & \centering $0.0103 \newline \pm 0.0017$
      & \centering $\bf0.0079 \newline \pm 0.0044$
      & \centering $0.0102 \newline \pm 0.0017$
      &            $\bf 0.0075 \newline \pm 0.0021$
\\
\midrule
\centering
Protein  \newline $(46K, 9)$
      & \centering $0.948 \newline \pm 0.024$
      & \centering $1.010 \newline \pm 0.038$
      & \centering $0.998 \newline \pm 0.007$
      &            $\bf 0.902 \newline \pm 0.033$
\\
\midrule
\centering
News  \newline $(40K, 61)$
      & \centering $\bf0.762 \newline \pm 0.013$
      & \centering $\bf0.758 \newline \pm 0.038$
      & \centering $0.866 \newline \pm 0.085$
      &            $\bf0.7523 \newline \pm 0.024$
\\
\midrule
\centering
Cifar10  \newline $(60K, 1K)$
      & \centering $0.293 \newline \pm 0.031$
      & \centering $0.259 \newline \pm 0.003$
      & \centering $0.298 \newline \pm 0.020$
      &            $\bf 0.232 \newline \pm 0.003$
\\
\midrule
\centering
Cifar10  \newline 120K iters 
      & \centering $0.161 \newline \pm 0.018$
      & \centering $0.142 \newline \pm 0.003$
      & \centering $0.187 \newline \pm 0.021$
      &            $\bf 0.123 \newline \pm 0.003$
\\
\bottomrule
\end{tabular}
\end{center}
\vspace{-0.1in}
\caption{\small
The first column gives the number of samples $N$ and the dimensionality $D$
of each dataset in the form $(N, D)$.
The subsequent columns show the regression MSE or classification error (lower is better)
on the \emph{test set} for each method.
The last row is for Cifar10 where we took the best models found by each method in 24K
iterations and trained it for $120K$ iterations.
When we trained the VGG-19 architecture using our training procedure,
we got test errors $0.310$ (24K iterations) and $0.151$ (120K iterations).
\label{tb:testresults}
}
\vspace{-0.2in}
\end{table}
}

\newcommand{\lbinf}{}

\newcommand{\insertLabPenCNN}{
\begin{table}
\centering
\begin{minipage}{2.8in}
\begin{tabular}{|c||ccc|cc|c|c|}
\hline
& \inlabelfont{c3} & \inlabelfont{c5} &
\inlabelfont{c7} & \inlabelfont{mp} &
\inlabelfont{ap} & \inlabelfont{fc} & \inlabelfont{sm} \\
\hline
\hline
\inlabelfont{c3}  & $0$ & $0.2$ & $0.3$  & $\lbinf$ & $\lbinf$ & $\lbinf$   & $\lbinf$\\
\inlabelfont{c5}  & $0.2$ & $0$ & $0.2$  & $\lbinf$ & $\lbinf$ & $\lbinf$   & $\lbinf$\\
\inlabelfont{c7}  & $0.3$ & $0.2$ & $0$  & $\lbinf$ & $\lbinf$ & $\lbinf$  & $\lbinf$ \\
\hline
\inlabelfont{mp}  & $\lbinf$ & $\lbinf$ & $\lbinf$  & $0$ & $0.25$ & $\lbinf$   & $\lbinf$\\
\inlabelfont{ap}  & $\lbinf$ & $\lbinf$ & $\lbinf$  & $0.25$ & $0$ & $\lbinf$  & $\lbinf$ \\
\hline
\inlabelfont{fc} & $\lbinf$ & $\lbinf$ & $\lbinf$  & $\lbinf$ &
$\lbinf$ & $0$   & $\lbinf$\\
\hline
\inlabelfont{sm}  & $\lbinf$ & $\lbinf$ & $\lbinf$  & $\lbinf$ &
$\lbinf$ & $\lbinf$   & $0$\\
\hline
\end{tabular}
\end{minipage}
\hspace{0.15in}
\begin{minipage}{2.45in}
\vspace{-0.05in}
\caption{\small
The label mismatch cost matrix $\mislabmat$ we used in our CNN
experiments.
$\mislabmat(\textrm{\inlabelfont{x}},\textrm{\inlabelfont{y}})$ denotes the penalty for
transporting a unit
mass from a layer with label \inlabelfont{x} to a layer with label \inlabelfont{y}.
The labels abbreviated are
\convthree, \convfive, \convseven, \maxpool, \avgpool, \fc, and \softmax{} in order.
A blank indicates $\infty$ cost.
We have not shown the \iplab{} and \oplab{} layers, but they are similar to
the \fc{} column, $0$ in the diagonal and $\infty$ elsewhere.
\label{tb:mislabmatcnn}
}
\end{minipage}
\end{table}
}

\newcommand{\insertLabPenMLP}{
\begin{table}
\centering
\begin{minipage}{2.8in}
\begin{tabular}{|c||ccc|cc|c|c|}
\hline
& \inlabelfont{re} & \inlabelfont{cr} &
\inlabelfont{<rec>} & \inlabelfont{lg} &
\inlabelfont{ta} & \inlabelfont{lin} \\
\hline
\hline
\inlabelfont{re}      & $0$ & $.1$ & $.1$  & $.25$ & $.25$  & $\lbinf$\\
\inlabelfont{cr}      & $.1$ & $0$ & $.1$  & $.25$ & $.25$  & $\lbinf$\\
\inlabelfont{<rec>}  & $.1$ & $.1$ & $0$  & $.25$ & $.25$  & $\lbinf$ \\
\hline
\inlabelfont{lg}  & $.25$ & $.25$ & $.25$  & $0$ & $.1$   & $\lbinf$\\
\inlabelfont{ta}  & $.25$ & $.25$ & $.25$  & $.1$ & $0$  & $\lbinf$ \\
\hline
\inlabelfont{lin}  & $\lbinf$ & $\lbinf$ & $\lbinf$  & $\lbinf$ &
$\lbinf$ &  $0$\\
\hline
\end{tabular}
\end{minipage}
\hspace{0.10in}
\begin{minipage}{2.50in}
\vspace{0.05in}
\caption{\small
The label mismatch cost matrix $\mislabmat$ we used in our MLP
experiments.
The labels abbreviated are
\relu, \crelu, \inlabelfont{<rec>}, \logistic, \tanhlabel, and \linear{} in order.
\inlabelfont{<rec>} is place-holder for any other rectifier such as
\leakyrelu, \softplus, \elu.
A blank indicates $\infty$ cost.
The design here was simple. Each label gets $0$ cost with itself.
A rectifier gets $0.1$ cost with another rectifier and $0.25$ with a sigmoid; vice
versa for all sigmoids.
The rest of the costs are infinity.
We have not shown the \iplab{} and \oplab, but they are similar to
the \inlabelfont{lin}{} column, $0$ in the diagonal and $\infty$ elsewhere.
\label{tb:mislabmatmlp}
}
\end{minipage}
\end{table}
}

\newcommand{\insertpathlengthalgo}{%
\begin{algorithm}
\vspace{0.02in}
\begin{algorithmic}[1]
\REQUIRE $\Gcal=(\Lcal,\Ecal)$,
$\Lcal$ is topologically sorted in $S$.
\STATE $\distoprw(\opnode) = 0$, $\distoprw(u) = \textrm{\inlabelfont{nan}}\;\,
  \forall u\neq \opnode$.
\WHILE{$S$ is not empty}
\STATE $u\leftarrow {\rm pop\_last}(S)$
\STATE $\Delta \leftarrow \{\distoprw(c): c \in {\rm children}(u)\}$ \label{line:child}
\STATE $\distoprw(u) \leftarrow 1 + {\rm average}(\Delta)$
\ENDWHILE
\STATE \textbf{Return} $\distoprw$.
\end{algorithmic}
\caption{$\;$Compute $\distoprw(u)$ for all $u\in\Lcal$ \label{alg:rwpl}}
\end{algorithm}
}


\newcommand{\insertResultsFig}{
\newcommand{\modselfigwidth}{1.505in}
\newcommand{\modselfighsp}{\hspace{-0.00in}}
\begin{figure*}[t]
\begin{minipage}{3.15in}
\subfloat{
\includegraphics[width=\modselfigwidth]{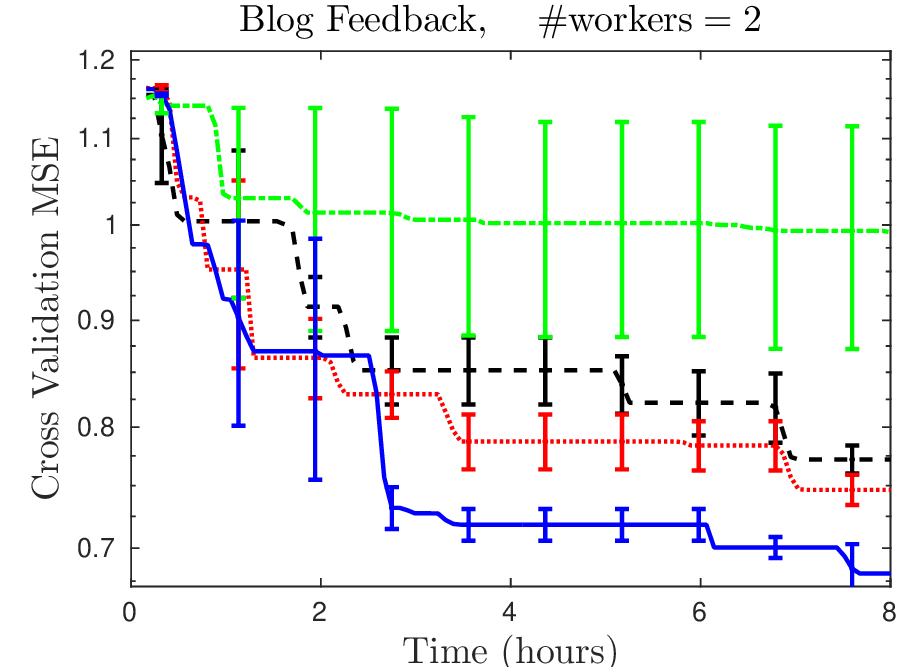}
\label{fig:modselblog}} \modselfighsp
\subfloat{
\includegraphics[width=\modselfigwidth]{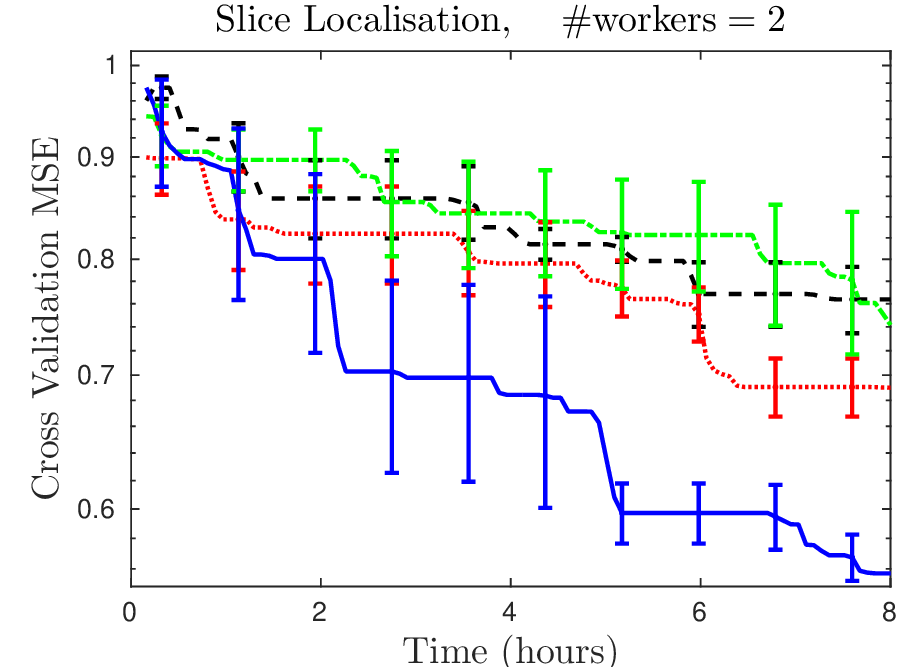}
\label{fig:modselslice}} 
\\[-0.1in]
\subfloat{
\includegraphics[width=\modselfigwidth]{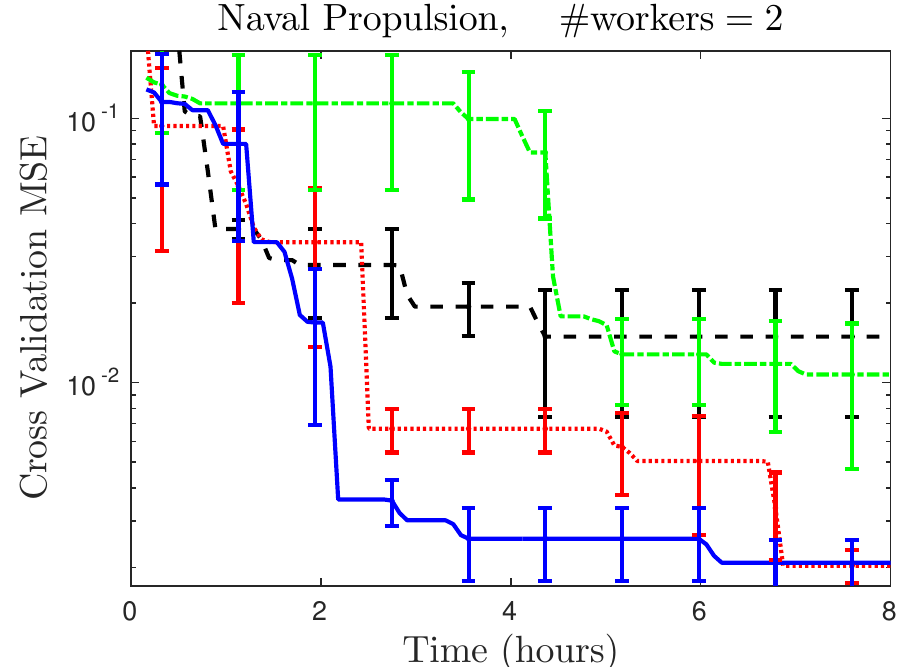}
\label{fig:modselnaval}} \modselfighsp
\subfloat{
\includegraphics[width=\modselfigwidth]{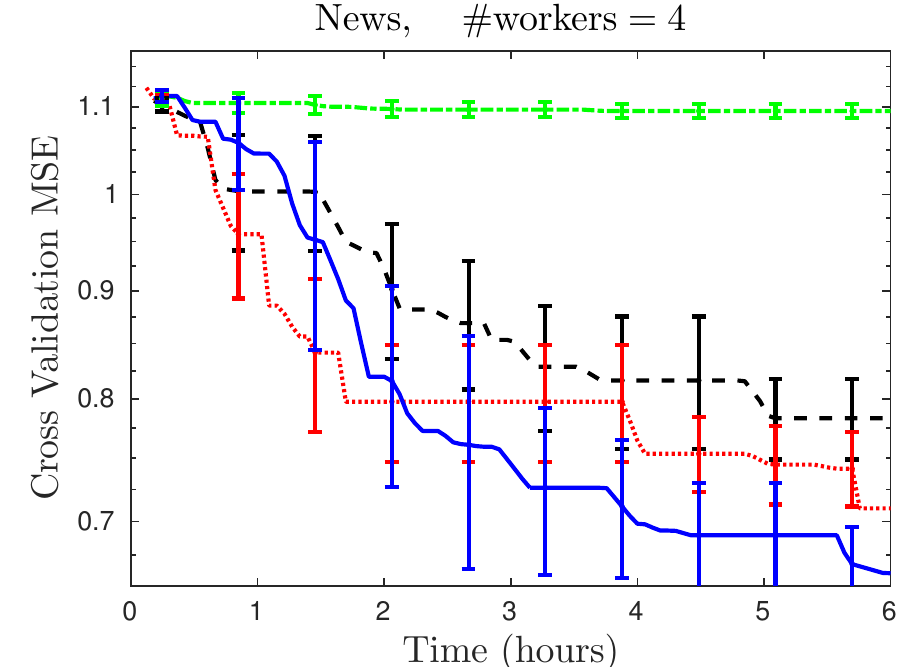}
\label{fig:modselnews}} 
\\[-0.1in]
\subfloat{
\includegraphics[width=\modselfigwidth]{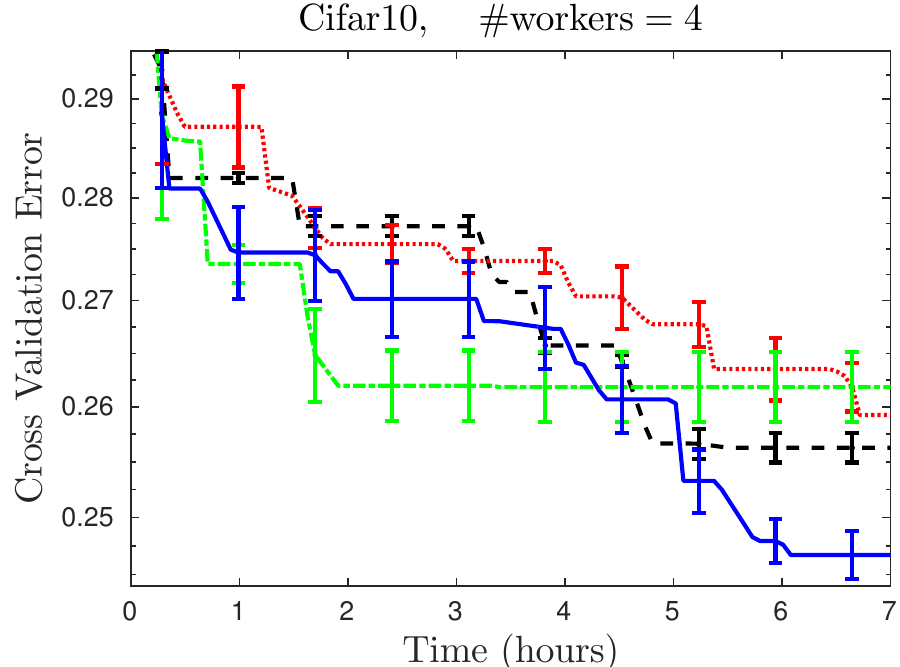}
\label{fig:modselcifar}} \modselfighsp
\hspace{0.45in}
\subfloat{
\includegraphics[width=0.7in]{figs/legend}
\label{fig:modselcifar}
} 
\end{minipage}
\hspace{0.01in}
\begin{minipage}{3in}
\newcommand{\vrcolwidth}{9.57mm}
{\footnotesize
\begin{tabular}{%
m{12.8mm}|m{\vrcolwidth}|m{12.79mm}|m{\vrcolwidth}|m{12.79mm}}
\toprule
\centering Method &  \centering\rand & \centering \evoalg &  \centering\treebo & \nnbo  \\
\toprule
\centering Blog \newline $(60K,281)$
      & \centering $0.780 \newline \pm 0.034$
      & \centering $0.806 \newline \pm 0.040$
      & \centering $0.928 \newline \pm 0.053$
      &            $\bf 0.731 \newline \pm 0.029$
\\
\midrule
\centering
Indoor  \newline $(21K,529)$
      & \centering $\bf 0.115 \newline \pm 0.023$
      & \centering $0.147 \newline \pm 0.010$
      & \centering $0.168 \newline \pm 0.023$
      &            $\bf 0.117 \newline \pm 0.008$
\\
\midrule
\centering
Slice  \newline $(54K,385)$
      & \centering $0.758 \newline \pm 0.041$
      & \centering $0.733 \newline \pm 0.041$
      & \centering $0.759 \newline \pm 0.079$
      &            $\bf 0.615 \newline \pm 0.044$
\\
\midrule
\centering
Naval \newline $(12K, 17)$
      & \centering $0.0103 \newline \pm 0.0017$
      & \centering $\bf0.0079 \newline \pm 0.0044$
      & \centering $0.0102 \newline \pm 0.0017$
      &            $\bf 0.0075 \newline \pm 0.0021$
\\
\midrule
\centering
Protein  \newline $(46K, 9)$
      & \centering $0.948 \newline \pm 0.024$
      & \centering $1.010 \newline \pm 0.038$
      & \centering $0.998 \newline \pm 0.007$
      &            $\bf 0.902 \newline \pm 0.033$
\\
\midrule
\centering
News  \newline $(40K, 61)$
      & \centering $\bf0.762 \newline \pm 0.013$
      & \centering $\bf0.758 \newline \pm 0.038$
      & \centering $0.866 \newline \pm 0.085$
      &            $\bf0.7523 \newline \pm 0.024$
\\
\midrule
\centering
Cifar10  \newline $(60K, 1K)$
      & \centering $0.293 \newline \pm 0.031$
      & \centering $0.259 \newline \pm 0.003$
      & \centering $0.298 \newline \pm 0.020$
      &            $\bf 0.232 \newline \pm 0.003$
\\
\midrule
\centering
Cifar10  \newline 120K iters 
      & \centering $0.161 \newline \pm 0.018$
      & \centering $0.142 \newline \pm 0.003$
      & \centering $0.187 \newline \pm 0.021$
      &            $\bf 0.123 \newline \pm 0.003$
\\
\bottomrule
\end{tabular}
}
\end{minipage}
\caption{\small
\emph{Cross validation results:}
In all figures, the $x$ axis is time.
The $y$ axis is the mean squared error (MSE) in the first 6 figures and
the classification error in the last. Lower is better in all cases.
The title of each figure states the dataset and the number of parallel workers (GPUs).
All figures were averaged over at least $5$ independent runs of each method.
Error bars indicate one standard error. \hspace{-0.1in}
\label{fig:modsel}
\vspace{-0.15in}
}
\end{figure*}
}

\newcommand{\insertFigMLPCopy}{
\newcommand{\modselfigwidth}{1.195in}
\newcommand{\modselfighsp}{\hspace{0.00in}}
\newcommand{\modselfighsptwo}{\hspace{0.15in}}
\begin{figure*}[t]
\centering
\subfloat{
\includegraphics[height=\modselfigwidth]{figs/blog}
\label{fig:modselblog}} \modselfighsp
\subfloat{
\includegraphics[height=\modselfigwidth]{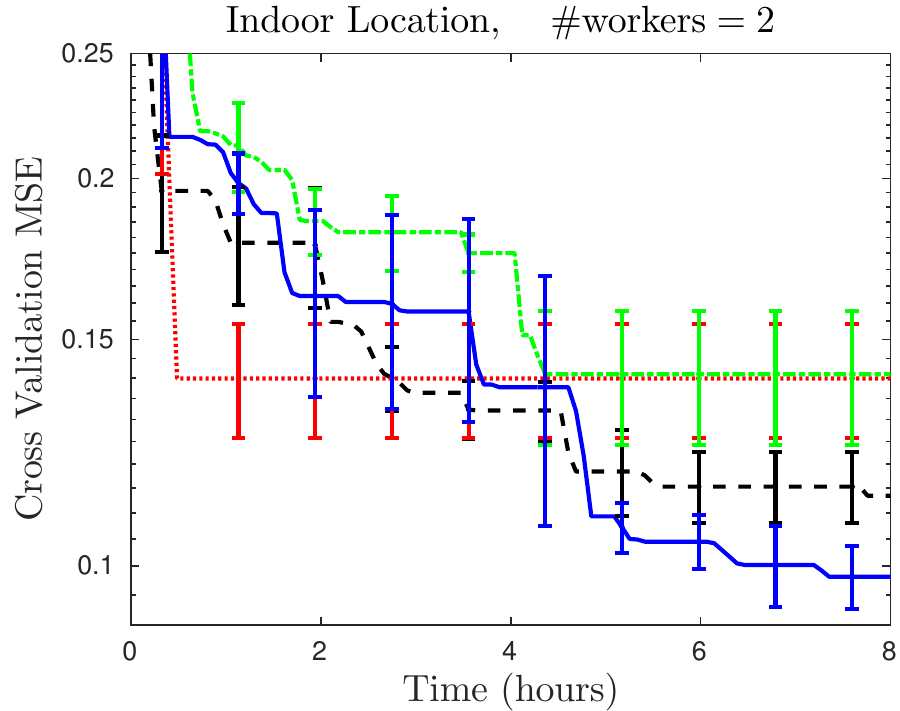}
\label{fig:modselindoor}} \modselfighsp
\subfloat{
\includegraphics[height=\modselfigwidth]{figs/slice}
\label{fig:modselslice}} \modselfighsp
\subfloat{
\includegraphics[height=\modselfigwidth]{figs/naval}
\label{fig:modselnaval}} \modselfighsp
\\[-0.1in]
\subfloat{
\includegraphics[height=\modselfigwidth]{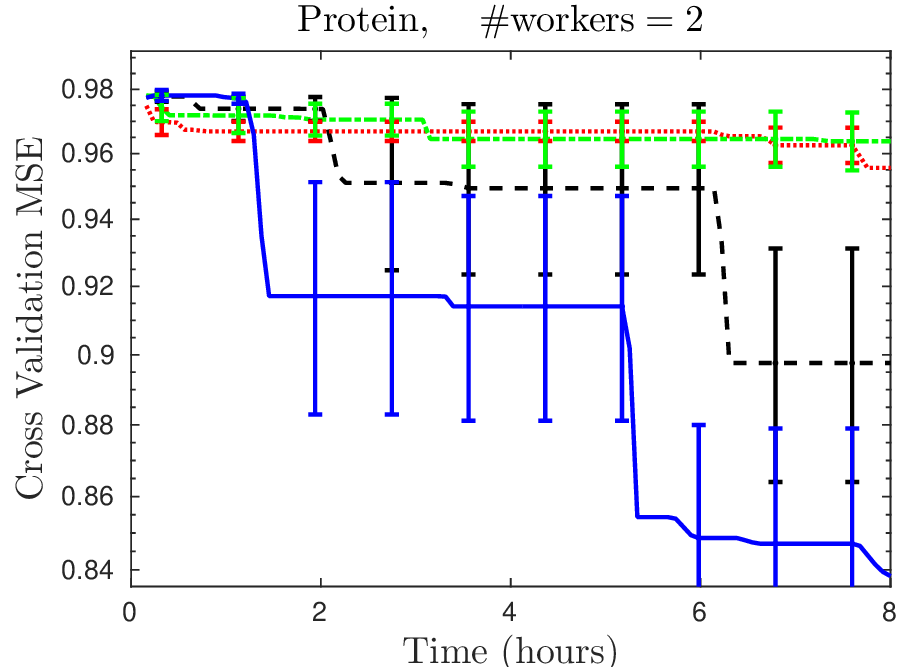}
\label{fig:modselprotein}} \modselfighsp
\subfloat{
\includegraphics[height=\modselfigwidth]{figs/news}
\label{fig:modselnews}} \modselfighsp
\subfloat{
\includegraphics[height=\modselfigwidth]{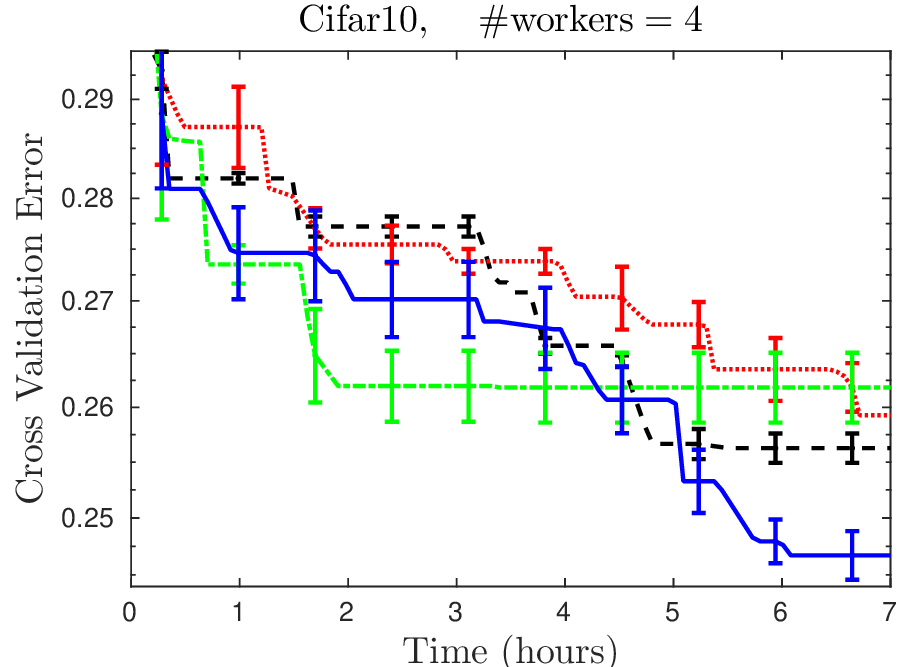}
\label{fig:modselcifar}} \modselfighsp
\hspace{0.15in}
\subfloat{
\includegraphics[width=0.7in]{figs/legend}
\label{fig:modselcifar}
} 
\vspace{-0.1in}
\caption{\small
\emph{Cross validation results:}
In all figures, the $x$ axis is time.
The $y$ axis is the mean squared error (MSE) in the first 6 figures and
the classification error in the last. Lower is better in all cases.
The title of each figure states the dataset and the number of parallel workers (GPUs).
All figures were averaged over at least $5$ independent runs of each method.
Error bars indicate one standard error. \hspace{-0.1in}
\label{fig:modsel}
\vspace{-0.15in}
}
\end{figure*}
}

\newcommand{\insertTestResultsTableVerCopy}{
\newcommand{\vrcolwidth}{11.57mm}
\begin{table}
\begin{center}
\small
\begin{tabular}{%
m{12.8mm}|m{\vrcolwidth}|m{12.79mm}|m{\vrcolwidth}|m{12.79mm}}
\toprule
\centering Method &  \centering\rand & \centering \evoalg &  \centering\treebo & \nnbo  \\
\toprule
\centering Blog \newline $(60K,281)$
      & \centering $0.780 \newline \pm 0.034$
      & \centering $0.806 \newline \pm 0.040$
      & \centering $0.928 \newline \pm 0.053$
      &            $\bf 0.731 \newline \pm 0.029$
\\
\midrule
\centering
Indoor  \newline $(21K,529)$
      & \centering $\bf 0.115 \newline \pm 0.023$
      & \centering $0.147 \newline \pm 0.010$
      & \centering $0.168 \newline \pm 0.023$
      &            $\bf 0.117 \newline \pm 0.008$
\\
\midrule
\centering
Slice  \newline $(54K,385)$
      & \centering $0.758 \newline \pm 0.041$
      & \centering $0.733 \newline \pm 0.041$
      & \centering $0.759 \newline \pm 0.079$
      &            $\bf 0.615 \newline \pm 0.044$
\\
\midrule
\centering
Naval \newline $(12K, 17)$
      & \centering $0.0103 \newline \pm 0.0017$
      & \centering $\bf0.0079 \newline \pm 0.0044$
      & \centering $0.0102 \newline \pm 0.0017$
      &            $\bf 0.0075 \newline \pm 0.0021$
\\
\midrule
\centering
Protein  \newline $(46K, 9)$
      & \centering $0.948 \newline \pm 0.024$
      & \centering $1.010 \newline \pm 0.038$
      & \centering $0.998 \newline \pm 0.007$
      &            $\bf 0.902 \newline \pm 0.033$
\\
\midrule
\centering
News  \newline $(40K, 61)$
      & \centering $\bf0.762 \newline \pm 0.013$
      & \centering $\bf0.758 \newline \pm 0.038$
      & \centering $0.866 \newline \pm 0.085$
      &            $\bf0.7523 \newline \pm 0.024$
\\
\midrule
\centering
Cifar10  \newline $(60K, 1K)$
      & \centering $0.293 \newline \pm 0.031$
      & \centering $0.259 \newline \pm 0.003$
      & \centering $0.298 \newline \pm 0.020$
      &            $\bf 0.232 \newline \pm 0.003$
\\
\midrule
\centering
Cifar10  \newline 120K iters 
      & \centering $0.161 \newline \pm 0.018$
      & \centering $0.142 \newline \pm 0.003$
      & \centering $0.187 \newline \pm 0.021$
      &            $\bf 0.123 \newline \pm 0.003$
\\
\bottomrule
\end{tabular}
\end{center}
\vspace{-0.1in}
\caption{\small
The first column gives the number of samples $N$ and the dimensionality $D$
of each dataset in the form $(N, D)$.
The subsequent columns show the regression MSE or classification error (lower is better)
on the \emph{test set} for each method.
The last row is for Cifar10 where we took the best models found by each method in 24K
iterations and trained it for $120K$ iterations.
When we trained the VGG-19 architecture using our training procedure,
we got test errors $0.310$ (24K iterations) and $0.151$ (120K iterations).
\label{tb:testresults}
}
\vspace{-0.2in}
\end{table}
}

\begin{abstract}
\vspace{-0.05in}
Bayesian Optimisation (BO) refers to a class of methods for global optimisation
of a function $\func$ which is only accessible via point evaluations.
It is typically used in settings where $\func$ is expensive to evaluate.
A common use case for BO in machine learning is model selection, where it is not
possible to
analytically model the generalisation performance of a statistical model, and we
resort to noisy and expensive training and validation procedures to choose the best model.
Conventional BO methods have focused on Euclidean and categorical domains,
which, in the context of model selection,
only permits tuning scalar hyper-parameters of machine learning algorithms.
However, with the surge of interest in deep learning, there is an increasing demand
to tune neural network \emph{architectures}.
In this work, we develop \nnbo, a Gaussian process based BO framework for neural
architecture search.
To accomplish this, we develop a distance metric in the space
of neural network architectures
which can be computed efficiently  via an optimal transport program.
This distance might be of independent interest to the deep learning community as it may
find applications outside of BO.
We demonstrate that \nnbos outperforms other alternatives for architecture search
in several cross validation based model selection tasks on multi-layer perceptrons
and convolutional neural networks.
\end{abstract}

\vspace{-0.10in}
\section{Introduction}
\label{sec:intro}
\vspace{-0.10in}

In many real world problems, we are required to sequentially evaluate a noisy
black-box function $\func$ with the goal of finding its optimum in some domain $\Xcal$.
Typically, each evaluation is expensive in such applications, and we need to keep
the number of evaluations to a minimum.
Bayesian optimisation (BO) refers to an approach for global optimisation
that is popularly used in such settings.
It uses Bayesian models for $\func$
to infer function values at unexplored regions and guide the selection
of points for future evaluations.
BO has been successfully applied for many optimisation problems in
optimal policy search, industrial design, and scientific experimentation.
That said, the quintessential use case for BO in machine learning is
\emph{model selection}~\citep{snoek12practicalBO,hutter2011smac}.
For instance, consider selecting the regularisation parameter $\lambda$
and kernel bandwidth $h$ for an SVM.
We can set this up as a zeroth order optimisation problem where our domain is a two
dimensional space of $(\lambda, h)$ values, and each function evaluation
trains the SVM on a training set, and computes the accuracy on a validation set.
The goal is to find the model, i.e. hyper-parameters, with the highest validation
accuracy.

The majority of the BO literature has focused on settings where the domain
$\Xcal$ is either Euclidean or categorical.
This
suffices for many tasks, such as the SVM example above. 
However, with
recent successes in deep learning, neural networks are increasingly becoming
the method of choice for many machine learning applications.
A number of recent work have designed novel neural network architectures
to significantly outperform the previous state of the
art~\citep{he2016deep,szegedy2015going,simonyan2014very,huang2017densely}.
This motivates studying model selection over
the space of neural architectures to optimise for generalisation performance.
A critical challenge in this endeavour is that evaluating a network via train
and validation procedures is very expensive.
This paper proposes a BO framework for this problem.


While there are several approaches to BO, those based on Gaussian processes (GP)
~\citep{rasmussen06gps} are most common in the BO literature.
In its most unadorned form, a BO algorithm operates sequentially, starting at time $0$
with a GP prior for $\func$; at time $t$, it incorporates results of evaluations
from $1,\dots,t-1$ in the form of a posterior for $\func$.
It then uses this posterior to construct an acquisition function $\acqt$,
where $\acqt(x)$ is a measure of the value of evaluating $\func$ at $x$ at time $t$
if our goal is to maximise $\func$.
Accordingly, it chooses to evaluate $\func$ at
the maximiser of the acquisition, i.e. $\xt = \argmax_{x\in\Xcal}\acqt(x)$.
There are two key ingredients to realising this plan for GP based BO.
First, we need to quantify the similarity between two points $x, x'$ in the domain
in the form of a kernel $\kernel(x, x')$.
The kernel is needed to define the GP, which
allows us to reason about an unevaluated value $\func(x')$ when we have
already evaluated $\func(x)$.
Secondly, we need a method to maximise $\acqt$.

These two steps are fairly straightforward in conventional domains.  For
example, in Euclidean spaces, we can use one of many popular kernels such as
Gaussian, Laplacian, or \matern; we can maximise $\acqt$ via off the shelf
branch-and-bound or gradient based methods.  However, when each $x\in\Xcal$ is
a neural network architecture, this is not the case.  Hence, our challenges in
this work are two-fold.  First, we need to \emph{quantify (dis)similarity
between two networks}.  Intuitively, in Fig.~\ref{fig:mainnnegs},
network~\ref{fig:mainnneg1} is more similar to network~\ref{fig:mainnneg2},
than it is to~\ref{fig:mainnneg3}.  Secondly, we need to be able to traverse
the space of such networks to \emph{optimise the acquisition function}.  Our
main contributions are as follows.
\vspace{-0.08in}
\begin{enumerate}[leftmargin=0.2in]
\item We develop a (pseudo-)distance for neural network
architectures called \nndists (Optimal Transport Metrics for 
Architectures of Neural Networks)
that can be computed efficiently via an optimal transport program.\hspace{-0.1in}
\vspace{-0.07in}

\item
We develop a BO framework for optimising functions on neural network
architectures called \nnbos (Neural Architecture Search with Bayesian Optimisation
and Optimal Transport).
This includes an evolutionary algorithm to
optimise the acquisition function.
\vspace{-0.07in}

\item 
Empirically, we demonstrate that \nnbos outperforms other baselines
on model selection tasks for
multi-layer perceptrons (MLP) and convolutional neural networks (CNN).
Our python implementations of \nndists and \nnbos
are available at
\href{https://github.com/kirthevasank/nasbot}{\small \incmtt{github.com/kirthevasank/nasbot}}.

\end{enumerate}

\textbf{Related Work:}
Recently, there has been a surge of interest in methods for neural architecture
search~\citep{
miikkulainen2017evolving,kitano1990designing,floreano2008neuroevolution,xie2017genetic,%
liu2017hierarchical,real2017large,stanley2002evolving,%
liu2017progressive,negrinho2017deeparchitect,cortes2016adanet,%
baker2016designing,zoph2016neural,zoph2017learning,zhong2017practical%
}.
We discuss them in detail in the Appendix due to space constraints. Broadly,
they fall into two categories, based on either evolutionary algorithms (\evoalg) or
reinforcement learning (RL).
\evoalgs provide a simple mechanism to explore
the space of architectures by making a sequence of changes to networks that
have already been evaluated.  However, as we will discuss later, they are not ideally
suited for optimising functions that are expensive to evaluate.
While RL methods have seen recent success,
architecture search is in essence an \emph{optimisation} problem --
find the network with the lowest validation error.
There is no explicit need to maintain a notion of state and solve credit
assignment~\citep{sutton1998reinforcement}.
Since RL is a fundamentally more difficult
problem than optimisation~\citep{jiang2016contextual},
these approaches
 need to try a very large number of architectures to find the optimum.
This is not desirable, especially in computationally constrained settings.

None of the above methods have been designed with a focus on the expense of evaluating
a neural network, with an emphasis on being judicious in selecting which architecture to
try next.
Bayesian optimisation (BO) uses introspective Bayesian models
to carefully determine future evaluations and is well suited for expensive evaluations.
BO usually consumes more computation to determine future points than other methods,
but this pays dividends when the evaluations are very expensive.
While there has been some work on BO for architecture search~\citep{
snoek12practicalBO,jenatton2017bayesian,mendoza2016towards,swersky2014raiders,%
bergstra2013making}, they have only been applied to optimise feed forward structures,
e.g. Fig.~\ref{fig:mainnneg1}, but not Figs.~\ref{fig:mainnneg2}, ~\ref{fig:mainnneg3}.
We compare \nnbos to one such method and demonstrate that feed forward structures
are inadequate for many problems.
%


\vspace{-0.10in}
\section{Set Up}
\label{sec:prelims}
\vspace{-0.10in}

Our goal is to maximise a function $\func$ defined on a
space $\Xcal$ of neural network architectures.
When we evaluate $\func$ at $x\in\Xcal$, we obtain a possibly noisy observation $y$
of $\func(x)$.
In the context of architecture search,
$\func$ is the performance on a validation set after $x$ is trained on the training set.
If $\xopt=\argmax_{\Xcal}\func(x)$ is the optimal architecture, and
$\xt$ is the architecture evaluated at time $t$, we want
$\func(\xopt) - \max_{t\leq n}\func(\xt)$ to vanish fast as the number of
evaluations $n\rightarrow \infty$.
We begin with a review of BO
and then present a graph theoretic formalism
for neural network architectures.

\vspace{-0.05in}
\subsection{A brief review of Gaussian Process based Bayesian Optimisation}
\vspace{-0.10in}
\label{sec:gpbointro}

A GP is a random process defined on some domain $\Xcal$, and is
characterised by a
mean function $\mu:\Xcal\rightarrow\RR$ and a
(covariance) kernel $\kernel:\Xcal^2\rightarrow\RR$.
Given $n$ observations $\Dcal_n = \{(x_i,y_i)\}_{i=1}^n$, where $\xtt{i}\in\Xcal,
\ytt{i} = \func(\xtt{i}) + \epsilon_i\in \RR$, and $\epsilon_i\sim\Ncal(0,\eta^2)$,
the posterior process $\func|\Dcal_n$ is also a GP with mean $\mutt{n}$ and covariance
$\kerneltt{n}$.
Denote $Y\in\RR^n$ with $Y_i=y_i$,
$k,k'\in\RR^n$ with $k_i =
\kernel(x,x_i),k'_i=\kernel(x',x_i)$,
and $K\in \RR^{n\times n}$ with
$K_{i,j} = \kernel(x_i,x_j)$.
Then, $\mutt{n},\kerneltt{n}$ can be computed via,
\begin{align*}
\hspace{-0.05in}
\mu_n(x) &= k^\top(K + \eta^2I)^{-1}Y, \hspace{0.35in}
\numberthis \label{eqn:gpPost}
\kernel_n(x,x') = \kernel(x,x') - k^\top(K + \eta^2I)^{-1}k'.
\hspace{0.05in}
\end{align*}
For more background on GPs,
we refer readers to~\citet{rasmussen06gps}.
When tasked with optimising a function $\func$ over a domain $\Xcal$, BO
models $\func$ as a sample from a GP.
At time $t$, we have already evaluated $\func$ at points
$\{\xtt{i}\}_{i=1}^{t-1}$ and obtained observations $\{\ytt{i}\}_{i=1}^{t-1}$.
To determine the next point for evaluation $\xt$,
we first use the posterior GP to define an \emph{acquisition function}
$\acqt:\Xcal\rightarrow\RR$,
which measures the utility of evaluating $\func$ at any $x\in\Xcal$ according
to the posterior.
We then maximise the acquisition $\xt=\argmax_\Xcal \acqt(x)$,
and evaluate $\func$ at $\xt$.
The expected improvement acquisition~\citep{mockus91bo},
\begin{align*}
\acqt(x) = \EE\big[\max\{0, \func(x) - \tau_{t-1}\} \big|\{(\xtt{i}, \ytt{i})\}_{i=1}^{t-1}
            \big],
\numberthis
\label{eqn:eiacq}
\end{align*}
measures the expected improvement over the current maximum value according to the
posterior GP.
Here $\tau_{t-1} = \argmax_{i\leq t-1} \func(\xtt{i})$ denotes the current best
value.
This expectation can be computed in closed form for GPs.
We use EI in this work, but the ideas
apply just as well to other acquisitions~\citep{brochu12bo}.



\textbf{GP/BO in the context of architecture search:}
Intuitively, $\kernel(x,x')$ is a measure of similarity between $x$ and $x'$.
If $\kernel(x,x')$ is large, then $\func(x)$ and $\func(x')$ are highly correlated.
Hence, the GP effectively imposes a smoothness condition on $\func:\Xcal\rightarrow\RR$;
i.e. since networks~${\rm \subref*{fig:mainnneg1}}$ and~${\rm \subref*{fig:mainnneg2}}$ in
Fig.~$\rm \ref{fig:mainnnegs}$ are similar, they are likely to have similar cross validation
performance.
In BO, when selecting the next point, we balance
between \emph{exploitation}, choosing points that we believe
will have high $\func$ value, and \emph{exploration}, choosing points
that we do not know much about so that we do not get stuck at a bad optimum.
For example, if we have already evaluated
$f({\rm \subref*{fig:mainnneg1}})$, then exploration incentivises us to 
choose~$\rm \subref*{fig:mainnneg3}$
over~$\rm \subref*{fig:mainnneg2}$ since we can reasonably
gauge $f({\rm \subref*{fig:mainnneg2}})$ from $f({\rm \subref*{fig:mainnneg1}})$.
On the other hand, if $f({\rm \subref*{fig:mainnneg1}})$ has high value, then exploitation
incentivises choosing ${\rm \subref*{fig:mainnneg2}}$,
as it is more likely to  be the optimum
than~${\rm \subref*{fig:mainnneg3}}$.

\subsection{A Mathematical Formalism for Neural Networks}
\label{sec:nngraphformalism}
\vspace{-0.05in}

\insertFigNNEgsMain

Our formalism will view a neural
network as a graph whose vertices are the layers of the network.
We will use the CNNs in Fig.~\ref{fig:mainnnegs} to illustrate the concepts.
%
A neural network $\Gcal = (\layers, \edges)$ is defined by a set of layers $\layers$
and directed edges $\edges$. An edge $(u,v)\in\edges$ is a ordered pair of layers.
In Fig.~\ref{fig:mainnnegs}, the layers are depicted by rectangles and the edges by arrows.
A layer $u\in\layers$ is equipped with a layer label $\laylabel(u)$
which denotes the type of operations performed at the layer.
For instance, in Fig.~\ref{fig:mainnneg1},
$\laylabel(1) = \textrm{\convthree},\, \laylabel(5) = \textrm{\maxpool}$
denote a $3\times 3$ convolution and
 a max-pooling operation.
The attribute $\layunits$ denotes the number of computational
units in a layer.
In Fig.~\ref{fig:mainnneg2},
$\layunits(5) = 32$ and $\layunits(7) = 16$
are the number of convolutional filters and fully connected nodes.

In addition, each network has \emph{decision layers} which are used to obtain the
predictions of the network.
For a classification task, the decision layers perform \softmax{} operations and output the
probabilities an input datum belongs to each class.
For regression, the decision layers perform \linear{} combinations of the outputs
of the previous layers and output a single scalar.
All networks have at least one decision layer.
When a network has multiple decision layers, we average the output of each decision
layer to obtain the final output.
The decision layers are shown in green in
Fig.~\ref{fig:mainnnegs}.
Finally, every network has a unique \emph{input layer} $\ipnode$
and \emph{output layer} $\opnode$ with labels
$\laylabel(\ipnode) = \textrm{\iplab}$ and $\laylabel(\opnode) = \textrm{\oplab}$.
It is instructive to think of the role of $\ipnode$ as feeding a data point to the
network and the role of $\opnode$ as averaging the
results of the decision layers.
The input and output layers are shown in pink in Fig.~\ref{fig:mainnnegs}.
We refer to all layers that are not input, output or decision layers as
\emph{processing layers}.

The directed edges are to be interpreted as follows.
The output of each layer is fed to each of its children; so both layers 2 and 3 in
Fig.~\ref{fig:mainnneg2} take the output of layer $1$ as input.
When a layer has multiple parents, the inputs are concatenated;
so layer 5 sees an input of $16+16$ filtered channels coming in from layers $3$ and $4$.
Finally, we mention that neural networks are also characterised by the values of the
weights/parameters between layers.
In architecture search, we typically do not consider these weights.
Instead, an algorithm will (somewhat ideally) assume  access to an optimisation
oracle that
can minimise the loss function on the training set and find the optimal weights.

We next describe a distance $d:\Xcal^2\rightarrow\RR_+$ for neural architectures.
Recall that our eventual goal is a kernel for the GP;
given a distance $d$, we will aim for $\kernel(x,x') =$ $e^{-\beta d(x,x')^p}$,
where $\beta,p\in\RR_+$,
as the kernel. %
Many popular kernels take this form.
For e.g. when $\Xcal\subset\RR^n$ and $d$ is the $L^2$ norm, $p=1,2$ correspond to the
Laplacian and Gaussian kernels respectively.%

\vspace{-0.05in}
\section{The \nndists{} Distance}
\label{sec:nndistmain}
\vspace{-0.10in}

To motivate this distance, note that 
the performance of a neural network is determined by the
amount of computation at each layer,
the types of these operations,
and how the layers are connected.
A meaningful distance  should account for these factors.
To that end, \nndists is defined as the minimum of a matching scheme
which attempts to match the computation at the layers of one network to
the layers of the other.
We incur penalties for matching layers with different types of
operations or those at structurally different positions.
We will find a matching that minimises these penalties, and the total
penalty at the minimum will give rise to a distance.
We first describe two concepts, layer masses and path lengths, which we will use
to define \nndist.

\textbf{Layer masses:}
The layer masses
$\laymass:\layers\rightarrow\RR_+$ will be the quantity that we match
between the layers of two networks when comparing them.
$\laymass(u)$ quantifies the significance of layer $u$.
For processing layers,
$\laymass(u)$ will represent the amount of computation carried out by layer $u$ and
is computed via the product of $\layunits(u)$ and the number of incoming units.
For example, in Fig.~\ref{fig:mainnneg2}, $\laymass(5)= 32 \times(16+16)$ as there are $16$
filtered channels each coming from layers $3$ and $4$ respectively.
As there is no computation at the input and output layers, we cannot define the layer
mass directly as we did for the processing layers.
Therefore, we use $\laymass(\ipnode) = \laymass(\opnode) = \dlmassfrac
\sum_{u\in\processlayers}\laymass(u)$ where
$\processlayers$ denotes the set of processing layers,
and $\dlmassfrac\in(0,1)$ is a parameter to be determined.
Intuitively, we are using an amount of mass that is proportional
to the amount of computation in the processing layers.
Similarly, the decision layers occupy a significant role in the architecture as they
directly influence the output.
While there is computation being performed at these layers, this might be problem
dependent -- there is more computation performed at the softmax layer in a 10 class
classification problem than in a 2 class problem.
Furthermore, we found that setting the layer mass for
decisions layers  based on computation underestimates their contribution
to the network.
Following the same intuition as we did for the input/output layers, we assign
an amount of mass proportional to the mass in the processing layers.
Since the outputs of the decision layers are averaged,
we distribute the mass among all decision layers;
that is, if $\decisionlayers$ are decision layers,
$\forall\,u\in\decisionlayers,
\laymass(u) = \frac{\dlmassfrac}{|\decisionlayers|} \sum_{u\in\processlayers}\laymass(u)$.
In all our experiments, we use $\dlmassfrac=0.1$.
In Fig.~\ref{fig:mainnnegs}, the layer masses for each layer are shown in
parantheses.

\textbf{Path lengths from/to {\normalfont $\ipnode$}/{\normalfont $\opnode$}:}
In a neural network $\Gcal$, a path from $u$ to $v$ is a sequence of layers
$u_1,\dots,u_s$ where $u_1=u$, $u_s=v$ and 
$(u_i,u_{i+1}) \in\edges$ for all $i\leq s-1$.
The length of this path is the number of hops from one node to another in order to get
from $u$ to $v$. 
For example, in Fig.~\ref{fig:mainnneg3},
$(2, 5, 8, 13)$ is a path from layer $2$ to $13$ of length $3$.
Let the shortest (longest) path length from $u$ to $v$ be the smallest (largest)
number of hops from one node to another among all paths from $u$ to $v$.
Additionally,
define the random walk path length as the expected number of hops to get from $u$ to
$v$, if, from any layer we hop to one of its children chosen uniformly at random.
For example, in Fig.~\ref{fig:mainnneg3}, the shortest, longest and random walk path lengths
from layer $1$ to layer $14$ are 5, 7, and 5.67 respectively.
%
%
For any $u\in\Lcal$, let $\distopsp(u), \distoplp(u), \distoprw(u)$ denote the length of
the shortest, longest and random walk paths from $u$ {to} the output $\opnode$.
Similarly, let $\distipsp(u), \distiplp(u), \distiprw(u)$ denote the corresponding
lengths for walks {from} the input $\ipnode$ to $u$.
As the layers of a neural network can be topologically ordered\footnote{%
A topological ordering is an ordering of the layers
$u_{1},\dots,u_{{|\Lcal|}}$ such that $u$ comes before $v$ if $(u,v) \in \edges$.
}, the above path lengths are well defined and finite.
Further, for any $s\in\{\textrm{sp,lp,rw}\}$ and
$t\in\{\textrm{ip,op}\}$, $\distpathlength_{t}^{s}(u)$
can   be computed for all $u\in\layers$, in 
$\bigO(|\edges|)$ time 
(see Appendix~\ref{app:nndistimplementation} for details).

We are now ready to describe \nndist.
Given two networks $\Gone=(\Lone,\Eone), \Gtwo=(\Ltwo,\Etwo)$ with $\none,\ntwo$ layers
respectively, we will attempt to match the layer masses in both networks.
We let $\otvar\in\RR_+^{\none\times\ntwo}$ be such that $\otvar(i,j)$ denotes the amount of
mass matched between layer $i\in\Gone$ and $j\in\Gtwo$.
The \nndists distance is computed by solving the following optimisation problem.
\begingroup
\allowdisplaybreaks
\begin{align*}
& \minimise_{\otvar}\quad \mismatchpen(\otvar)  + \nonmatchpen(\otvar) +  \structpencoeff\structpen(\otvar)
\numberthis
\label{eqn:nndistdefnmain}
\\
& \subto\;
\sum_{j\in\Ltwo} \otvar_{ij} \leq \laymass(i),\;
\sum_{i\in\Lone} \otvar_{ij} \leq \laymass(j),\; \forall i,j
\end{align*}
\endgroup
\vspace{-0.15in}

The label mismatch term $\mismatchpen$, penalises
matching masses that have different labels,
while the structural term $\structpen$ penalises matching masses 
at structurally different positions with respect to each other.
If we choose not to match any mass in either network, we incur a
non-assignment penalty $\nonmatchpen$.
$\structpencoeff > 0$
determines the trade-off between the structural and other terms.
The inequality constraints  ensure that
we do not over assign the masses in a layer.
We now describe $\mismatchpen,\nonmatchpen,$ and $\structpen$.

\insertLabelMismatchTableSmall

\emph{Label mismatch penalty \emph{$\mismatchpen$}}:
We begin with a label penalty matrix $\mislabmat\in\RR^{L\times L}$ where
$L$ is the number of all label types and
$\mislabmat(\textrm{\inlabelfont{x}},\textrm{\inlabelfont{y}})$ denotes the penalty for
transporting a unit
mass from a layer with label \inlabelfont{x} to a layer with label \inlabelfont{y}.
We then construct a matrix $\mislabmatprob\in\RR^{\none\times\ntwo}$ with
$\mislabmatprob(i,j) = \mislabmat(\laylabel(i), \laylabel(j))$ corresponding to
the mislabel cost for matching unit mass from each layer $i\in\Lone$ to each
layer $j\in\Ltwo$.
We then set
$\mismatchpen(\otvar) = \langle \otvar, \mislabmatprob \rangle = \sum_{i\in\Lone,j\in\Ltwo}
\otvar(i,j)C(i,j)$ to be the sum of all matchings from $\Lone$ to $\Ltwo$ weighted
by the label penalty terms.
This matrix $\mislabmat$, illustrated in Table~\ref{tb:mislabmatsmall}, is a parameter
that needs to be specified for \nndist.
They can be specified with an intuitive understanding
of the functionality of the layers;
e.g. many values in $\mislabmat$ are $\infty$, while for similar layers,
we choose a value less than $1$.

\emph{Non-assignment penalty \emph{$\nonmatchpen$}}:
We set this to be the amount of mass that is unassigned in both networks,
i.e. $\nonmatchpen(\otvar) =
\sum_{i\in\Lone} \big(\laymass(i) - \sum_{j\in\Ltwo} \otvar_{ij}\big)
+ \sum_{j\in\Ltwo}$ $\big(\laymass(j) - \sum_{i\in\Lone} \otvar_{ij}\big)$.
This essentially implies that the cost for not assigning unit mass is $1$.
The costs in Table~\ref{tb:mislabmatsmall} are defined relative to this.
For similar layers \inlabelfont{x}, \inlabelfont{y},
$\mislabmat(\textrm{\inlabelfont{x}},\textrm{\inlabelfont{y}}) \ll 1$
and for disparate layers 
$\mislabmat(\textrm{\inlabelfont{x}},\textrm{\inlabelfont{y}}) \gg 1$.
That is, we would rather match \convthree{} to \convfive{} than not assign it,
provided the structural penalty for doing so is small;
conversely, we would rather not assign a \convthree{}, than assign it to \fc{}.
This also explains why we did not use a trade-off parameter like $\structpencoeff$
for $\mismatchpen$ and $\nonmatchpen$ -- it is simple to specify reasonable
values for
 $\mislabmat(\textrm{\inlabelfont{x}},\textrm{\inlabelfont{y}})$
from an understanding of their functionality.

\emph{Structural penalty \emph{$\structpen$}}:
We define a matrix $\strmatprob\in\RR^{\none\times\ntwo}$ where
$\strmatprob(i,j)$ is small if layers $i\in\Lone$ and $j\in\Ltwo$ are at
structurally similar positions in their respective networks.
We then set $\structpen(\otvar) =
\langle \otvar, \strmatprob \rangle$.
For $i\in\Lone,\;j\in\Ltwo$, we let
$\strmatprob(i,j) = \frac{1}{6}\sum_{s\in\{\textrm{sp, lp, rw}\}}
\sum_{t\in\{\textrm{ip,op}\}} |\distpathlength^{s}_{t}(i) - 
\distpathlength^{s}_{t}(j)|$ be the
average of all path length differences, where $\distpathlength^s_t$ are the
path lengths
defined previously.
We define $\structpen$ in terms of
the shortest/longest/random-walk path lengths from/to the input/output,
because they capture various notions of
information flow in a neural network; a layer's input is influenced by
the paths the data takes before reaching the layer and its output influences
all layers it passes through before reaching the decision layers.
If the path lengths are similar for two layers, they are
likely to be at similar structural positions.
Further, this form allows us to
solve~\eqref{eqn:nndistdefnmain} efficiently via an OT program and prove distance
properties about the solution.
If we need to compute pairwise distances for several networks,
as is the case in BO,
the path lengths can be pre-computed in $\bigO(|\edges|)$ time,
and used to construct $\strmatprob$ for two networks at the moment
of computing the distance between them.

This completes the description of our matching program.
In Appendix~\ref{app:nndist}, we prove that~\eqref{eqn:nndistdefnmain} can be
formulated as an Optimal Transport (OT) program~\citep{villani2008optimal}.
OT is a well studied  problem with several efficient solvers~\citep{peyre2016ot}.
Our theorem below, 
shows that the solution of~\eqref{eqn:nndistdefnmain} is
 a distance.

\insertTableEAModifiersSmall

\vspace{0.05in}
\begin{theorem}
\label{thm:metric}
Let $d(\Gone,\Gtwo)$ be the solution of~\eqref{eqn:nndistdefnmain} for networks
$\Gone,\Gtwo$.
Under mild regularity conditions on $M$,
$d(\cdot,\cdot)$ is a pseudo-distance. That is, for all networks
$\Gone,\Gtwo,\Gthree$, it satisfies, $d(\Gone,\Gtwo) \geq 0$,
$d(\Gone,\Gtwo) = d(\Gtwo,\Gone)$,
$d(\Gone,\Gone) = 0$ and
$d(\Gone,\Gthree) \leq d(\Gone, \Gtwo) + d(\Gtwo, \Gthree)$.
\end{theorem}

For what follows, define $\dbar(\Gone,\Gtwo) = d(\Gone,\Gtwo)/(\totmass(\Gone) +
\totmass(\Gtwo))$
where $\totmass(\Gcal_i) = \sum_{u\in\Lcal_i}\laymass(u)$ is the total mass of a network.
Note that $\dbar \leq 1$.
While $\dbar$ does not satisfy the triangle inequality, it provides a
useful  measure of dissimilarity normalised by the amount of computation.
Our experience suggests that $d$ puts more emphasis on the amount of computation at the
layers over structure and vice versa for $\dbar$.
Therefore, it is prudent to combine both quantities in any downstream application.
The caption in Fig.~\ref{fig:mainnnegs} gives $d,\dbar$ values for the examples
in that figure when $\structpencoeff = 0.5$.

We conclude this section with a couple of remarks.  First, \nndists shares
similarities with Wasserstein (earth mover's) distances which also have an OT
formulation.  However, it is not a Wasserstein distance itself---in particular,
the supports of the masses and the cost matrices change depending on the two
networks being compared.
Second, while there has been prior work for defining various distances and
kernels on graphs, we cannot use them in BO because neural networks have
additional complex properties in addition to graphical structure, such as the
type of operations performed at each layer, the number of neurons, etc.  The
above work either define the distance/kernel between vertices or assume the
same vertex (layer)
set~\citep{messmer1998new,gao2010survey,%
wallis2001graph,%
smola2003kernels,kondor2002diffusion}, none of which apply in our setting.
While some methods do allow different vertex sets~\citep{vishwanathan2010graph},
they cannot handle layer masses and layer similarities.
Moreover, the computation of the above distances are more expensive than \nndist.
Hence, these methods cannot be directly plugged into  BO framework for architecture
search.

In Appendix~\ref{app:nndist}, we provide additional material on \nndist.
This includes the proof of Theorem~\ref{thm:metric}, a discussion on some design choices,
and implementation details such as the computation of the path lengths.
Moreover, we provide illustrations to demonstrate that \nndists is a meaningful distance
for architecture search.
For example, a t-SNE embedding places similar architectures
close to each other. 
Further, scatter plots showing the validation error vs distance on real datasets
demonstrate that networks with
small distance tend to perform similarly on the problem.


\vspace{-0.05in}
\section{\nnbo}
\label{sec:method}
\vspace{-0.10in}

We now describe \nnbo, our BO algorithm for neural architecture search.
Recall that in order to realise the BO scheme outlined in Section~\ref{sec:gpbointro},
we need to  specify (a) a kernel $\kernel$
for neural architectures and (b) a method to optimise
the acquisition $\acqt$ over these architectures.
Due to space constraints, we will only describe the key ideas and defer all details
to Appendix~\ref{app:implementation}.

As described previously, we will use a negative exponentiated distance for $\kernel$.
Precisely, $\kernel = \alpha e^{-\beta \dist} + \alphabar d^{-\betabar \dbar}$,
where $\dist,\dbar$ are the \nndists distance and its normalised version.
We mention that while this has the form of popular
kernels, we do not know yet if it is in fact a kernel.
In our experiments, we did not encounter an instance where the eigenvalues of
the kernel matrix were negative.
In any case, there are several methods to circumvent
this issue in kernel methods~\citep{sutherland2015scalable}.

We use an evolutionary algorithm (\evoalg) approach to optimise the
acquisition function~\eqref{eqn:eiacq}.
For this, we begin with an initial pool of networks and evaluate the acquisition
$\acqt$ on those networks.
Then we generate a set of $\Nmut$ mutations of this pool as follows.
First, we stochastically select $\Nmut$ candidates from the set of networks
already evaluated such that those with higher $\acqt$ values are more likely to be
selected than those with lower values.
Then we modify each candidate, to produce a new architecture.
These modifications, described in Table~\ref{tb:nnmodifierssmall},
might change the architecture either by increasing or decreasing the number of computational
units in a layer, by adding or deleting layers, or by changing the connectivity of
existing layers.
Finally, we evaluate the acquisition on this $\Nmut$ mutations, add it to the initial pool,
and repeat for the prescribed number of steps.
While \evoalgs works fine for cheap functions, such as the acquisition $\acqt$ which
is analytically available,
it is not suitable when evaluations are expensive, such as 
training a neural network.
This is because \evoalgs selects points for future evaluations that are already close
to points that have been evaluated, and is hence inefficient at exploring the
space.
In our experiments, we compare \nnbos to the same \evoalgs scheme used to optimise the
acquisition and demonstrate
the former outperforms the latter.

We conclude this section by observing that this framework for \nnbo/\nndists
has additional flexibility to what has been described.
If one wishes to tune over drop-out probabilities, regularisation
penalties and batch normalisation at each layer,  they can be treated as part
of the layer label, via an augmented label penalty matrix $M$
which accounts for these considerations.  If one wishes to jointly
tune other scalar hyper-parameters (e.g. learning rate), they can use an
existing kernel for euclidean spaces and define the GP over the joint
architecture + hyper-parameter space via a product kernel.  BO methods for
early stopping in iterative training
procedures~\citep{kandasamy2017boca,klein2016fast,kandasamy2016multigp,%
kandasamy2016gaussian,kandasamy2016multi}
can be easily
incorporated by defining a \emph{fidelity space}.  Using a line of
work in scalable GPs~\citep{wilson2015kernel,snelson2006sparse}, one can apply
our methods to challenging problems which might require trying a very large
number ($\sim$100K) of architectures.  These extensions will enable deploying
\nnbos in large scale settings, but are tangential to our goal of introducing a
BO method for architecture search.

\vspace{-0.05in}
\section{Experiments}
\label{sec:experiments}
\vspace{-0.10in}

\insertFigMLP

\textbf{Methods:}
We compare \nnbos to the following baselines.
\rand: random search;
\evoalgs (Evolutionary algorithm):
the same \evoalgs procedure described above.
\treebos~\citep{jenatton2017bayesian}: a BO method which only searches over feed forward
structures.
Random search is a natural baseline to compare optimisation methods.
However,
unlike in Euclidean spaces, there is no natural way to randomly explore
the space of architectures.
Our \rands implementation, operates in exactly the same way as \nnbo,
except that the \evoalgs procedure 
is fed a random sample from $\unif(0,1)$ instead of
the GP acquisition each time it evaluates an architecture.
Hence, \rands is effectively picking a random network from the same space 
explored by \nnbo;
neither method has an unfair advantage because it considers a different space.
While there are other methods for architecture search, their implementations are highly
nontrivial and are not made available.



\textbf{Datasets:}
We use the following datasets:
blog feedback~\citep{buza2014feedback},
indoor location~\citep{torres2014ujiindoorloc},
slice localisation~\citep{graf20112d},
naval propulsion~\citep{coraddu2016machine},
protein tertiary structure~\citep{rana2013physicochemical},
news popularity~\citep{fernandes2015proactive},
Cifar10~\citep{krizhevsky2009cifar}.
The first six%
{} are regression problems for which we use MLPs.
The last is a classification task on images for which we use CNNs.
Table~\ref{tb:testresults} gives the size and dimensionality of each dataset.
For the first $6$ datasets, we use a $0.6-0.2-0.2$ train-validation-test split
and normalised the input and output to have
zero mean and unit variance.
Hence, a constant predictor will have a mean squared error of approximately $1$.
For Cifar10 we use $40K$ for training and $10K$ each for validation and testing.

\textbf{Experimental Set up:}
Each method is executed in an asynchronously parallel set up of 2-4 GPUs,
That is, it can evaluate multiple models in parallel, with each model on a single GPU.
When the evaluation of one model finishes, the methods can incorporate the result
and immediately re-deploy
the next job without waiting for the others to finish.
For the blog, indoor, slice, naval and protein datasets we use
2 GeForce GTX 970 (4GB) GPUs and a computational budget of 8 hours
for each method.
For the news popularity dataset we use $4$  GeForce GTX 980 (6GB) GPUs with a
budget of 6 hours and for Cifar10 we use 4 K80 (12GB) GPUs
with a budget of 10 hours.
For the regression datasets, we train each model with stochastic gradient descent (SGD)
with a fixed step size of $10^{-5}$, a batch size of 256 for 20K batch iterations.
For Cifar10, we start with a step size of $10^{-2}$, and reduce it gradually.
We train in batches of 32 images for 60K batch iterations.
The methods evaluate between $70$-$120$ networks depending
on the size of the networks chosen and the number of GPUs.

\insertTestResultsTableHor

\textbf{Results:}
Fig.~\ref{fig:modsel} plots the best validation score for each method against
time.  In Table~\ref{tb:testresults}, we present the results on the test set
with the best model chosen on the basis of validation set performance.
On the Cifar10 dataset, we also trained the best models for longer ($150K$
iterations). These results are in the last column of Table~\ref{tb:testresults}.
We see that \nnbos is the most consistent of all methods.
The average time taken by \nnbos to determine the next architecture to evaluate was
$46.13$s.
For \rand, \evoalg, and \treebos this was
$26.43$s, $0.19$s, and $7.83$s respectively.
The time taken to train and validate models was on the order of 10-40
minutes depending on the model size.
Fig.~\ref{fig:modsel} includes this time taken to determine the next point.
Like many BO algorithms, while \nnbo's selection criterion is time consuming,
it pays off when evaluations are expensive.
In Appendices~\ref{app:implementation} and~\ref{app:experiments}, we provide
additional details on the experiment set up and conduct
synthetic ablation studies by
holding out
different components of the \nnbos framework.  We also illustrate some of the
best architectures found---on many datasets, common features were long skip
connections and multiple decision layers.

Finally, we note that while our Cifar10 experiments fall short of the current
state of the art~\citep{zoph2016neural,liu2017progressive,liu2017hierarchical},
the amount of computation in these work is several orders of magnitude more
than ours (both the computation invested to train a single model and the number
of models trained).
Further, they use constrained spaces specialised for CNNs, while \nnbos is
deployed in a very general model space.
We believe that our results can also be improved by
employing enhanced training techniques such as image whitening, image flipping,
drop out, etc. For example, using our training procedure on the VGG-19
architecture~\citep{simonyan2014very} yielded a test set error of
\textcolor{black}{$0.1018$ after $150K$ iterations}. However, VGG-19 is known
to do significantly better on Cifar10.
That said, we believe our results are encouraging and lay out the premise for
BO for neural architectures.


\vspace{-0.1in}
\section{Conclusion}
\label{sec:conclusion}
\vspace{-0.10in}


We described \nnbo, a BO framework for neural architecture search.
\nnbos finds better architectures for MLPs and CNNs  more efficiently
than other baselines on several datasets.
A key contribution of this work is the efficiently computable \nndists distance for
neural network architectures, which
may be of independent interest as it might find applications outside of BO.
Our code for \nnbos and \nndists will be made available.

\subsection*{Acknolwedgements}
We would like to thank Guru Guruganesh and Dougal Sutherland for the insightful
discussions.
This research is partly funded by DOE grant DESC0011114, NSF grant
IIS1563887, and the Darpa D3M program.
KK is supported by a Facebook fellowship and a Siebel scholarship.

{\small
\renewcommand{\bibsection}{\section*{References\vspace{-0.1em}} }
\setlength{\bibsep}{1.1pt}
\bibliography{kky,koopmans}

\begin{thebibliography}{54}
\providecommand{\natexlab}[1]{#1}
\providecommand{\url}[1]{\texttt{#1}}
\expandafter\ifx\csname urlstyle\endcsname\relax
  \providecommand{\doi}[1]{doi: #1}\else
  \providecommand{\doi}{doi: \begingroup \urlstyle{rm}\Url}\fi

\bibitem[Baker et~al.(2016)Baker, Gupta, Naik, and Raskar]{baker2016designing}
Bowen Baker, Otkrist Gupta, Nikhil Naik, and Ramesh Raskar.
\newblock Designing neural network architectures using reinforcement learning.
\newblock \emph{arXiv preprint arXiv:1611.02167}, 2016.

\bibitem[Bergstra et~al.(2013)Bergstra, Yamins, and Cox]{bergstra2013making}
James Bergstra, Daniel Yamins, and David~Daniel Cox.
\newblock Making a science of model search: Hyperparameter optimization in
  hundreds of dimensions for vision architectures.
\newblock 2013.

\bibitem[Brochu et~al.(2010)Brochu, Cora, and de~Freitas]{brochu12bo}
Eric Brochu, Vlad~M. Cora, and Nando de~Freitas.
\newblock {A Tutorial on Bayesian Optimization of Expensive Cost Functions,
  with Application to Active User Modeling and Hierarchical Reinforcement
  Learning}.
\newblock \emph{CoRR}, 2010.

\bibitem[Buza(2014)]{buza2014feedback}
Krisztian Buza.
\newblock Feedback prediction for blogs.
\newblock In \emph{Data analysis, machine learning and knowledge discovery},
  pages 145--152. Springer, 2014.

\bibitem[Coraddu et~al.(2016)Coraddu, Oneto, Ghio, Savio, Anguita, and
  Figari]{coraddu2016machine}
Andrea Coraddu, Luca Oneto, Aessandro Ghio, Stefano Savio, Davide Anguita, and
  Massimo Figari.
\newblock Machine learning approaches for improving condition-based maintenance
  of naval propulsion plants.
\newblock \emph{Proceedings of the Institution of Mechanical Engineers, Part M:
  Journal of Engineering for the Maritime Environment}, 230\penalty0
  (1):\penalty0 136--153, 2016.

\bibitem[Cortes et~al.(2016)Cortes, Gonzalvo, Kuznetsov, Mohri, and
  Yang]{cortes2016adanet}
Corinna Cortes, Xavi Gonzalvo, Vitaly Kuznetsov, Mehryar Mohri, and Scott Yang.
\newblock Adanet: Adaptive structural learning of artificial neural networks.
\newblock \emph{arXiv preprint arXiv:1607.01097}, 2016.

\bibitem[Fernandes et~al.(2015)Fernandes, Vinagre, and
  Cortez]{fernandes2015proactive}
Kelwin Fernandes, Pedro Vinagre, and Paulo Cortez.
\newblock A proactive intelligent decision support system for predicting the
  popularity of online news.
\newblock In \emph{Portuguese Conference on Artificial Intelligence}, 2015.

\bibitem[Floreano et~al.(2008)Floreano, D{\"u}rr, and
  Mattiussi]{floreano2008neuroevolution}
Dario Floreano, Peter D{\"u}rr, and Claudio Mattiussi.
\newblock Neuroevolution: from architectures to learning.
\newblock \emph{Evolutionary Intelligence}, 1\penalty0 (1):\penalty0 47--62,
  2008.

\bibitem[Gao et~al.(2010)Gao, Xiao, Tao, and Li]{gao2010survey}
Xinbo Gao, Bing Xiao, Dacheng Tao, and Xuelong Li.
\newblock A survey of graph edit distance.
\newblock \emph{Pattern Analysis and applications}, 13\penalty0 (1):\penalty0
  113--129, 2010.

\bibitem[Ginsbourger et~al.(2011)Ginsbourger, Janusevskis, and
  Le~Riche]{ginsbourger2011dealing}
David Ginsbourger, Janis Janusevskis, and Rodolphe Le~Riche.
\newblock Dealing with asynchronicity in parallel gaussian process based global
  optimization.
\newblock In \emph{ERCIM}, 2011.

\bibitem[Graf et~al.(2011)Graf, Kriegel, Schubert, P{\"o}lsterl, and
  Cavallaro]{graf20112d}
Franz Graf, Hans-Peter Kriegel, Matthias Schubert, Sebastian P{\"o}lsterl, and
  Alexander Cavallaro.
\newblock 2d image registration in ct images using radial image descriptors.
\newblock In \emph{International Conference on Medical Image Computing and
  Computer-Assisted Intervention}, pages 607--614. Springer, 2011.

\bibitem[He et~al.(2016)He, Zhang, Ren, and Sun]{he2016deep}
Kaiming He, Xiangyu Zhang, Shaoqing Ren, and Jian Sun.
\newblock Deep residual learning for image recognition.
\newblock In \emph{Proceedings of the IEEE conference on computer vision and
  pattern recognition}, pages 770--778, 2016.

\bibitem[Huang et~al.(2017)Huang, Liu, Weinberger, and van~der
  Maaten]{huang2017densely}
Gao Huang, Zhuang Liu, Kilian~Q Weinberger, and Laurens van~der Maaten.
\newblock Densely connected convolutional networks.
\newblock In \emph{CVPR}, 2017.

\bibitem[Hutter et~al.(2011)Hutter, Hoos, and Leyton-Brown]{hutter2011smac}
Frank Hutter, Holger~H Hoos, and Kevin Leyton-Brown.
\newblock Sequential model-based optimization for general algorithm
  configuration.
\newblock In \emph{LION}, 2011.

\bibitem[Jenatton et~al.(2017)Jenatton, Archambeau, Gonz{\'a}lez, and
  Seeger]{jenatton2017bayesian}
Rodolphe Jenatton, Cedric Archambeau, Javier Gonz{\'a}lez, and Matthias Seeger.
\newblock Bayesian optimization with tree-structured dependencies.
\newblock In \emph{International Conference on Machine Learning}, 2017.

\bibitem[Jiang et~al.(2016)Jiang, Krishnamurthy, Agarwal, Langford, and
  Schapire]{jiang2016contextual}
Nan Jiang, Akshay Krishnamurthy, Alekh Agarwal, John Langford, and Robert~E
  Schapire.
\newblock Contextual decision processes with low bellman rank are
  pac-learnable.
\newblock \emph{arXiv preprint arXiv:1610.09512}, 2016.

\bibitem[Kandasamy et~al.(2016{\natexlab{a}})Kandasamy, Dasarathy, Oliva,
  Schneider, and P{\'o}czos]{kandasamy2016gaussian}
Kirthevasan Kandasamy, Gautam Dasarathy, Junier~B Oliva, Jeff Schneider, and
  Barnab{\'a}s P{\'o}czos.
\newblock Gaussian process bandit optimisation with multi-fidelity evaluations.
\newblock In \emph{Advances in Neural Information Processing Systems}, pages
  992--1000, 2016{\natexlab{a}}.

\bibitem[Kandasamy et~al.(2016{\natexlab{b}})Kandasamy, Dasarathy, Oliva,
  Schneider, and Poczos]{kandasamy2016multigp}
Kirthevasan Kandasamy, Gautam Dasarathy, Junier~B Oliva, Jeff Schneider, and
  Barnabas Poczos.
\newblock Multi-fidelity gaussian process bandit optimisation.
\newblock \emph{arXiv preprint arXiv:1603.06288}, 2016{\natexlab{b}}.

\bibitem[Kandasamy et~al.(2016{\natexlab{c}})Kandasamy, Dasarathy, Poczos, and
  Schneider]{kandasamy2016multi}
Kirthevasan Kandasamy, Gautam Dasarathy, Barnabas Poczos, and Jeff Schneider.
\newblock The multi-fidelity multi-armed bandit.
\newblock In \emph{Advances in Neural Information Processing Systems}, pages
  1777--1785, 2016{\natexlab{c}}.

\bibitem[Kandasamy et~al.(2017)Kandasamy, Dasarathy, Schneider, and
  Poczos]{kandasamy2017boca}
Kirthevasan Kandasamy, Gautam Dasarathy, Jeff Schneider, and Barnabas Poczos.
\newblock {Multi-fidelity Bayesian Optimisation with Continuous
  Approximations}.
\newblock \emph{arXiv preprint arXiv:1703.06240}, 2017.

\bibitem[Kitano(1990)]{kitano1990designing}
Hiroaki Kitano.
\newblock Designing neural networks using genetic algorithms with graph
  generation system.
\newblock \emph{Complex systems}, 4\penalty0 (4):\penalty0 461--476, 1990.

\bibitem[Klein et~al.(2016)Klein, Falkner, Bartels, Hennig, and
  Hutter]{klein2016fast}
Aaron Klein, Stefan Falkner, Simon Bartels, Philipp Hennig, and Frank Hutter.
\newblock Fast bayesian optimization of machine learning hyperparameters on
  large datasets.
\newblock \emph{arXiv preprint arXiv:1605.07079}, 2016.

\bibitem[Kondor and Lafferty(2002)]{kondor2002diffusion}
Risi~Imre Kondor and John Lafferty.
\newblock Diffusion kernels on graphs and other discrete input spaces.
\newblock In \emph{ICML}, volume~2, pages 315--322, 2002.

\bibitem[Krizhevsky and Hinton(2009)]{krizhevsky2009cifar}
Alex Krizhevsky and Geoffrey Hinton.
\newblock Learning multiple layers of features from tiny images, 2009.

\bibitem[Liu et~al.(2017{\natexlab{a}})Liu, Zoph, Shlens, Hua, Li, Fei-Fei,
  Yuille, Huang, and Murphy]{liu2017progressive}
Chenxi Liu, Barret Zoph, Jonathon Shlens, Wei Hua, Li-Jia Li, Li~Fei-Fei, Alan
  Yuille, Jonathan Huang, and Kevin Murphy.
\newblock Progressive neural architecture search.
\newblock \emph{arXiv preprint arXiv:1712.00559}, 2017{\natexlab{a}}.

\bibitem[Liu et~al.(2017{\natexlab{b}})Liu, Simonyan, Vinyals, Fernando, and
  Kavukcuoglu]{liu2017hierarchical}
Hanxiao Liu, Karen Simonyan, Oriol Vinyals, Chrisantha Fernando, and Koray
  Kavukcuoglu.
\newblock Hierarchical representations for efficient architecture search.
\newblock \emph{arXiv preprint arXiv:1711.00436}, 2017{\natexlab{b}}.

\bibitem[Maaten and Hinton(2008)]{maaten2008visualizing}
Laurens van~der Maaten and Geoffrey Hinton.
\newblock Visualizing data using t-sne.
\newblock \emph{Journal of machine learning research}, 9\penalty0
  (Nov):\penalty0 2579--2605, 2008.

\bibitem[Mendoza et~al.(2016)Mendoza, Klein, Feurer, Springenberg, and
  Hutter]{mendoza2016towards}
Hector Mendoza, Aaron Klein, Matthias Feurer, Jost~Tobias Springenberg, and
  Frank Hutter.
\newblock Towards automatically-tuned neural networks.
\newblock In \emph{Workshop on Automatic Machine Learning}, pages 58--65, 2016.

\bibitem[Messmer and Bunke(1998)]{messmer1998new}
Bruno~T Messmer and Horst Bunke.
\newblock A new algorithm for error-tolerant subgraph isomorphism detection.
\newblock \emph{IEEE Transactions on Pattern Analysis and Machine
  Intelligence}, 20\penalty0 (5):\penalty0 493--504, 1998.

\bibitem[Miikkulainen et~al.(2017)Miikkulainen, Liang, Meyerson, Rawal, Fink,
  Francon, Raju, Navruzyan, Duffy, and Hodjat]{miikkulainen2017evolving}
Risto Miikkulainen, Jason Liang, Elliot Meyerson, Aditya Rawal, Dan Fink,
  Olivier Francon, Bala Raju, Arshak Navruzyan, Nigel Duffy, and Babak Hodjat.
\newblock Evolving deep neural networks.
\newblock \emph{arXiv preprint arXiv:1703.00548}, 2017.

\bibitem[Mockus and Mockus(1991)]{mockus91bo}
J.B. Mockus and L.J. Mockus.
\newblock {Bayesian approach to global optimization and application to
  multiobjective and constrained problems}.
\newblock \emph{Journal of Optimization Theory and Applications}, 1991.

\bibitem[Negrinho and Gordon(2017)]{negrinho2017deeparchitect}
Renato Negrinho and Geoff Gordon.
\newblock Deeparchitect: Automatically designing and training deep
  architectures.
\newblock \emph{arXiv preprint arXiv:1704.08792}, 2017.

\bibitem[Peyr\'e and Cuturi(2017)]{peyre2016ot}
Gabriel Peyr\'e and Marco Cuturi.
\newblock \emph{{Computational Optimal Transport}}.
\newblock Available online, 2017.

\bibitem[Rana(2013)]{rana2013physicochemical}
PS~Rana.
\newblock Physicochemical properties of protein tertiary structure data set,
  2013.

\bibitem[Rasmussen and Williams(2006)]{rasmussen06gps}
C.E. Rasmussen and C.K.I. Williams.
\newblock \emph{{Gaussian Processes for Machine Learning}}.
\newblock Adaptative computation and machine learning series. University Press
  Group Limited, 2006.

\bibitem[Real et~al.(2017)Real, Moore, Selle, Saxena, Suematsu, Le, and
  Kurakin]{real2017large}
Esteban Real, Sherry Moore, Andrew Selle, Saurabh Saxena, Yutaka~Leon Suematsu,
  Quoc Le, and Alex Kurakin.
\newblock Large-scale evolution of image classifiers.
\newblock \emph{arXiv preprint arXiv:1703.01041}, 2017.

\bibitem[Simonyan and Zisserman(2014)]{simonyan2014very}
Karen Simonyan and Andrew Zisserman.
\newblock Very deep convolutional networks for large-scale image recognition.
\newblock \emph{arXiv preprint arXiv:1409.1556}, 2014.

\bibitem[Smola and Kondor(2003)]{smola2003kernels}
Alexander~J Smola and Risi Kondor.
\newblock Kernels and regularization on graphs.
\newblock In \emph{Learning theory and kernel machines}, pages 144--158.
  Springer, 2003.

\bibitem[Snelson and Ghahramani(2006)]{snelson2006sparse}
Edward Snelson and Zoubin Ghahramani.
\newblock Sparse gaussian processes using pseudo-inputs.
\newblock In \emph{Advances in neural information processing systems}, pages
  1257--1264, 2006.

\bibitem[Snoek et~al.(2012)Snoek, Larochelle, and Adams]{snoek12practicalBO}
Jasper Snoek, Hugo Larochelle, and Ryan~P Adams.
\newblock {Practical Bayesian Optimization of Machine Learning Algorithms}.
\newblock In \emph{Advances in Neural Information Processing Systems}, 2012.

\bibitem[Stanley and Miikkulainen(2002)]{stanley2002evolving}
Kenneth~O Stanley and Risto Miikkulainen.
\newblock Evolving neural networks through augmenting topologies.
\newblock \emph{Evolutionary computation}, 10\penalty0 (2):\penalty0 99--127,
  2002.

\bibitem[Sutherland(2015)]{sutherland2015scalable}
Dougal~J Sutherland.
\newblock \emph{{Scalable, Active and Flexible Learning on Distributions}}.
\newblock PhD thesis, Carnegie Mellon University Pittsburgh, PA, 2015.

\bibitem[Sutton and Barto(1998)]{sutton1998reinforcement}
Richard~S Sutton and Andrew~G Barto.
\newblock \emph{Reinforcement learning: An introduction}, volume~1.
\newblock MIT press Cambridge, 1998.

\bibitem[Swersky et~al.(2014)Swersky, Duvenaud, Snoek, Hutter, and
  Osborne]{swersky2014raiders}
Kevin Swersky, David Duvenaud, Jasper Snoek, Frank Hutter, and Michael~A
  Osborne.
\newblock Raiders of the lost architecture: Kernels for bayesian optimization
  in conditional parameter spaces.
\newblock \emph{arXiv preprint arXiv:1409.4011}, 2014.

\bibitem[Szegedy et~al.(2015)Szegedy, Liu, Jia, Sermanet, Reed, Anguelov,
  Erhan, Vanhoucke, and Rabinovich]{szegedy2015going}
Christian Szegedy, Wei Liu, Yangqing Jia, Pierre Sermanet, Scott Reed, Dragomir
  Anguelov, Dumitru Erhan, Vincent Vanhoucke, and Andrew Rabinovich.
\newblock Going deeper with convolutions.
\newblock In \emph{Proceedings of the IEEE conference on computer vision and
  pattern recognition}, pages 1--9, 2015.

\bibitem[Torres-Sospedra et~al.(2014)Torres-Sospedra, Montoliu,
  Mart{\'\i}nez-Us{\'o}, Avariento, Arnau, Benedito-Bordonau, and
  Huerta]{torres2014ujiindoorloc}
Joaqu{\'\i}n Torres-Sospedra, Ra{\'u}l Montoliu, Adolfo Mart{\'\i}nez-Us{\'o},
  Joan~P Avariento, Tom{\'a}s~J Arnau, Mauri Benedito-Bordonau, and
  Joaqu{\'\i}n Huerta.
\newblock Ujiindoorloc: A new multi-building and multi-floor database for wlan
  fingerprint-based indoor localization problems.
\newblock In \emph{Indoor Positioning and Indoor Navigation (IPIN), 2014
  International Conference on}, pages 261--270. IEEE, 2014.

\bibitem[Villani(2008)]{villani2008optimal}
C{\'e}dric Villani.
\newblock \emph{Optimal transport: old and new}, volume 338.
\newblock Springer Science \& Business Media, 2008.

\bibitem[Vishwanathan et~al.(2010)Vishwanathan, Schraudolph, Kondor, and
  Borgwardt]{vishwanathan2010graph}
S~Vichy~N Vishwanathan, Nicol~N Schraudolph, Risi Kondor, and Karsten~M
  Borgwardt.
\newblock Graph kernels.
\newblock \emph{Journal of Machine Learning Research}, 11\penalty0
  (Apr):\penalty0 1201--1242, 2010.

\bibitem[Wallis et~al.(2001)Wallis, Shoubridge, Kraetz, and
  Ray]{wallis2001graph}
Walter~D Wallis, Peter Shoubridge, M~Kraetz, and D~Ray.
\newblock Graph distances using graph union.
\newblock \emph{Pattern Recognition Letters}, 22\penalty0 (6-7):\penalty0
  701--704, 2001.

\bibitem[Wilson and Nickisch(2015)]{wilson2015kernel}
Andrew Wilson and Hannes Nickisch.
\newblock Kernel interpolation for scalable structured gaussian processes
  (kiss-gp).
\newblock In \emph{International Conference on Machine Learning}, pages
  1775--1784, 2015.

\bibitem[Xie and Yuille(2017)]{xie2017genetic}
Lingxi Xie and Alan Yuille.
\newblock Genetic cnn.
\newblock \emph{arXiv preprint arXiv:1703.01513}, 2017.

\bibitem[Zhong et~al.(2017)Zhong, Yan, and Liu]{zhong2017practical}
Zhao Zhong, Junjie Yan, and Cheng-Lin Liu.
\newblock Practical network blocks design with q-learning.
\newblock \emph{arXiv preprint arXiv:1708.05552}, 2017.

\bibitem[Zoph and Le(2016)]{zoph2016neural}
Barret Zoph and Quoc~V Le.
\newblock Neural architecture search with reinforcement learning.
\newblock \emph{arXiv preprint arXiv:1611.01578}, 2016.

\bibitem[Zoph et~al.(2017)Zoph, Vasudevan, Shlens, and Le]{zoph2017learning}
Barret Zoph, Vijay Vasudevan, Jonathon Shlens, and Quoc~V Le.
\newblock Learning transferable architectures for scalable image recognition.
\newblock \emph{arXiv preprint arXiv:1707.07012}, 2017.

\end{thebibliography}
}
\bibliographystyle{plainnat}

\newpage

\appendix

\section{Additional Details on \nndist}
\label{app:nndist}

\subsection{Optimal Transport Reformulation}

We begin with a review optimal transport.
Throughout this section,
$\langle \cdot,\cdot\rangle$ denotes the Frobenius dot product.
$\one_n,\zero_n\in\RR^n$ denote a vector of ones and zeros respectively.

\textbf{A review of Optimal Transport}%
~\citep{villani2008optimal}\textbf{:}
Let $y_1\in\RR_+^{\none}, y_2\in\RR_+^{\ntwo}$ be such that
$\one_{\none}^\top y_1 = \one_{\ntwo}^\top y_2$.
Let $C\in\RR_+^{\none\times\ntwo}$.
The following optimisation problem,
\begin{align*}
& \minimise_{\otvar}\quad\;\; \langle \otvar, C \rangle  
\numberthis
\label{eqn:otdefn}
\\
& \subto \quad \otvar>0,\;\; \otvar\one_{\ntwo} = y_1, \;\; \otvar^\top\one_{\none} = y_2.
\end{align*}
is called an \emph{optimal transport} program.
One interpretation of this set up is that $y_1$ denotes the supplies at $n_1$ warehouses,
$y_2$ denotes the demands at $n_2$ retail stores,
$C_{ij}$ denotes the cost of transporting a unit mass of supplies from warehouse $i$ to
store $j$ and $\otvar_{ij}$ denotes the mass of material transported from $i$ to $j$.
The program attempts to find transportation plan which minimises the total
cost of transportation $\langle \otvar, C \rangle$.

\textbf{OT formulation of~\eqref{eqn:nndistdefnmain}:}
We now describe the OT formulation of the \nndists distance.
In addition to providing an efficient way to solve~\eqref{eqn:nndistdefnmain},
the OT formulation will allow us to prove the metric properties of the solution.
When computing the distance between $\Gone,\Gtwo$, for $i=1,2$,
let $\totmass(\Gcal_i) = \sum_{u\in\layers_i}\laymass(u)$ denote the total mass in
$\Gcal_i$, and $\nbarii{i} = n_i + 1$ where $n_i = |\layers_i|$.
$y_1 = [\{\laymass(u)\}_{u\in\Lone}, \totmass(\Gtwo)] \in \RR^{\nbarone}$ will be
the supplies in our OT problem,
and $y_2 = [\{\laymass(u)\}_{u\in\Ltwo}, \totmass(\Gone)] \in \RR^{\nbartwo}$ will
be the demands.
To define the cost matrix, we augment the mislabel and structural penalty matrices
$\mislabmatprob, \strmatprob$ with an additional row and column of zeros;
i.e. $\mislabmatot = [\mislabmatprob\, \zero_{n_1}; \zero_{\nbartwo}^\top]
\in\RR^{\nbarone\times\nbartwo}$; $\strmatot$ is defined similarly.
Let $\nasmatot = [\zero_{n_1,n_2}\,\one_{n_1}; \one_{n_2}^\top\, 0] \in
\RR^{\nbarone\times\nbartwo}$.
We will show that~\eqref{eqn:nndistdefnmain} is equivalent to the following OT program.
\begin{align*}
& \minimise_{\otvarp}\quad \langle \otvarp, C' \rangle 
\label{eqn:nndistdefntwo}
\numberthis  \\
& \subto \quad \otvarp\one_{\nbartwo} = y_1, \quad \otvarp^\top\one_{\nbarone} = y_2.
\end{align*}
One interpretation of~\eqref{eqn:nndistdefntwo} is that the last row/column appended to
the cost matrices serve as a non-assignment layer and that the cost for transporting
unit mass to this layer from all other layers is $1$.
The costs for mislabelling was defined relative to this non-assignment cost.
The costs for similar layers is much smaller than $1$;
therefore, the optimiser is incentivised
to transport mass among similar layers rather than not assign it provided that
the structural penalty is not too large.
Correspondingly, the cost for very disparate layers is much larger so that we would
never match, say, a convolutional layer with a pooling layer.
In fact, the $\infty$'s in Table~\ref{tb:mislabmatsmall} can be replaced by any value larger
than $2$ and the solution will be the same.
The following theorem shows that~\eqref{eqn:nndistdefnmain} and~\eqref{eqn:nndistdefntwo}
are equivalent.

\vspace{0.1in}
\begin{theorem}
Problems~\eqref{eqn:nndistdefnmain} and~\eqref{eqn:nndistdefntwo} are equivalent, in
that they both have the same minimum and we can recover the solution of one from the
other.
\begin{proof}
We will show that there exists a bijection between feasible points in both problems
with the same value for the objective.
First let $\otvar\in\RR^{\none\times\ntwo}$ be a feasible point for~\eqref{eqn:nndistdefnmain}.
Let $\otvarp\in\RR^{\nbarone\times\nbartwo}$ be such that its first $\none\times\ntwo$ block is
$\otvar$ and,
$\otvar_{\nbarone j} = \sum_{i=1}^{\none} \otvar_{ij},\;
 \otvar_{i \nbartwo} = \sum_{j=1}^{\ntwo} \otvar_{ij},$ and
$ \otvar_{\nbarone,\nbartwo} = \sum_{ij}\otvar_{ij}$.
Then, for all $i\leq \none$, $\sum_{j} \otvarp_{ij} = \laymass(j)$ and
$\sum_{j} \otvarp_{\nbarone j} \otvarp_{ij} = \sum_{j} \laymass(j) - \sum_{ij}\otvar_{ij} +
\otvar_{\nbarone,\nbartwo} = \totmass(\Gtwo)$.
We then have, $\otvarp\one_{\nbartwo} = y_1$
Similarly, we can show $\otvarp^\top \one_{\nbarone} = y_2$.
Therefore, $\otvarp$ is feasible for~\eqref{eqn:nndistdefntwo}.
We see that the objectives are equal via simple calculations,
\begin{align*}
\langle \otvarp, C'\rangle &=
\langle \otvarp, \mislabmatot + \strmatot\rangle + \langle \otvarp, \nasmatot \rangle
\numberthis
\label{eqn:objsareequalarg}
\\
& =
\langle \otvar, \mislabmatprob + \strmatprob\rangle +
      \sum_{j=1}^{\ntwo}\otvarp_{ij} + \sum_{i=1}^{\none}\otvarp_{ij}
\\
&=  \langle \otvar, \mislabmatprob \rangle + \langle \otvar, \strmatprob\rangle 
+\;\sum_{i\in\Lone} \big(\laymass(i) - \sum_{j\in\Ltwo} \otvar_{ij}\big)
+ \;\sum_{j\in\Ltwo} \big(\laymass(j) - \sum_{i\in\Lone} \otvar_{ij}\big).
\end{align*}
The converse also follows via a straightforward argument.
For given $\otvarp$ that is feasible for~\eqref{eqn:nndistdefntwo}, we let
$\otvar$ be the first $\none\times\ntwo$ block. By the equality constraints and non-negativity
of $\otvarp$, $\otvar$ is feasible for~\eqref{eqn:nndistdefnmain}.
By reversing the argument in~\eqref{eqn:objsareequalarg} we see that the objectives are
also equal. 
\end{proof}
\end{theorem}

\subsection{Distance Properties of \nndist}

The following theorem shows that the solution of~\eqref{eqn:nndistdefnmain} is
a pseudo-distance.
This is a formal version of Theorem~\ref{thm:metric} in the main text.

\vspace{0.1in}
\begin{theorem}
\label{thm:distthm}
Assume that the mislabel cost matrix $\mislabmat$ satisfies the triangle inequality; i.e.
for all labels \emph{\inlabelfont{x}, \inlabelfont{y}, \inlabelfont{z}} we have
\emph{
$\mislabmat(\textrm{\inlabelfont{x}}, \textrm{\inlabelfont{z}})
\leq
\mislabmat(\textrm{\inlabelfont{x}}, \textrm{\inlabelfont{y}}) +
\mislabmat(\textrm{\inlabelfont{y}}, \textrm{\inlabelfont{z}})$}.
Let $d(\Gone,\Gtwo)$ be the solution of~\eqref{eqn:nndistdefnmain} for networks
$\Gone,\Gtwo$.
Then $d(\cdot,\cdot)$ is a pseudo-distance. That is, for all networks
$\Gone,\Gtwo,\Gthree$, it satisfies, $d(\Gone,\Gtwo) > 0$,
$d(\Gone,\Gtwo) = d(\Gtwo,\Gone)$,
$d(\Gone,\Gone) = 0$ and
$d(\Gone,\Gthree) \leq d(\Gone, \Gtwo) + d(\Gtwo, \Gthree)$.
\end{theorem}

\insertFigPseudoDistanceIllus

Some remarks are in order.
First, observe that while
$d(\cdot,\cdot)$ is a pseudo-distance, it is not a distance;
i.e. $d(\Gone,\Gtwo) = 0\,\nRightarrow\, \Gone=\Gtwo$.
For example, while the networks in Figure~\ref{fig:nnpd} have different
descriptors according to our formalism in Section~\ref{sec:nngraphformalism},
their distance is $0$.
However, it is not hard to see that their functionality is the same -- in both cases,
the output of layer $1$ is passed through $16$ \convthree{} filters and then fed
to a layer with $32$ \convthree{} filters -- and hence, this property is desirable
in this example.
It is not yet clear however, if the topology induced by our metric equates two
functionally dissimilar networks.
We leave it to future work to study equivalence classes induced by the \nndists
distance.
Second, despite the OT formulation, this is not a Wasserstein distance.
In particular, the supports of the masses and the cost matrices change depending
on the two networks being compared.

\begin{proof}[\textbf{Proof of Theorem~\ref{thm:distthm}}]
We will  use the OT formulation~\eqref{eqn:nndistdefntwo} in this proof.
The first three properties are straightforward.
Non-negativity follows from non-negativity of $\otvarp, C'$ in~\eqref{eqn:nndistdefntwo}.
It is symmetric since the cost matrix for $d(\Gtwo,\Gone)$ is $C'^\top$ if the cost
matrix for $d(\Gone,\Gtwo)$ is $C$ and $\langle \otvarp, C' \rangle = \langle \otvarp^\top, C'^\top
\rangle$ for all $\otvarp$.
We also have $d(\Gone, \Gone) = 0$ since, then, $C'$ has a zero diagonal.

To prove the triangle inequality, we will use a gluing lemma,
similar to what is used in the proof of Wasserstein
distances~\citep{peyre2016ot}.
Let $\Gone,\Gtwo,\Gthree$ be given and $m_1, m_2, m_3$ be their total masses.
Let the solutions to $d(\Gone,\Gtwo)$ and $d(\Gtwo,\Gthree)$ be
$P\in\RR^{\nbarone\times\nbartwo}$ and
$Q\in\RR^{\nbartwo\times\nbarthree}$ respectively.
When solving~\eqref{eqn:nndistdefntwo}, we see that adding extra mass to the
non-assignment layers does not change the objective, as an optimiser can transport mass
between the two layers with $0$ cost.
Hence, we can assume w.l.o.g that~\eqref{eqn:nndistdefntwo} was
solved with
$y_i = \big[\{\laymass(u)\}_{u\in\Lcal_i}, \big(\sum_{j\in\{1,2,3\}} \totmass(\Gcal_j) -
\totmass(\Gcal_i)\big)\big] \in \RR^{\nbarii{i}}$ for $i=1,2,3$,
when computing the distances $d(\Gone,\Gtwo)$, $d(\Gone,\Gthree)$, $d(\Gtwo,\Gthree)$;
i.e. the total mass was $m_1+m_2+m_3$ for all three pairs.
We can similarly assume that $P,Q$ account for this extra mass, i.e.
$P_{\nbarone\nbartwo}$ and $Q_{\nbartwo\nbarthree}$ have been increased by $m_3$ and
$m_1$ respectively from their solutions in~\eqref{eqn:nndistdefntwo}.

\newcommand{\ytwoj}{(y_{2})_j}
To apply the gluing lemma, let $S = P \diag(1/y_2) Q \in \RR^{\nbarone\times\nbarthree}$,
where $\diag(1/y_2)$ is a diagonal matrix whose $(j,j)$\ssth element is $1/\ytwoj$
(note $y_2>0$).
We see that $S$ is feasible for~\eqref{eqn:nndistdefntwo} when computing
$d(\Gone,\Gthree)$,
\begin{align*}
R\one_{\nbarthree} &= P\diag(1/y_2)Q\one_{\nbarthree} 
 = P\diag(1/y_2)y_2 =
P\one_{\nbartwo} = y_1.
\end{align*}
Similarly, $R^\top\one_{\nbarone} = y_3$.
Now, let $U',V',W'$ be the cost matrices $C'$
in~\eqref{eqn:nndistdefntwo}
when computing $d(\Gone,\Gtwo)$, $d(\Gtwo,\Gthree)$, and $d(\Gone,\Gthree)$ respectively.
We will use the following technical lemma whose proof is given below.

\vspace{-0.05in}
\begin{lemma}
\label{lem:costmatrixtriangle}
For all $i\in\Lone,\,j\in\Ltwo,\,k\in\Lthree$, we have $W'_{ik} \leq U'_{ij} + V'_{jk}$.
\end{lemma}
\vspace{-0.05in}

\insertthmpostspacing
Applying Lemma~\ref{lem:costmatrixtriangle} yields the triangle inequality.
\begin{align*}
d(\Gone,\Gthree) &\,\leq \langle R, W' \rangle \;
=\, \sum_{i\in\Lone,k\in\Lthree} W'_{ik} \sum_{j\in\Ltwo} \frac{P_{ij}Q_{jk}}{\ytwoj}
\leq\,
\sum_{i,j,k} (U'_{ij} + V'_{jk})  \frac{P_{ij}Q_{jk}}{\ytwoj} \\
&=
\sum_{ij} \frac{U'_{ij} P_{ij}}{\ytwoj}\sum_{k}Q_{jk} \,+\
\sum_{jk} \frac{V'_{jk} Q_{jk}}{\ytwoj}\sum_{k}P_{ij} \\
&=\;
\sum_{ij} {U'_{ij} P_{ij}} \;+\;
\sum_{jk} {V'_{jk} Q_{jk}} 
\,=\, d(\Gone,\Gtwo) + d(\Gtwo,\Gthree)
\end{align*}
The first step uses the fact that $d(\Gone,\Gthree)$ is the minimum of all
feasible solutions and
the third step uses Lemma~\ref{lem:costmatrixtriangle}.
The fourth step rearranges terms and the fifth step uses
$P^\top\one_{\nbarone} = Q\one_{\nbarthree} = y_2$.
\end{proof}

\begin{proof}[\textbf{Proof of Lemma~\ref{lem:costmatrixtriangle}}]
Let $\Wp = \Wplmm + \Wpstr + \Wpnas$ be the decomposition into the label mismatch,
structural and non-assignment parts of the cost matrices;
define similar quantities $\Uplmm, \Upstr,\Upnas,\Vplmm, \Vpstr,\Vpnas$ for
$\Up,\Vp$.
Noting $a\leq b + c$ and $d\leq e + f$ implies $a + d \leq b + e + c + f$,
it is sufficient to show the triangle inquality for each component individually.
For the label mismatch term,
$(\Wplmm)_{ik} \leq (\Uplmm)_{ij} + (\Vplmm)_{jk}$ follows directly from
the conditions on $M$ by setting
$\textrm{\inlabelfont{x}} = \laylabel(i)$,
$\textrm{\inlabelfont{y}} = \laylabel(j)$,
$\textrm{\inlabelfont{z}} = \laylabel(k)$,
where $i,j,k$ are indexing in $\Lone,\Ltwo,\Lthree$ respectively.

For the non-assignment terms, when $(\Wpnas)_{ik} = 0$ the claim is true trivially.
$(\Wpnas)_{ik} = 1$, either when $(i=\nbarone, k\leq \nthree)$ or
$(i\leq \none, k = \nbarthree)$.
In the former case, when $j\leq \ntwo$, $(\Upnas)_{jk} = 1$ and when
$j=\nbartwo$, $(\Vpnas)_{\nbartwo} = 1$ as $k\leq \nthree$.
We therefore have, $(\Wpnas)_{ik} = (\Upnas)_{ij} + (\Vpnas)_{jk} = 1$.
A similar argument shows equality for the
$(i\leq \none, k = \nbarthree)$ case as well.

Finally, for the structural terms we note that $\Wpstr$ can be written
as $\Wpstr = \sum_t\Wpt$ as can $\Upt, \Vpt$.
Here $t$ indexes over the choices for the types of distances considered,
i.e. $t\in \{\textrm{sp, lp, rw}\} \times \{\textrm{ip, op}\}$.
It is sufficient to show
$(\Wpt)_{ik} \leq (\Upt)_{ij} + (\Vpt)_{jk}$.
This inequality takes the form,
\[
|\dploneit - \dplthreekt| \leq |\dploneit - \dpltwojt| + |\dpltwojt - \dplthreekt|.
\]
Where $\dplnnllt{g}{\ell}$ refers to distance type $t$ in network $g$ for layer $s$.
The above is simply the triangle inequality for real numbers.
This concludes the proof of Lemma~\ref{lem:costmatrixtriangle}.
\end{proof}

\subsection{Implementation \& Design Choices}
\label{app:nndistimplementation}

\textbf{Masses on the decision \& input/output layers:}
It is natural to ask why one needs to model the mass in the decision and input/output
layers.
For example, a seemingly natural choice is to use $0$ for these layers.
Using $0$ mass, is a reasonable strategy if we were to allow only one decision layer.
However, when there are multiple decision layers, consider comparing the
following two networks: 
the first has a feed forward MLP with non-linear layers, the second is
the same network but with an additional linear decision layer $u$, with one edge from 
$\ipnode$ to
$u$ and an edge from $u$ to $\opnode$.
This latter models the function as a linear + non-linear term
which might be suitable for some problems unlike modeling it only as a non-linear term.
If we do not add layer masses for the input/output/decision layers, then the distance
between
both networks would be $0$ - as there will be equal mass in the FF part for both networks
and
they can be matched with 0 cost.

\insertpathlengthalgo

\textbf{Computing path lengths $\distpathlength_s^t$:}
Algorithm~\ref{alg:rwpl} computes all path lengths in $O(|\edges|)$ time.
Note that topological sort of a connected digraph also takes $O(|\edges|)$ time.
The topological sorting ensures that $\distoprw$ is always computed for the children
in step 4.
For $\distopsp,\distoplp$ we would replace the averaging of $\Delta$ in step 5
with the minimum and maximum of $\Delta$ respectively.

\vspace{-0.05in}

For $\distiprw$ we make the following changes to Algorithm~\ref{alg:rwpl}.
In step 1, we set $\distiprw(\ipnode) = 0$, in step 3, we ${\rm pop\_first}$
and $\Delta$ in step 4 is computed using the parents.
$\distipsp,\distiplp$ are computed with the same procedure but by replacing the
averaging with minimum or maximum as above.

\textbf{Label Penalty Matrices:}
The label penalty matrices used in our \nnbos implementation, described below,
satisfy the triangle inequality condition in Theorem~\ref{thm:distthm}.

\insertLabPenCNN
\insertLabPenMLP

\underline{CNNs:}
Table~\ref{tb:mislabmatcnn} shows the label penalty matrix $M$ for used in
our CNN experiments
with labels \convthree, \convfive, \convseven, \maxpool, \avgpool, \softmax, \iplab,
\oplab.
\inlabelfont{conv$k$} denotes a $k\times k$ convolution while
\avgpool{} and \maxpool{} are pooling operations.
In addition, we also use \resthree, \resfive, \resseven{} layers which are inspired
by ResNets.
A \inlabelfont{res$k$} uses 2 concatenated \inlabelfont{conv$k$} layers but
the input to the first layer is added to the output of the second layer before the
relu activation -- See Figure 2 in~\citet{he2016deep}.
The layer mass for \inlabelfont{res$k$} layers is twice that of a  \inlabelfont{conv$k$} 
layer.
The costs for the \inlabelfont{res} in the label penalty matrix is the same
as the \inlabelfont{conv} block.
The cost between a \inlabelfont{res$k$} and  \inlabelfont{conv$j$} is
$\mislabmat(\textrm{\inlabelfont{res$k$}},\textrm{\inlabelfont{conv$j$}}) = 
0.9 \times \mislabmat(\textrm{\inlabelfont{conv$k$}},\textrm{\inlabelfont{conv$j$}}) + 
  0.1 \times 1$;
i.e. we are using a convex combination of the \inlabelfont{conv} costs and
the non-assignment cost.
The intuition is that a \inlabelfont{res$k$} is similar to \inlabelfont{conv$k$} block
except for the residual addition.

\underline{MLPs:}
Table~\ref{tb:mislabmatmlp} shows the label penalty matrix $M$ for used in
our MLP experiments
with labels \relu, \crelu, \leakyrelu, \softplus, \elu, \logistic, \tanhlabel, \linear, 
\iplab, \oplab.
Here  the first seven are common non-linear activations;
 \relu, \crelu, \leakyrelu, \softplus, \elu{}  rectifiers while
\logistic{} and \tanhlabel{} are sigmoidal activations.

\textbf{Other details:}
Our implementation of \nndists differs from what is described in the main text
in two ways.
First, in our CNN experiments, for a \fc{} layer $u$, we use
$0.1 \times \laymass(u) \times \langle\textrm{\#-incoming-channels}\rangle$
as the mass, i.e. we multiply it by $0.1$ from what is described in the main text.
This is because, in the convolutional and pooling channels, each unit is an image
where as in the \fc{} layers each unit is a scalar.
One could, in principle, account for the image sizes at the various layers when computing
the layer masses, but this also has the added complication of depending on the size of the
input image which varies from problem to problem.
Our approach is simpler and yields reasonable results.

Secondly, we use a slightly different form for $\strmatprob$.
First,
for $i\in\Lone,\;j\in\Ltwo$, we let
$\strmatproball(i,j) = \frac{1}{6}\sum_{s\in\{\textrm{sp, lp, rw}\}} $ $
\sum_{t\in\{\textrm{ip,op}\}} |\distpathlength^{s}_{t}(i) - \distpathlength^{s}_{t}(j)|$
be the
average of \emph{all} path length differences;
i.e. $\strmatproball$ captures the path length differences when considering all layers.
For CNNs,
we similarly construct matrices $\strmatprobconv, \strmatprobpool, \strmatprobfc$,
except they only consider the convolutional, pooling and fully connected layers
respectively in the path lengths. For $\strmatprobconv$, the distances to the output
(from the input) can be computed by zeroing outgoing (incoming) edges to layers
that are not convolutional.
We can similarly construct $\strmatprobpool$ and $\strmatprobfc$ only counting
the pooling and fully connected layers.
Our final cost matrix for the structural penalty is the average of these four matrices,
$\strmatprob = (\strmatproball + \strmatprobconv + \strmatprobpool + \strmatprobfc)/4$.
For MLPs, we adopt a similar strategy by computing matrices
$\strmatproball, \strmatprobrect, \strmatprobsigmoid$ with all layers, only
rectifiers, and only sigmoidal layers and let
$\strmatprob = (\strmatproball + \strmatprobrect + \strmatprobsigmoid)/3$.
The intuition is that by considering certain types of layers, we are
accounting for different types of information flow due to different operations.

\subsection{Some Illustrations of the \nndists Distance}
\label{app:nndistillustration}

We illustrate that \nndists computes reasonable distances on neural network architectures
via a two-dimensional t-SNE visualisation~\citep{maaten2008visualizing} of the
network architectures based.
Given a distance matrix between $m$ objects, t-SNE embeds them in a $d$ dimensional space
so that objects with small distances are placed closer to those that have larger
distances.
Figure~\ref{fig:tsne} shows the t-SNE embedding using the \nndists distance and its
noramlised version.
We have indexed 13 networks in both figures in a-n and displayed their architectures in
Figure~\ref{fig:tsne_nns}.
Similar networks are placed close to each other indicating
that \nndists induces a meaningful topology among neural network architectures.

\insertdistembedding
\insertdistembeddingnns

Next, we show that the distances induced by \nndists are correlated with validation error
performance.
In Figure~\ref{fig:distcorr} we provide the following scatter plot for networks trained
in our experiments for the Indoor, Naval and Slice datasets.
Each point in the figure is for pair of networks. The $x$-axis is the \nndists distance
between the pair
and the $y$-axis is the difference in the validation error on the dataset.
In each figure we used $300$ networks giving rise to $~45K$ pairwise points in each
scatter plot.
As the figure indicates, when the distance is small the difference in performance is close
to $0$.
However, as the distance increases, the points are more scattered.
Intuitively, one should expect that while networks that are far apart could perform
similarly or differently, similar networks should perform similarly.
Hence, \nndists induces a useful topology in the space of architectures that is smooth
for validaiton performance on real world datasets. 
This demonstrates that
it can be incorporated in a BO framework to
 optimise a network based on its validation error.

\insertdistcorrelations


\section{Implementation of \nnbo}
\label{app:implementation}

Here, we describe our BO framework for \nnbos in full detail.

\subsection{The Kernel}
\label{sec:kernel}

As described in the main text, we use a negative exponentiated distance as our kernel.
Precisely, we use,
\vspace{-0.05in}
\newcommand{\kernelarg}{(\cdot,\cdot)}
\begin{align*}
\kernel\kernelarg =
  \alpha e^{-\sum_{i}\beta_i \disti^p\kernelarg}
 + \alphabar e^{-\sum_{i}\betabar_i \distbari^{\pbar}\kernelarg}.
\numberthis
\label{eqn:otmannkernel}
\end{align*}

\vspace{-0.10in}

Here, $\disti,\distbari$, are the \nndists distance and its normalised counterpart
developed in Section~\ref{sec:nndistmain},
computed with different values for $\structpencoeff\in\{\structpencoeffi\}_i$.
$\beta_i,\betabar_i$ manage the relative contributions of $\disti,\distbari$,
while $(\alpha,\alphabar)$ manage the contributions of each kernel in the sum.
An ensemble approach of the above form, instead of trying to pick a single best value,
ensures that \nnbos accounts for the different
topologies induced by the different distances  $\disti,\distbari$.
In the experiments we report, we used $\{\structpencoeffi\}_i = \{0.1, 0.2, 0.4, 0.8\}$,
 $p=1$ and $\pbar=2$. Our experience suggests that \nnbos was not 
particularly sensitive to these
choices expect when we used only very large or only very small values in
$\{\structpencoeffi\}_i$.

\nnbo, as described above has $11$ hyper-parameters of its own;
$\alpha, \alphabar, \{(\beta_i, \betabar_i)\}_{i=1}^4$ and the GP noise variance $\eta^2$.
While maximising the GP marginal likelihood is a common approach to pick 
hyper-parameters, this might cause over-fitting when there are many of them.
Further, as training large neural networks is typically expensive, we have to
content with few observations for the GP in practical settings.
Our solution is to start with a (uniform) prior over these hyper-parameters and sample
hyper-parameter values from the posterior under the GP
likelihood~\citep{snoek12practicalBO}, which we found to be robust.
While it is possible to treat $\structpencoeff$ itself as a hyper-parameter of the kernel,
this will require us to re-compute all pairwise distances of networks that
have  already  been evaluated each time we change the hyper-parameters.
On the other hand, with the above approach, we can compute and store distances for
different $\structpencoeffi$ values whenever a new network
is evaluated, and then compute the kernel cheaply
for different values of $\alpha, \alphabar, \{(\beta_i, \betabar_i)\}_{i}$.

\vspace{-0.05in}
\subsection{Optimising the Acquisition}
\vspace{-0.05in}
\label{sec:acqopt}

We use a evolutionary algorithm (\evoalg) approach to optimise the
acquisition function~\eqref{eqn:eiacq}.
We begin with an initial pool of networks and evaluate the acquisition
$\acqt$ on those networks.
Then we generate a set of $\Nmut$ mutations of this pool as follows.
First, we stochastically select $\Nmut$ candidates from the set of networks
already evaluated such that those with higher $\acqt$ values are more likely to be
selected than those with lower values.
Then we apply a mutation operator to each candidate,
to produce a modified architecture.
Finally, we evaluate the acquisition on this $\Nmut$ mutations, add it to the initial pool,
and repeat for the prescribed number of steps.

\textbf{Mutation Operator:}
To describe the mutation operator,
we will first define a library of modifications to a neural network.
These modifications, described in Table~\ref{tb:nnmodifiers},
might change the architecture either by increasing or decreasing the number of computational
units in a layer, by adding or deleting layers, or by changing the connectivity of
existing layers.
They provide a simple mechanism to explore the space of architectures that are close
to a given architecture.
The \emph{one-step mutation operator} takes a given network and applies one of the
modifications in Table~\ref{tb:nnmodifiers} picked at random to produce a new network.
The \emph{$k$-step mutation operator} takes a given network, and repeatedly applies
the one-step operator $k$ times -- the new network will have undergone $k$ changes
from the original one. 
One can also define a compound operator, which picks the number of steps
probabilistically.
In our implementation of \nnbo, we used such a compound operator with probabilities
$(0.5, 0.25, 0.125, 0.075, 0.05)$;
i.e. it chooses a one-step operator with probability $0.5$, a $4$-step operator with
probability $0.075$, etc.
Typical implementations of EA in Euclidean spaces define the mutation operator
via a Gaussian (or other) perturbation of a chosen candidate.
It is instructive to think of the probabilities for each step in our scheme above
as being analogous to the width of the Gaussian chosen for perturbation.%

\insertTableEAModifiers

\textbf{Sampling strategy:}
The sampling strategy for \evoalgs is as follows.
Let $\{\ztt{i}\}_{i}$, where $\ztt{i}\in\Xcal$ be the points evaluated so far.
We sample $\Nmut$ new points 
from  a distribution $\pi$  where $\pi(\ztt{i}) \propto \exp(g(\ztt{i})/\sigma$).
Here $g$ is the function to be optimised (for \nnbo, $\acqt$ at time $t$).
$\sigma$ is the standard deviation of all previous evaluations.
As the probability for large $g$ values is higher, they are more likely to get selected.
$\sigma$ provides normalisation to account for different ranges of function values.

Since our candidate selection scheme at each step favours networks that have
high acquisition value, our EA scheme is more likely to search at regions that
are known to have high acquisition.
The stochasticity in this selection scheme
and the fact that we could take multiple steps in the
mutation operation ensures that we still sufficiently explore the space.
Since an evaluation of $\acqt$ is cheap, we can use many \evoalgs steps to
explore several architectures and optimise $\acqt$.

\textbf{Other details:}
The \evoalgs procedure is also initialised with the same initial pools in
Figures~\ref{fig:initPoolCNN},~\ref{fig:initPoolMLP}.
In our \nnbos implementation, we increase the
total number of \evoalgs evaluations $n_{\text{\evoalgs}}$
at rate $\bigO(\sqrt{t})$ where $t$ is the current
time step in \nnbo.
We set $\Nmut$ to be  $\bigO(\sqrt{n_{\text{\evoalgs}}})$.
Hence, initially we are only considering a small neighborhood around the initial pool,
but as we proceed along BO,  we expand to a larger region, and also
spend more effort to optimise $\acqt$.

\textbf{Considerations when performing modifications:}
The modifications in Table~\ref{tb:nnmodifiers} is straightforward in MLPs.
But in CNNs one needs to ensure that the image sizes are the
same when concatenating them as an input to a layer.
This is because strides can shrink the size of the image.
When we perform a modification we check if this condition is violated and if so, disallow
that modification.
When a \skiplayer{} modifier attempts to add a connection from a layer with a large image
size to one with a smaller one, we add \avgpool{} layers at stride 2 so that
the connection can be made (this can be seen, for e.g. in the second network in
Fig.~\ref{fig:bestNetCifarHEI}).

\subsection{Other Implementation Details}

\textbf{Initialisation:}
We initialise \nnbos (and other methods) with an initial pool of $10$ networks.
These networks are illustrated in Fig.~\ref{fig:initPoolCNN} for CNNs and
Fig.~\ref{fig:initPoolMLP} for MLPs at the end of the document.
These are the same networks used to initialise the \evoalgs procedure to optimise
the acquisition.
All initial networks have feed forward structure.
For the CNNs, the first $3$ networks have structure similar to the VGG
nets~\citep{simonyan2014very} and the remaining have blocked feed forward structures
as in~\citet{he2016deep}.
We also use blocked structures for the MLPs with the layer labels decided arbitrarily.

\textbf{Domain:}
For \nnbo, and other methods,
we impose the following constraints on the search space.
If the \evoalgs modifier (explained below) generates a network that violates these
constraints, we simply skip it.
\begin{itemize}
\item Maximum number of layers: $60$
\item Maximum mass: $10^8$
\item Maximum in/out degree: $5$
\item Maximum number of edges: $200$
\item Maximum number of units per layer: $1024$
\item Minimum number of units per layer: $8$
\end{itemize}

\textbf{Layer types:} We use the layer types detailed in
Appendix~\ref{app:nndistimplementation} for both CNNs and MLPs.
For CNNs, all pooling operations are done at stride 2.
For convolutional layers, we use either stride 1 or 2 (specified in the
illustrations).
For all layers in a CNN we use \relu{} activations.

\textbf{Parallel BO:}
We use a parallelised experimental set up where multiple models can be
evaluated in parallel.
We handle parallel BO via the hallucination technique
in~\citet{ginsbourger2011dealing}.

Finally, we emphasise that many of the above choices were made arbitrarily, and we
were able to get \nnbos working efficiently with our first choice for these
parameters/specifications.
Note that
many end-to-end systems require specification of such choices.


\section{Addendum to Experiments}
\label{app:experiments}

\subsection{Baselines}

\rand:
Our \rands implementation, operates in exactly the same way as \nnbo,
except that the \evoalgs procedure (Sec.~\ref{sec:acqopt})
is fed a random sample from $\unif(0,1)$ instead of
the GP acquisition each time it evaluates an architecture.
That is, we follow the same schedule for $n_{\text{\evoalg}}$ and $\Nmut$ as we did for
\nnbos.
Hence \rands has the opportunity to explore the same space as \nnbo, but
picks the next evaluation randomly from this space.

\evoalg: This is as described in Appendix~\ref{app:implementation} except that we
fix $\Nmut=10$ all the time.
In our experiments where we used a budget based on time, it was difficult to predict
the total number of evaluations so as to set $\Nmut$ in perhaps a more intelligent way.

\treebo: As the implementation from~\citet{jenatton2017bayesian} was not made available,
we wrote our own.
It differs from the version described in the paper in a few ways.
We do not tune for a regularisation penalty and step size as they do to keep it
line with the rest of our experimental set up.
We set the depth of the network to $60$ as we allowed $60$ layers for the other methods.
We also check for the other constraints given in Appendix~\ref{app:implementation}
before evaluating a network.
The original paper uses a tree structured kernel which can allow for efficient
inference with a large number of samples.
For simplicity, we construct the entire kernel matrix and perform standard GP inference.
The result of the inference is the same, and the number of GP samples was always below
$120$ in our experiments so a sophisticated procedure was not necessary.

\subsection{Details on Training}
In all methods, for each proposed network architecture, we trained the network
on the train data set, and periodically evaluated its performance on the
validation data set. For MLP experiments, we optimised network parameters using
stochastic gradient descent with a fixed step size of $10^{-5}$ and
a batch size of 256 for 20,000 iterations.
We computed the validation set MSE every 100 iterations; from this we returned
the minimum MSE that was achieved.
For CNN experiments, we optimised network
parameters using stochastic gradient descent with a batch size of 32.
We started with a learning rate of $0.01$ and reduced it gradually.
We also used batch normalisation and trained the model for
$60,000$ batch iterations.
We computed the validation set classification error every $4000$ iterations;
from this we returned the minimum classification error that was achieved.

After each method returned an optimal neural network architecture, we again
trained each optimal network architecture on the train data set, periodically
evaluated its performance on the validation data set, and finally computed the
MSE or classification error on the test data set. For MLP experiments, we used
the same optimisation procedure as above; we then computed the test set MSE at
the iteration where the network achieved the minimum validation set MSE. For
CNN experiments, we used the same optimisation procedure as above, except here
the optimal network architecture was trained for 120,000 iterations; we then
computed the test set classification error at the iteration where the network
achieved the minimum validation set classification error.

\subsection{Optimal Network Architectures and Initial Pool}
Here we illustrate and compare the optimal neural network architectures found
by different methods. In Figures
\ref{fig:bestNetCifarHEI}-\ref{fig:bestNetCifarTREE}, we show some optimal
network architectures found on the Cifar10 data by \nnbo, \evoalg, \rand, and
\treebo, respectively.  We also show some optimal network architectures found
for these four methods on the Indoor data, in Figures
\ref{fig:bestNetIndoorHEI}-\ref{fig:bestNetIndoorTREE}, and on the Slice data,
in Figures \ref{fig:bestNetSliceHEI}-\ref{fig:bestNetSliceTREE}.
A common feature among all optimal architectures found by \nnbos was the presence of long
skip connections and multiple decision layers.

In Figure~\ref{fig:initPoolMLP}, we show the initial pool of MLP network
architectures, and in Figure~\ref{fig:initPoolCNN}, we show the initial pool of
CNN network architectures.
On the Cifar10 dataset, VGG-19 was one of the networks in the initial pool.
While all methods beat VGG-19 when trained for 24K iterations (the number of iterations we
used when picking the model), \treebo{} and \rand{} lose to VGG-19 (see
Section~\ref{sec:experiments} for details).
This could be because the performance after shorter training periods may not exactly
correlate with performance after longer training periods.

\subsection{Ablation Studies and Design Choices}

We conduct experiments comparing the various design choices in \nnbo.
Due to computational constraints, we carry them out on synthetic functions.

In Figure~\ref{fig:ablkernel}, we compare \nnbos using only the normalised distance,
only the unnormalised distance, and the combined kernel as in~\eqref{eqn:otmannkernel}.
While the individual distances performs well, the combined form outperforms both.

Next, we modify our \evoalgs procedure to optimise the acquisition.
We execute \nnbos using only the \evoalgs modifiers which change the
computational units (first four modifiers in Table~\ref{tb:nnmodifiers}),
then using the
modifiers which only change the structure of the networks (bottom 5 in
Table~\ref{tb:nnmodifiers}),
and finally using all 9 modifiers, as used in all our experiments.
The combined version outperforms the first two.

Finally, we experiment with different choices for $p$ and $\pbar$
in~\eqref{eqn:otmannkernel}.
As the figures indicate, the performance was not particularly sensitive to these choices.

\newcommand{\avgmass}{{\rm am}}
\newcommand{\indeg}{\rm deg_{i}}
\newcommand{\outdeg}{\rm deg_{o}}
\newcommand{\distipop}{\rm \delta}
\newcommand{\avgstride}{\rm {str}}
\newcommand{\fracconv}{\rm {frac\_conv3}}
\newcommand{\fracsigmoid}{\rm {frac\_sigmoid}}

Below we describe the three synthetic functions $f_1, f_2, f_3$ used in our synthetic
experiments. 
$f_3$ applies for CNNs while $f_1, f_2$ apply for MLPs.
Here $\avgmass$ denotes the average mass per layer, 
$\indeg$ is the average in degree the  layers,
$\outdeg$ is the average out degree,
$\distipop$ is the shortest distance from $\ipnode$ to $\opnode$,
$\avgstride$ is the average stride in CNNS,
$\fracconv$ is the fraction of layers that are \convthree,
$\fracsigmoid$ is the fraction of layers that are sigmoidal.

\begin{align*}
f_0 &=  \exp(-0.001 * |\avgmass - 1000|) +
         \exp(-0.5 * |\indeg - 5|) + 
         \exp(-0.5 * |\outdeg - 5|) + \\
    &\hspace{0.3in}
         \exp(-0.1 * |\distipop - 5|) +
         \exp(-0.1 * ||\Lcal| - 30|) +
         \exp(-0.05 * ||\Ecal| - 100|) \\
f_1 &= f_0 + \exp(-3 * |\avgstride - 1.5|) +
         \exp(-0.3 * ||\Lcal| - 50|) + \\
&\hspace{0.3in}
        \exp(-0.001 * |\avgmass - 500|) + 
        \fracconv \\
f_2 &= f_0 + 
        \exp(-0.001 * |\avgmass - 2000|) + 
         \exp(-0.1 * ||\Ecal| - 50|) +
        \fracsigmoid \\
f_3 &= f_0 + \fracsigmoid
\end{align*}

\insertFigAblation

\section{Additional Discussion on Related Work}

Historically, 
evolutionary (genetic) algorithms (\evoalg) have been the most common method used for
designing architectures~\citep{%
miikkulainen2017evolving,kitano1990designing,floreano2008neuroevolution,xie2017genetic,%
liu2017hierarchical,real2017large,stanley2002evolving}.
\evoalgs techniques are popular as they provide a simple mechanism to explore
the space of architectures by making a sequence of changes to networks that
have already been evaluated.  However, as we will discuss later, \evoalgs
algorithms, while conceptually and computationally simple, are typically not
best suited for optimising functions that are expensive to evaluate.
A related line of work first sets up a search space for architectures via incremental
modifications, and then explores this space via random exploration, MCTS, or A*
search~\citep{liu2017progressive,negrinho2017deeparchitect,cortes2016adanet}.
Some of the methods above can only optimise among feed forward structures, e.g.
Fig.~\ref{fig:mainnneg1}, but cannot handle spaces with arbitrarily structured 
networks,
e.g. Figs.~\ref{fig:mainnneg2}, ~\ref{fig:mainnneg3}.

The most successful recent architecture search methods that can handle
arbitrary structures have adopted reinforcement
learning (RL)~\citep{baker2016designing,zoph2016neural,zoph2017learning,zhong2017practical}.
However, architecture search is in essence an \emph{optimisation} problem --
find the network with the highest function value.
There is no explicit need to maintain a notion of state and solve the credit assignment
problem in RL~\citep{sutton1998reinforcement}.
Since RL is fundamentally more difficult than optimisation~\citep{jiang2016contextual},
these methods
typically need to try a very large number of architectures to find the optimum.
This is not desirable, especially in computationally constrained settings.


\newpage

\insertFigBestNetCifarHEI

\insertFigBestNetCifarGA

\insertFigBestNetCifarRAND

\insertFigBestNetCifarTREE


\insertFigBestNetIndoorHEI

\insertFigBestNetIndoorGA

\insertFigBestNetIndoorRAND

\insertFigBestNetIndoorTREE


\insertFigBestNetSliceHEI

\insertFigBestNetSliceGA

\insertFigBestNetSliceRAND

\insertFigBestNetSliceTREE


\insertFigInitialPoolCNN

\insertFigInitialPoolMLP

\end{document}